\documentclass{article} %
\usepackage{iclr2025_conference,times}

\usepackage[utf8]{inputenc} %
\usepackage[T1]{fontenc}    %
\usepackage[colorlinks=true,
            linkcolor=myblue,
            citecolor=myblue,
            urlcolor=red,
            ]{hyperref}       %
\usepackage{url}            %
\usepackage{booktabs}       %
\usepackage{amsfonts}       %
\usepackage{nicefrac}       %
\usepackage{microtype}      %
\usepackage{xcolor}         %
\usepackage{wrapfig}
\usepackage{multirow}
\usepackage{lipsum} %
\usepackage{graphicx}
\usepackage{subcaption}

\usepackage{amsmath}
\usepackage{amssymb}
\usepackage{mathtools}
\usepackage{amsthm}
\usepackage{wasysym}
\usepackage{makecell}
\usepackage{changepage,threeparttable}

\usepackage{array}
\usepackage{epsfig}
\usepackage{graphicx}
\usepackage{float}
\usepackage{wrapfig}
\usepackage{amsmath,amssymb,amsthm}
\usepackage{bm,xspace}
\usepackage{comment}
\usepackage{multirow}
\usepackage{balance}
\usepackage{url}
\usepackage{booktabs}
\usepackage{etoolbox,siunitx}
\usepackage{calc}
\usepackage{pifont,hologo}
\usepackage{color}
\usepackage{adjustbox}
\usepackage{amsmath}
\usepackage{enumitem}
\usepackage{subcaption}
\usepackage{titlesec}
\usepackage[normalem]{ulem}  %
\PassOptionsToPackage{table}{xcolor}
\usepackage{colortbl}
\usepackage[table]{xcolor}

\newcommand\rebuttal[1]{\textcolor{black}{#1}}

\definecolor{myblue}{HTML}{118ab2}
\definecolor{red}{HTML}{ef476f}
\definecolor{orange}{HTML}{cc7700}
\definecolor{mygray}{HTML}{efefef}
\definecolor{darkgreen}{HTML}{228B22}
\definecolor{mydarkgray}{HTML}{757575}

\newcommand{\figref}[1]{Fig.~\ref{#1}}
\newcommand{\tabref}[1]{Tab.~\ref{#1}}
\newcommand{\secref}[1]{Sec.~\ref{#1}}
\renewcommand{\eqref}[1]{Eq.~\ref{#1}}

\newcolumntype{x}[1]{>{\centering\arraybackslash}p{#1}}
\newcolumntype{y}[1]{>{\raggedright\arraybackslash}p{#1}}
\newcolumntype{z}[1]{>{\raggedleft\arraybackslash}p{#1}}
\newcommand{\tablestyle}[2]{\setlength{\tabcolsep}{#1}\renewcommand{\arraystretch}{#2}\centering\footnotesize}

\DeclareMathSymbol{@}{\mathord}{letters}{"3B}

\newcommand{\citesplit}[1]{\vspace{-0.3em}\tiny{(\citeauthor{#1}},\\\tiny{\citeyear{#1})}}

\makeatletter
\DeclareRobustCommand\onedot{\futurelet\@let@token\@onedot}
\def\@onedot{\ifx\@let@token.\else.\null\fi\xspace}
\def\eg{\emph{e.g}\onedot}

\newcommand*{\Rom}[1]{\expandafter\@slowromancap\romannumeral #1@}
\newcommand*{\rom}[1]{\expandafter\romannumeral #1}

\def\1{\bm{1}}

\newcommand{\Ls}{\mathcal{L}}

\let\originalleft\left
\let\originalright\right
\renewcommand{\left}{\mathopen{}\mathclose\bgroup\originalleft}
\renewcommand{\right}{\aftergroup\egroup\originalright}

\definecolor{resnet}{HTML}{264653}
\definecolor{r3m}{HTML}{264653}
\definecolor{vit}{HTML}{2A9D8F}
\definecolor{vc1}{HTML}{2A9D8F}
\definecolor{multivit}{HTML}{E9C46A}
\definecolor{multimae}{HTML}{E9C46A}
\definecolor{spunet}{HTML}{F4A261}
\definecolor{ponderv2}{HTML}{F4A261}
\definecolor{pointnet}{HTML}{E76F51}
\definecolor{cyan}{HTML}{264653}

\definecolor{pink}{HTML}{FF7096}

\usepackage{pifont}%
\newcommand{\cmark}{\ding{51}}%
\newcommand{\xmark}{\ding{55}}%

\usepackage{hyperref}

\title{{\includegraphics[height=1.25em]{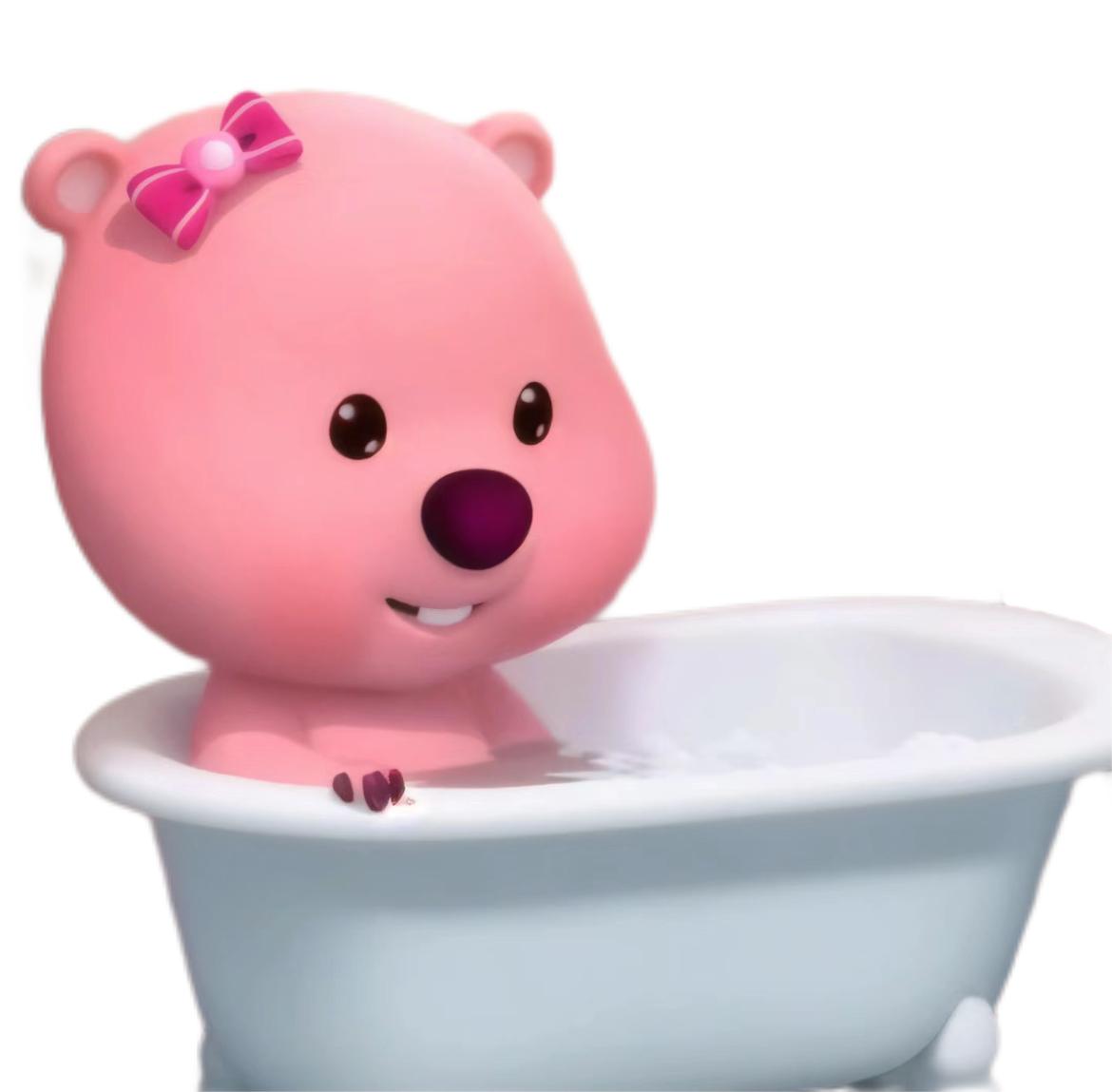}} \textcolor{pink}{\textbf{SPA}}: 3D \textcolor{pink}{SP}atial-\textcolor{pink}{A}wareness Enables \\Effective Embodied Representation}

\author{
  Haoyi Zhu\textsuperscript{1,2}, 
  Honghui Yang\textsuperscript{2,3}, 
  Yating Wang\textsuperscript{2,4},  %
  Jiange Yang\textsuperscript{2,5}, 
  Limin Wang\textsuperscript{2,5}, 
  Tong He\textsuperscript{2\dag} \\
  \textsuperscript{1}USTC, \textsuperscript{2}Shanghai AI Lab, \textsuperscript{3}ZJU, \textsuperscript{4}Tongji, \textsuperscript{5}NJU \\
  \textsuperscript{\dag} Corresponding Author
}

\iclrfinalcopy %
\begin{document}

\maketitle

\begin{abstract}

In this paper, we introduce SPA, a novel representation learning framework that emphasizes the importance of 3D spatial awareness in embodied AI. Our approach leverages differentiable neural rendering on multi-view images to endow a vanilla Vision Transformer (ViT) with intrinsic spatial understanding. We present the most comprehensive evaluation of embodied representation learning to date, covering 268 tasks across 8 simulators with diverse policies in both single-task and language-conditioned multi-task scenarios. The results are compelling: SPA consistently outperforms more than 10 state-of-the-art representation methods, including those specifically designed for embodied AI, vision-centric tasks, and multi-modal applications, while using less training data. Furthermore, we conduct a series of real-world experiments to confirm its effectiveness in practical scenarios. These results highlight the critical role of 3D spatial awareness for embodied representation learning. Our strongest model takes more than 6000 GPU hours to train and we are committed to open-sourcing all code and model weights to foster future research in embodied representation learning. Project Page: \url{https://haoyizhu.github.io/spa/}.

\end{abstract}

\begin{figure}[!htb]
\centering
\includegraphics[width=0.85\linewidth]{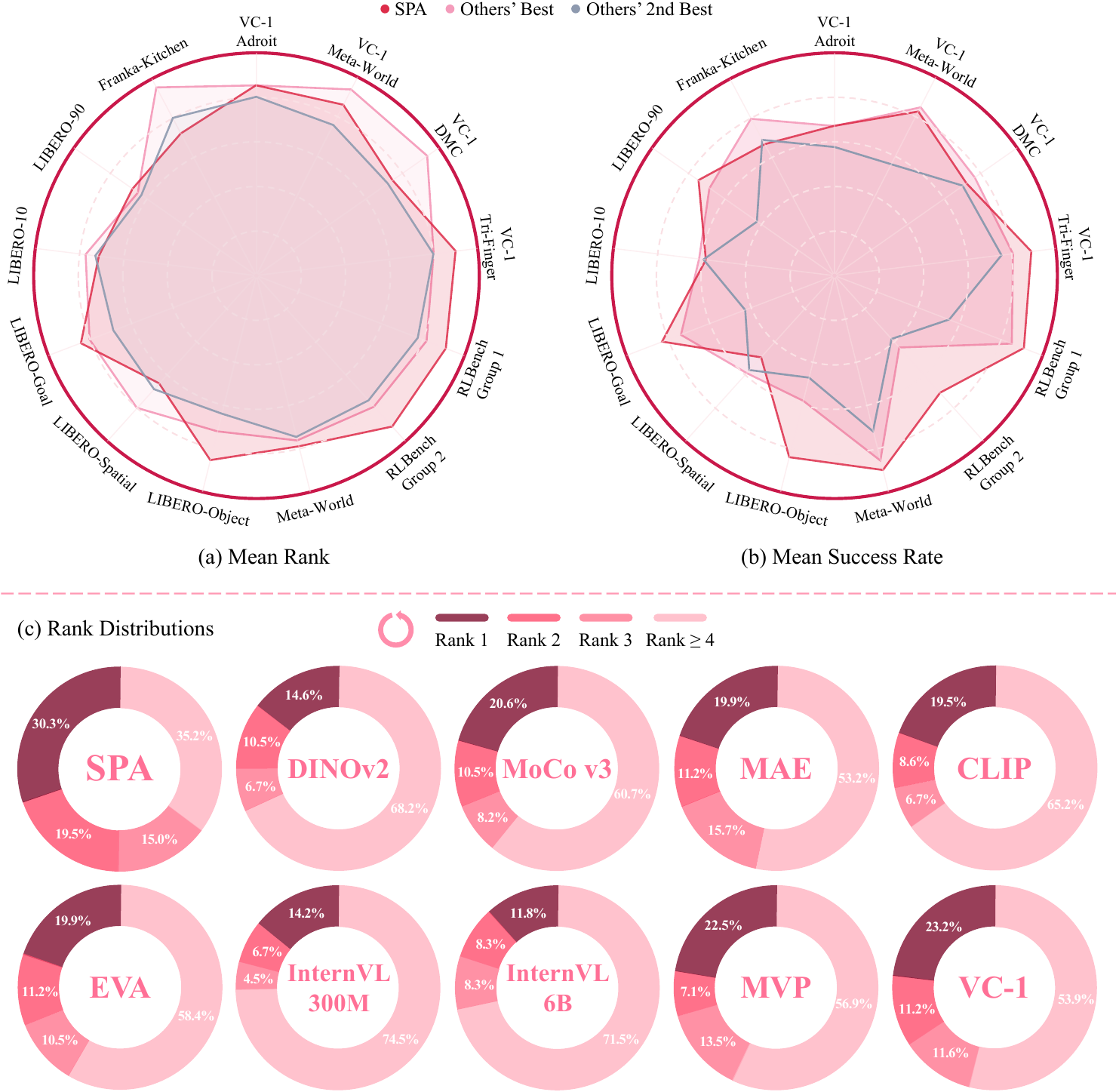}
\vspace{-0.9em}
\caption{\rebuttal{\textbf{Performance comparison across representations.}} \textit{Above}: (a) Mean rank and (b) mean success rate on benchmarks. Lines represent the performance of SPA, best, and second best performance on each benchmark. \textit{Bottom}: Rank distributions for 268 individual tasks, showing proportions from rank 1 to rank $\geq$ 4 counterclockwise. Our model demonstrates superior overall performance.}
\label{fig:radar_plot}
\vspace{-1.8em}
\end{figure}

\section{Introduction}

Vision systems have made remarkable progress in understanding 2D images~\citep{he2020momentum,chen2020simple,he2022masked,feichtenhofer2022masked,tong2022videomae,yang2023comae,oquab2023dinov2,radford2021learning,fang2023eva,chen2024internvl}. However, achieving true visual intelligence necessitates a comprehensive understanding of the 3D world. This is crucial for embodied AI, where agents must perceive, reason, and interact with complex 3D environments. 

Existing visual representation learning methods for embodied AI~\citep{nair2022r3m,radosavovic2023real,majumdar2023we,karamcheti2023voltron,shang2024theia,yang2024spatiotemporal} largely rely on paradigms from 2D vision, predominantly employing contrastive-based or masked autoencoder (MAE)-based approaches. 
However, they often struggle to fully capture the spatial relationships and 3D structures inherent in the physical world. This limitation arises from their primary emphasis on 2D semantic understanding, which, though valuable, is still insufficient for the sophisticated spatial reasoning required in embodied AI tasks, where agents need to navigate environments, manipulate objects, and make decisions using their 3D spatial awareness.

In this paper, we introduce \textcolor{pink}{\textbf{SPA}}, a general 3D spatial-aware representation learning framework for embodied AI. SPA leverages neural rendering~\citep{mildenhall2021nerf} as the pre-training pre-text task on multi-view images.
Unlike explicit 3D representations like point clouds or meshes—which prior work~\citep{wang2024rise,wang2024dexcap,ze20243d,zhu2024point} has shown to outperform pure 2D inputs in robot learning—multi-view images are easier to process and more readily available, making them ideal for large-scale training, such as from internet videos.
Specifically, given a vanilla 2D image backbone, \eg a Vision Transformer (ViT)~\citep{dosovitskiy2020vit},
we first extract multi-view feature maps from the input images.
Using known camera poses, we then construct a feature volume from these feature maps and sample rays to apply differentiable neural rendering.
This process generates multi-view RGB-D images and semantic maps for supervision without labels, enabling the pre-training of a 2D image backbone to enhance 3D spatial awareness.

To thoroughly validate our assumption and method, we collect 268 embodied tasks across 8 simulators using various policy methods. To our knowledge, this represents \textbf{the largest scale of embodied evaluation to date}. Previous work, such as R3M~\citep{nair2022r3m} and VC-1~\citep{majumdar2023we}, evaluated fewer than 20 tasks, potentially leading to incomplete or biased conclusions. Our evaluation spans both single-task and language-conditioned multi-task learning. We compare over 10 state-of-the-art representation learning methods, categorized as embodied-specific~\citep{nair2022r3m,majumdar2023we,radosavovic2023real}, vision-centric~\citep{oquab2023dinov2,chen2021empirical,he2022masked}, and multi-modal~\citep{radford2021learning,fang2023eva,chen2024internvl}. Our method consistently outperforms others, underscoring the importance of 3D spatial awareness for embodied AI. Notably, multi-modal models like CLIP~\citep{radford2021learning}, consistently perform poorly. This holds even the vision-language model scales the ViT to 6B parameters~\citep{chen2024internvl}. 
Through a camera pose estimation task and feature map visualization, we demonstrate that SPA has learned superior 3D spatial understanding. Further, we find that 3D awareness shows a positive correlation with embodied performance.
Finally, we conduct several real-world tasks, where SPA also demonstrates superior performance.
Our contribution can be summarized as follows.
\begin{itemize}
    \item We propose a significant \textit{spatial hypothesis}: 3D spatial awareness is crucial for embodied representation learning. Our experiments provide clear evidence for the hypothesis.
    \item We introduce SPA, a novel paradigm for representation learning in embodied AI. It enhances a vanilla Vision Transformer (ViT) with 3D awareness using differentiable neural rendering as the pre-text task on multi-view images.
    \item We conduct the largest evaluation benchmark for embodied representation learning, significantly larger than previous studies. It involves 268 tasks, 8 simulators, and over 10 SOTA methods with diverse downstream policies and task settings.
    \item Through extensive experiments in both simulators and real-world settings, SPA outperforms more than 10 SOTA representation learning methods, demonstrating its effectiveness.
\end{itemize}

\section{Methodology}
\label{sec:method}
\begin{figure}[!tb]
\centering
\includegraphics[width=0.95\linewidth]{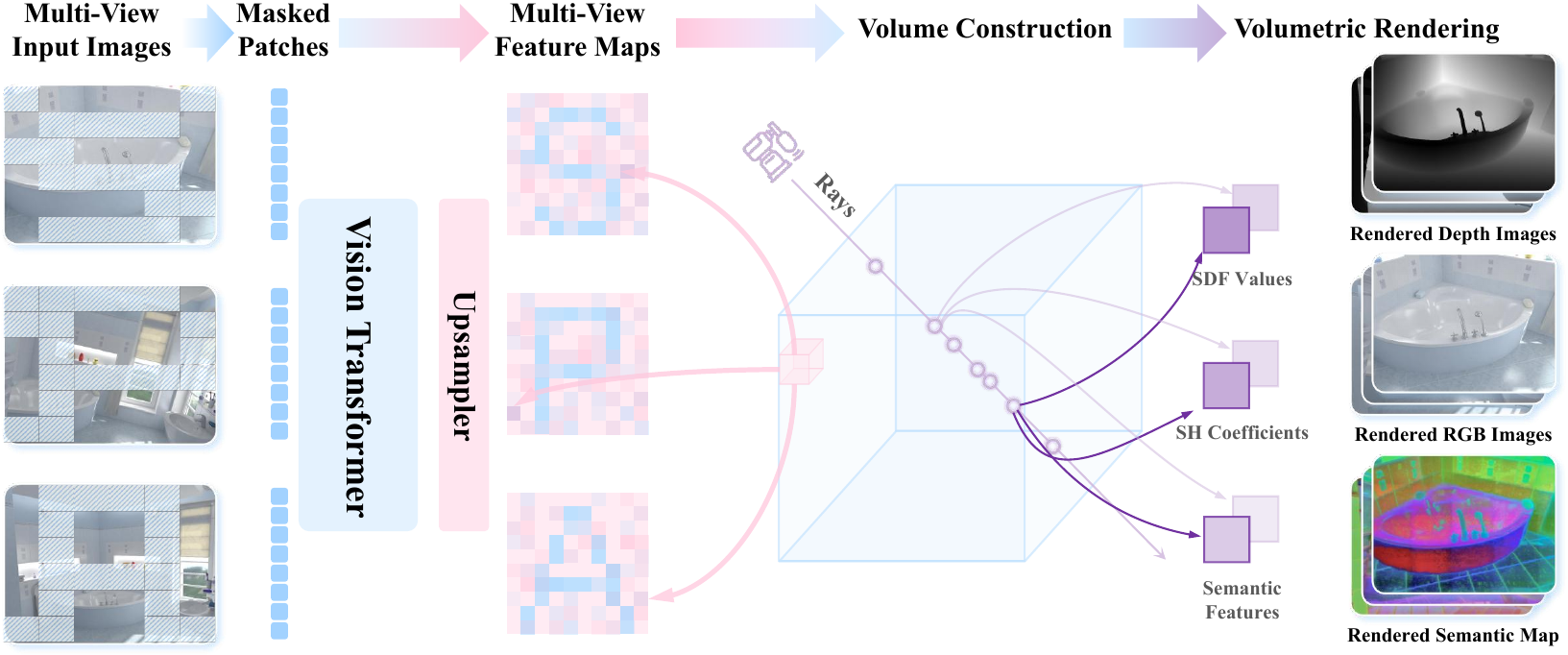}
\vspace{-0.75em}
\caption{\textbf{Pipeline Overview.} Given multi-view images, we randomly mask patches and input the remaining into a Vision Transformer. The upsampled latent features generate multi-view feature maps, from which we construct a feature volume to derive SDF values, SH coefficients, and semantic features. We then render depth, RGB, and semantic maps for loss computation.}
\label{fig:pipeline}
\vspace{-1.2em}
\end{figure}

In this section, %
we first describe our process for handling multi-view image inputs and feature extraction in Sec.~\ref{sec:method-input-pre-process}. Subsequently, we construct an explicit feature volume from these multi-view features, detailed in Sec.~\ref{sec:method-volume-construction}. Finally, we explain the image rendering from the feature volume and loss functions for network optimization in Sec.~\ref{sec:method-rendering} and Sec.~\ref{sec:method-loss}. Our pipeline is visualized in \figref{fig:pipeline}.

\subsection{Input Process and Feature Extraction}
\label{sec:method-input-pre-process}
Given a set of multi-view images $\mathbf{I} = \{I_1, I_2, \ldots, I_N\}$, where each $I_i \in \mathbb{R}^{3 \times H \times W}$ and $N \in \mathbb{Z}^+$, we utilize a 2D image backbone $F$, such as a ViT. The images are processed separately through $F$, yielding latent features $\mathbf{L} = \{l_1, l_2, \ldots, l_N\}$, where each $l_i = F(I_i) \in \mathbb{R}^{L \times C}$. Following MAE, we apply random masking to input images to enhance robustness, but without a ViT decoder and MAE's pixel reconstruction objective.
For each $l_i$, masked positions are filled with a mask token, and we concatenate the global class token with other patch tokens as read-out tokens similar to DPT~\citep{Ranftl2020}.
We then unpatchify them to obtain a latent feature map of size $\frac{H}{P} \times \frac{W}{P}$, where $P$ is the ViT patch size. Finally, two simple upsampling layers transform this into a feature map $M_i$ matching the input resolution. Each upsampling layer includes a convolution, a GELU~\citep{hendrycks2016gaussian} activation, and a pixel shuffle layer~\citep{shi2016real} with an upscale factor of $\sqrt{P}$.

\subsection{Dynamic Volume Construction}
\label{sec:method-volume-construction}

To enable multi-view interaction, we construct a 3D feature volume from multi-view feature maps, $\mathbf{M}$.
Unlike the bird's-eye view (BEV) construction in autonomous driving~\citep{li2022bevformer}, which usually relies on a fixed scene range around ego vehicle , our method dynamically adjusts the scene range based on the spatial extents of the environment to accommodate varying datasets.
Specifically, the scene's bounds are first estimated using available depth data, sparse points, or pre-defined rules.
We then partition the scene into a volume of size $X \times Y \times Z$, with voxel size dynamically adjusted to capture either fine object details or larger environments.
Voxel features, $\Tilde{\mathcal{V}}$, are initialized with learnable positional embeddings. Each voxel is projected onto the multi-view feature maps using the known transformation matrix $\mathbf{T}$. Deformable attention~\citep{zhu2021deformdetr} is then applied, where the multi-view features act as keys and values, and the voxel features as queries. Finally, a 3D convolution refines the output volume features to obtain $\mathcal{V}$.
The process can be formulated as:
\begin{equation}
\mathcal{V}=\operatorname{Conv3D}(\operatorname{DeformAttn}(\Tilde{\mathcal{V}}, \mathbf{M}, \mathbf{T})).
\end{equation}

\subsection{Differentiable Volumetric Rendering}
\label{sec:method-rendering}
After constructing the feature volume, we employ differentiable neural rendering~\citep{mildenhall2021nerf} to connect 2D and 3D domains. For better geometry representation, we utilize the implicit signed distance function (SDF) field modeling as in NeuS~\citep{wang2021neus}. The SDF represents the 3D distance from a query point to the nearest surface, implicitly capturing the 3D geometry. %

Given a feature volume $\mathcal{V}$, we apply a shallow 3D CNN $\phi$ to directly produce three outputs: an SDF feature volume $\mathcal{S} \in \mathbb{R}^{X \times Y \times Z}$, a spherical harmonic (SH)~\citep{yu2021plenoctrees,zhu2023x} coefficient field $\mathcal{K} \in \mathbb{R}^{D \times X \times Y \times Z}$ (where $D = 3 \cdot (l_\mathrm{max} + 1)^2$) for color rendering, and a semantic feature volume $\mathcal{F} \in \mathbb{R}^{C_\mathrm{semantic} \times X \times Y \times Z}$:
\begin{equation}
    \mathcal{S} \in \mathbb{R}^{X \times Y \times Z}, \quad \mathcal{K} \in \mathbb{R}^{D \times X \times Y \times Z}, \quad \mathcal{F} \in \mathbb{R}^{C_\mathrm{semantic} \times X \times Y \times Z} = \phi(\mathcal{V}).
\end{equation}
Unlike prior work~\citep{huang2023ponder,zhu2023ponderv2,yang2024unipad}, which employs an MLP to compute the attributes of each sampled point individually, we directly apply a 3D CNN to $\mathcal{V}$.
This eliminates the need for pointwise MLP computations, reducing redundant processing and enabling more efficient execution.
Consequently, our approach leads to substantial improvements in both time and memory efficiency, especially when sampling a large number of points during rendering.

To render a 2D pixel $i$, we sample $N$ ray points $\{\mathbf{p}_j = \mathbf{o} + t_j \mathbf{d}_i \mid j=1,\ldots,N, \, t_j<t_{j+1}\}$ from ray $\mathbf{r}_i$, where $\mathbf{o}$ is the camera origin and $\mathbf{d}_i$ is the viewing direction. Attributes for each point are obtained via trilinear sampling:
\begin{equation}
    s_j = \tau(\mathcal{S}, \mathbf{p}_j), \quad \mathbf{k}_j = \tau(\mathcal{K}, \mathbf{p}_j), \quad \mathbf{f}_j = \tau(\mathcal{F}, \mathbf{p}_j).
\end{equation}
The SH vector $\mathbf{k}_j = (k_l^m)_{0 \leq l \leq l_{\text{max}}, -l \leq m \leq l}$, where $k_l^m \in \mathbb{R}^3$, is used to compute view-dependent colors $\hat{\mathbf{c}}_j$ by querying the SH basis functions $Y_l^m: \mathbb{S}^2 \rightarrow \mathbb{R}$ based on the viewing direction $\mathbf{d}_j$:
\begin{equation}
    \hat{\mathbf{c}}_j = \operatorname{Sigmoid}\left( \sum_{l=0}^{l_{\text{max}}} \sum_{m=-l}^{l} k_l^m Y_l^m (\mathbf{d}_j) \right).
\end{equation}
Following the formulation in NeuS~\citep{wang2021neus}, the RGB color $\hat{\mathbf{C}}_i$, depth $\hat{\mathbf{D}}_i$, and semantic feature $\hat{\mathbf{F}}_i$ for pixel $i$ are computed by integrating the predicted values along the ray:
\begin{equation}
    \hat{\mathbf{C}}_i = \sum_{j=1}^{N} w_j \hat{\mathbf{c}}_j, \quad \hat{\mathbf{D}}_i = \sum_{j=1}^{N} w_j t_j, \quad \hat{\mathbf{F}}_i = \sum_{j=1}^{N} w_j \hat{\mathbf{f}}_j,
\end{equation}
where $w_j = T_j \alpha_j$ is the occlusion-aware weight, with $T_j = \prod_{k=1}^{j-1} (1 - \alpha_k)$ representing the accumulated transmittance and $\alpha_j$ being the opacity value. Specifically, $\alpha_j$ is computed as:
\begin{equation}
    \alpha_j = \max \left(\frac{\sigma_s(s_j) - \sigma_s(s_{j+1})}{\sigma_s(s_j)}, 0\right),
\end{equation}
where $\sigma_s(x) = (1 + e^{-sx})^{-1}$ is the sigmoid function modulated by a learnable parameter $s$.

\subsection{Loss Functions}
\label{sec:method-loss}
During pre-training, we randomly sample $K$ pixels from multi-view inputs in each iteration. The rendering loss is calculated based on the differences between the input pixel values and the predicted values. For the semantic feature map, we use the feature map from AM-RADIO~\citep{ranzinger2024radio} as supervision. Our framework has the capability to distill knowledge from multiple vision foundation models by adding multiple rendering heads. However, this paper does not explore that approach, as it is not the primary focus. The rendering loss is expressed as:
\begin{equation}
\Ls_{\text{render}} = \frac{1}{K} \sum_{i=1}^{K} \left( \lambda_{\text{color}} \cdot \| \mathbf{C}_i - \hat{\mathbf{C}}_i \| + \lambda_{\text{depth}} \cdot \| \mathbf{D}_i - \hat{\mathbf{D}}_i \| + \lambda_{\text{semantic}} \cdot \| \mathbf{F}_i - \hat{\mathbf{F}}_i \| \right).
\end{equation}
Additionally, we incorporate the Eikonal regularization loss $\Ls_{\text{eikonal}}$, near-surface SDF supervision loss $\Ls_{\text{sdf}}$, and free space SDF loss $\Ls_{\text{free}}$, which are standard in neural surface reconstruction. Detailed definitions of these losses are provided in Appendix~\ref{appx:loss}.
The total loss is defined as:
\begin{equation}
\Ls_{\text{total}} = \Ls_{\text{render}} + \lambda_{\text{eikonal}} \cdot \Ls_{\text{eikonal}} + \lambda_{\text{sdf}} \cdot \Ls_{\text{sdf}} + \lambda_{\text{free}} \cdot \Ls_{\text{free}}.
\end{equation}

\begin{figure}[!tb]
\centering
\includegraphics[width=1\linewidth]{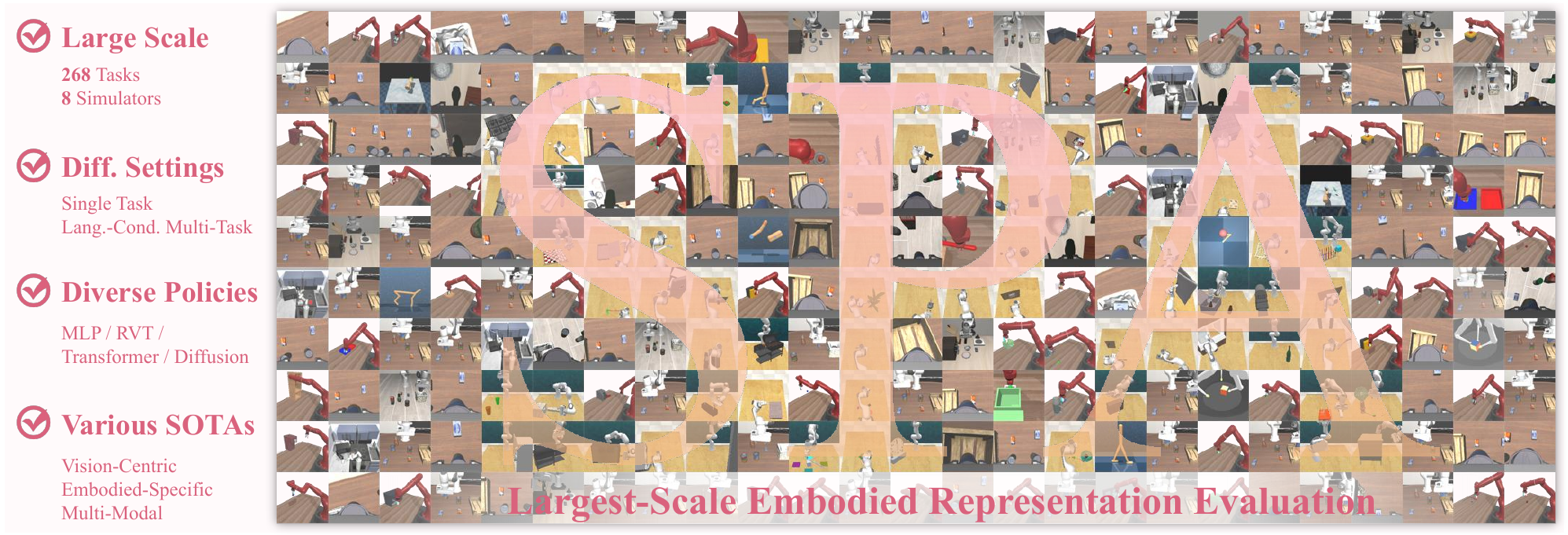}
\vspace{-1.6em}
\caption{\textbf{Overview of our large-scale embodied evaluation.} We conduct the largest-scale evaluation of embodied representation learning to date. Our study encompasses 268 tasks across 8 simulators, including both single-task and language-conditioned multi-task settings. We evaluate diverse policy architectures and assess various state-of-the-art representation methods. This thorough evaluation allows us to provide a comprehensive and unbiased analysis of different representations.}
\label{fig:evaluation}
\vspace{-1.2em}
\textbf{}\end{figure}

\section{Large-Scale Embodied Evaluation}
\label{sec:evaluation-setting}
Unlike the CV or NLP communities, where large-scale benchmarks are common, embodied representations have not been thoroughly assessed. The largest previous evaluation, VC-1~\citep{majumdar2023we}, includes only 17 tasks. This may lead to randomness and bias. Therefore, we have created \textbf{the largest embodied evaluation to date}, encompassing \textbf{268 tasks} across 8 simulators—\textbf{over 15 times larger} than VC-1's evaluation. Additionally, unlike previous approaches~\citep{majumdar2023we,nair2022r3m,radosavovic2023real} that used a small MLP policy under single-task settings, our evaluation spans multiple policy types (\eg MLP, diffusion, transformer) and includes both single-task and language-conditioned multi-task settings. This unprecedented scale and diversity ensure robust and convincing conclusions. During all evaluations, we adhere to standard practices by freezing the pre-trained representation model.
Our detailed evaluation settings can be found in Appendix~\ref{appx:eval-setups}. The overview of our evaluation is shown in \figref{fig:evaluation}.

We have included 3 \textit{single-task benchmarks}:\\
\noindent \textbf{1) VC-1~\citep{majumdar2023we}} involves 4 selected simulators with 14 tasks in total: Adroit (AD)~\citep{Kumar2016thesis}, Meta-World (MW)~\citep{yu2020meta}, DMControl (DMC)~\citep{tunyasuvunakool2020dm_control}, and TriFinger (TF)~\citep{wuthrich2020trifinger}. We use a 3-layer MLP as the policy network.\\ %
\noindent \textbf{2) Franka Kitchen~\citep{gupta2019relay}} 
involves 5 selected tasks. Each task spans two camera viewpoints and three random seeds. We utilize 25 demonstrations to train a 2-layer MLP policy.\\
\noindent \textbf{3) Meta-World~\citep{yu2020meta}} involves 48 selected tasks of varying difficulty. We implemented the Diffusion Policy~\citep{chi2023diffusion} on this benchmark and adhered to the setup in \cite{ze20243d} to generate 10 demonstrations for each single-task training, followed by evaluation through 20 rollouts.

We have also included 2 \textit{language-conditioned multi-task benchmarks}:\\
\noindent \textbf{1) RLBench~\citep{james2020rlbench}} features 71 selected tasks that can be successfully executed. We divide the tasks into two groups according to their category defined by PolarNet~\citep{chen23polarnet}. We employ RVT-2~\citep{goyal2024rvt2}, the SOTA method on this benchmark, as our policy.\\
\noindent \textbf{2) LIBERO~\citep{liu2024libero}} comprises 130 tasks across 5 suites: LIBERO-Spatial, LIBERO-Object, LIBERO-Goal, LIBERO-10, and LIBERO-90. We train a language-conditioned transformer policy provided by the original LIBERO on each suite with only 20 demonstrations per task.

\section{Training and Implementation Details}

\label{sec:traning-implementations}
In this section, we present the implementation and training of our SPA model. 
We first compile several multi-view datasets, training ViT-B models on each to assess the impact of different datasets (\secref{sec:dataset-investigation}). Finally, we integrate all factors and scale up both data and model size to train the strongest version of SPA using a ViT-large (ViT-L) backbone (\secref{sec:put-all-together}). More details can be found in Appendix~\ref{appx:impl-details}.

\subsection{Dataset Investigation}
\label{sec:dataset-investigation}

We collect several multi-view datasets. To investigate their effectiveness in SPA representation learning, we train a ViT-B model on one or two of the datasets, keeping the total training steps constant, and assess performance on the VC-1 benchmarks. For simplicity, semantic rendering is disabled. The datasets investigated are listed in the first column of \tabref{tab:dataset-ablation}.
Most datasets provide ground-truth depth, which we use for supervision. 
As our findings above reveal that depth supervision is helpful, for datasets lacking ground-truth depth, we employ a depth estimation model. For instance, Droid~\citep{khazatsky2024droid} only offers binocular images, so we apply CroCo-Stereo~\citep{croco_v2} for dense depth estimation. 
Additionally, due to inaccurate camera poses in Droid, we treat its data as single-view inputs. 
The results are presented in \tabref{tab:dataset-ablation}, with further details in Appendix~\ref{appx:impl-details}. Our analysis reveals that some datasets can be detrimental. For example, although RH20T~\citep{fang2023rh20t} is a large-scale robotic dataset, its lack of visual diversity—stemming from data collected in the same lab—negatively impacts representation learning.

\begin{table}[!hb]\centering
\caption{\textbf{Influence of different datasets.} We present the performance results on the VC-1 benchmark. \textit{Mean S.R.} refers to the mean success rate across all individual tasks.}\label{tab:dataset-ablation}
\vspace{-1.0em}
\resizebox{\linewidth}{!}{
\tablestyle{8pt}{0.95}
\begin{tabular}{l|cccc|c}\toprule
Datasets &AD &MW &DMC &TF &\makecell{Mean\\S.R.}\\\cmidrule{1-6}
ScanNet~\citep{dai2017scannet} &52.67$\pm$4.11 &90.93$\pm$3.22 &65.11$\pm$1.31 &70.75$\pm$1.08 & 73.68 \\
ScanNet++~\citep{yeshwanth2023scannet++} &56.00$\pm$2.83 &89.87$\pm$4.20 &62.24$\pm$4.51 &71.28$\pm$0.38 & 72.51\\
Arkitscenes~\citep{baruch2021arkitscenes} &50.67$\pm$5.73 &89.87$\pm$4.59 &60.51$\pm$2.55 &66.54$\pm$0.13 & 70.45\\
Droid~\citep{khazatsky2024droid} &53.33$\pm$5.25 &90.40$\pm$4.90 &60.99$\pm$3.72 &73.28$\pm$0.61 & 72.16\\
Hypersim~\citep{roberts2021hypersim} &52.67$\pm$4.11 &88.80$\pm$3.27 &60.84$\pm$2.06 &72.29$\pm$0.47 &71.29\\
Hypersim + ADT~\citep{pan2023aria} &52.00$\pm$2.83 &87.20$\pm$2.30 &63.61$\pm$1.04 &70.83$\pm$0.13 &71.41\\
Hypersim + S3DIS~\citep{armeni2017joint} &49.33$\pm$0.94 &94.13$\pm$2.04 &64.57$\pm$3.91 &71.74$\pm$0.75 &73.98\\
Hypersim + Structured3D~\citep{zheng2020structured3d} &46.67$\pm$4.11 &80.27$\pm$7.72 &58.02$\pm$ 2.34 &65.05$\pm$0.40 &65.35\\
Hypersim + RH20T~\citep{fang2023rh20t} &47.33$\pm$1.89 &86.93$\pm$4.99 &57.01$\pm$4.35 &64.28$\pm$0.46 &67.35\\
Hypersim + ASE~\citep{avetisyan2024scenescript} &47.33$\pm$4.11 &87.73$\pm$3.39 &60.62$\pm$4.14 &68.59$\pm$0.30 &69.54\\
\bottomrule
\end{tabular}
}
\vspace{-1.0em}

\end{table}

\subsection{Put All Together}
\label{sec:put-all-together}

Based on the previous analyses, we proceed to pre-train the final version of SPA. We use a mask ratio of $0.5$ and enable all three rendering losses. Following Ponder~\citep{huang2023ponder}, we set the weight for the RGB loss to $10$, the weights for the depth and semantic losses to $1$, and use $\lambda_{\mathrm{eikonal}} = 0.01$, $\lambda_{\mathrm{sdf}} = 10$, and $\lambda_{\mathrm{free}} = 1$.
The volume size is $128 \times 128 \times 32$. For stable training, we apply the Exponential Moving Average (EMA) technique with a decay of $0.999$. We use AdamW~\citep{loshchilov2017fixing} as the optimizer with a weight decay of $0.04$ and a learning rate of $8e^{-4}$. OneCycle~\citep{smith2019super} learning rate scheduler is adopted. We utilize 80 NVIDIA A100-SXM4-80GB GPUs, each with a batch size of $2$, and accumulate gradients over 8 batches, resulting in a total effective batch size of $2 \times 8 \times 80 = 1280$. Training is conducted over $2000$ epochs, sampling each dataset to match the size of ADT per epoch. The datasets used for the final version include ScanNet, ScanNet++, ADT, S3DIS, Hypersim, and Droid.

\section{Experiment Results}
\label{sec:experiment-results}

In this section, we present the results of our large-scale evaluation. Our experiments are designed to address the following research questions:\\
\noindent \textbf{Q1:} How does SPA compare to other methods in our large-scale embodied evaluation?\\
\noindent \textbf{Q2:} What insights do we gain about various representation learning approaches from our evaluation?\\
\noindent \textbf{Q3:} Does SPA really learn enhanced 3D awareness that results in improved embodied representation?\\
\noindent \textbf{Q4:} Can SPA facilitate robot learning in real-world environments in a zero-shot manner?
\subsection{Overall Comparisons (Q1, Q2)}
\label{sec:overall-comparisons}

\begin{table}[!tp]\centering
\vspace{-0.25em}
\caption{\textbf{Summary of different representation learning methods.} `\#Param.' is the total parameters of the encoder, while `\#Frames' indicates the total number of image frames used during pre-training.}\label{tab:dataset-summary}
\vspace{-1.0em}
\resizebox{\linewidth}{!}{
\tablestyle{4pt}{0.9}
\begin{tabular}{l|ccc|cccc|ccc|c}\toprule[0.14em]
\multirow{3}{*}{Method} &\multicolumn{3}{c|}{\textit{Vision-Centric}} &\multicolumn{4}{c|}{\textit{Multi-Modal}} &\multicolumn{3}{c|}{\textit{Embodied-Specific}} &\multirow{2}{*}{\makecell{\textit{Distilled}\\AM-RADIO\\\citesplit{ranzinger2024radio}}} \\
 &\makecell{MoCoV3\\\citesplit{chen2020improved}} 
 &\makecell{MAE\\\citesplit{he2022masked}} &\makecell{DINOV2\\\citesplit{oquab2023dinov2}} &\makecell{CLIP\\\citesplit{radford2021learning}} &\makecell{EVA\\\citesplit{fang2023eva}}
 &\makecell{InternViT-300M\\\citesplit{chen2024internvl}} &\makecell{InternViT-6B\\\citesplit{chen2024internvl}} 
&\makecell{MVP\\\citesplit{radosavovic2023real}} &\makecell{VC-1\\\citesplit{majumdar2023we}} &\makecell{SPA\\\vspace{-0.3em}(Ours)\\\;} & \\
Is Vanilla? &\cmark &\cmark &\xmark &\cmark &\cmark &\xmark &\xmark &\cmark &\cmark &\cmark &\xmark \\
Input Size &224 &224 &224 &224 &224 &448 &224 &256 &224 &224 &dynamic \\
Patch Size &16 &16 &14 &14 &14 &14 &14 &16 &16 &16 &16 \\
\#Param. &303M &303M &303M &303M &303M &303M &5.9B &303M &303M &303M &653M \\
\#Frames &1.28M &1.28M &1.2B &400M &14M &5.0B &5.0B &4.5M &5.6M &3.8M &1.4B \\
\bottomrule[0.14em]
\end{tabular}
}
\vspace{-1.8em}
\end{table}

\begin{table}[!tb]\centering
\vspace{-0.05em}
\caption{\textbf{Comparison of different representation learning methods.} `OOM' indicates an out-of-memory error during evaluation. The best and second-best results are \textbf{bolded} and \underline{underlined} respectively. The number in parentheses denotes the number of tasks. \rebuttal{S.R. denotes `Success Rate'.}}\label{tab:main-comparison}
\vspace{-1.0em}
\resizebox{0.95\linewidth}{!}{
\tablestyle{2pt}{0.95}
\begin{tabular}{l|l|ccc|cccc|cccc}\toprule[0.14em]
\multicolumn{2}{r|}{Method} &\multicolumn{3}{c|}{\textit{Vision-Centric}} &\multicolumn{4}{c|}{\textit{Multi-Modal}} &\multicolumn{3}{c}{\textit{Embodied-Specific}} \\
\multicolumn{2}{l|}{Benchmark} &MoCoV3 &MAE &DINOV2 &CLIP &EVA &\makecell{InternViT-\\300M} &\makecell{InternViT-\\6B} &MVP &VC-1 &SPA (Ours) \\\midrule
\multirow{4}{*}{\rotatebox{0}{VC-1}} &AD (2) &58.7$\pm$7.0 &58.0$\pm$2.0 &47.3$\pm$3.1 &48.7$\pm$3.1 &58.0$\pm$6.0 &53.3$\pm$3.1 &\underline{60.0$\pm$9.2} &53.3$\pm$4.2 &54.0$\pm$4.0 &\textbf{60.0$\pm$4.0} \\
&MW (5) &88.8$\pm$5.0 &90.0$\pm$4.6 &84.0$\pm$3.7 &77.1$\pm$3.2 &90.7$\pm$0.9 &84.0$\pm$3.7 &89.1$\pm$1.2 &\textbf{93.6$\pm$5.2} &87.5$\pm$3.8 &\underline{93.3$\pm$2.0} \\
&DMC (5) &67.3$\pm$3.3 &\textbf{74.4$\pm$1.8} &64.5$\pm$2.5 &53.9$\pm$3.6 &62.7$\pm$2.8 &53.3$\pm$0.4 &66.3$\pm$3.2 &69.4$\pm$2.6 &65.3$\pm$3.6 &\underline{71.1$\pm$5.0}\\
&TF (2) &67.9$\pm$0.2 &73.0$\pm$0.5 &68.5$\pm$0.4 &56.1$\pm$1.6 &67.2$\pm$0.2 &65.2$\pm$1.6 &70.7$\pm$0.9 &\underline{73.2$\pm$0.8} &70.9$\pm$1.1 &\textbf{73.6$\pm$2.0} \\\cmidrule{1-12}
\multirow{2}{*}{\rotatebox{0}{RLBench}} &Group 1 (35) &73.7 &78.3 &78.2 &76.8 &75.2 &74.1 &OOM &76.2 &\underline{80.1} &\textbf{80.5} \\
&Group 2 (36) &54.2 &\underline{57.7} &56.1 &55.7 &57.0 &54.9 &OOM &56.3 &55.7 &\textbf{61.2} \\\cmidrule{1-12}
\multicolumn{2}{c|}{Meta-World (48)} &\textbf{69.3$\pm$1.5} &67.8$\pm$1.7 &56.3$\pm$0.6 &66.7$\pm$1.7 &63.7$\pm$1.3 &57.5$\pm$1.7 &OOM &66.4$\pm$1.7 &68.6$\pm$1.5 &\underline{69.2$\pm$1.}7 \\\cmidrule{1-12}
\multirow{5}{*}{\rotatebox{0}{LIBERO}} &Object (10) &65.3$\pm$8.0 &71.7$\pm$13.1 &64.7$\pm$9.9 &50.2$\pm$7.0 &\underline{73.2$\pm$6.0} &67.7$\pm$6.0 &58.0$\pm$10.6 &63.7$\pm$4.8 &69.7$\pm$7.2 &\textbf{76.7$\pm$5.3} \\
&Spatial (10) &40.5$\pm$0.9 &57.2$\pm$2.9 &36.3$\pm$11.8 &32.2$\pm$0.6 &\textbf{59.3$\pm$7.7} &48.3$\pm$6.4 &42.0$\pm$10.3 &\underline{58.0$\pm$6.2} &50.5$\pm$7.5 &50.0$\pm$3.8 \\
&Goal (10) &49.2$\pm$8.1 &54.3$\pm$6.0 &22.2$\pm$2.3 &30.3$\pm$3.2 &56.8$\pm$2.9 &58.8$\pm$4.5 &33.2$\pm$2.0 &\underline{63.8$\pm$2.8} &57.5$\pm$6.6 &\textbf{65.3$\pm$2.5} \\
&10 (10) &34.2$\pm$3.8 &\textbf{41.2$\pm$4.5} &28.3$\pm$3.0 &27.5$\pm$3.9 &43.3$\pm$2.8 &38.2$\pm$1.3 &34.3$\pm$4.6 &39.0$\pm$0.9 &39.7$\pm$3.5 &\underline{40.2$\pm$3.6} \\
&90 (90) &30.0$\pm$1.4 &29.9$\pm$2.0 &27.5$\pm$2.2 &29.4$\pm$2.0 &31.3$\pm$2.3 &23.8$\pm$1.8 &27.1$\pm$2.1 &\underline{32.1$\pm$3.5} &30.6$\pm$3.3 &\textbf{32.2$\pm$1.6} \\\cmidrule{1-12}
\multicolumn{2}{c|}{Franka-Kitchen (5)} &\textbf{48.3$\pm$4.7} &\underline{42.7$\pm$2.6} &40.9$\pm$6.4 &30.8$\pm$3.3 &37.3$\pm$1.3 &28.5$\pm$1.7 &OOM &34.3$\pm$6.1 &37.5$\pm$3.5 &40.6$\pm$1.9 \\\midrule
\multicolumn{2}{c|}{Mean S.R.~$\uparrow$} &81.67 &\underline{85.13} &75.18 &77.10 &83.84 &75.41 &30.65 &84.85 &84.69 &\textbf{88.63} \\
\multicolumn{2}{c|}{Mean Rank~$\downarrow$} &4.51 &\underline{4.07} &5.61 &5.17 &4.37 &5.92 &7.57 &4.24 &4.13 &\textbf{3.20} \\
\bottomrule[0.14em]
\end{tabular}
}
\vspace{-2.25em}
\end{table}

\noindent \textbf{Evaluation Metrics.} We follow prior work~\citep{majumdar2023we,zhu2024point} in reporting two metrics: \textit{Mean Success Rate (Mean S.R.)} and \textit{Mean Rank}. Mean S.R. is the average success rate across all tasks, indicating overall performance, while Mean Rank reflects the average ranking of each method's success rate across tasks, providing a measure of relative performance. Since RLBench has fixed train and test sets, we report a single result for this benchmark. %

\renewcommand{\baselinestretch}{0.98}

\noindent \textbf{Baselines.} We evaluate 9 SOTA representation learning models, all using ViT-L backbone, categorized into vision-centric, multi-modal, and embodied-specific. This also includes a 6B multi-modal model~\citep{chen2024internvl}. 
\rebuttal{
The vision-centric methods are originally from the vision community; the multi-modal methods are typically CLIP-style language-image pre-trained models and are used specifically for VLMs; the embodied-specific methods are designed and pre-trained specifically for embodied AI tasks.
}
Details are summarized in \tabref{tab:dataset-summary}. 
The results on each benchmark are shown in \tabref{tab:main-comparison}. For detailed results on each task and each random seed, please refer to Appendix~\ref{appx:detailed-results}. We also have visualized the performance radar chart and the per-task rank distributions in \figref{fig:radar_plot}.

\noindent \textcolor{red}{\textbf{\textit{Finding 1:}}} We observe that SPA demonstrates superior performance in both mean success rate and mean rank. While no method ranks first across all individual benchmarks, consistent with the findings by \cite{majumdar2023we}, SPA achieves the best or second-best mean success rate \textbf{in 11 out of 13 benchmarks}. Additionally, it ranks in the top 3 for \textbf{over 65.5\% of individual tasks}, surpassing the second and third highest percentages of 46.8\% for MAE and 46.0\% for VC-1, respectively. These trends demonstrate the robustness and superiority of SPA.

\noindent \textcolor{red}{\textbf{\textit{Finding 2:}}} We observe that for vision-centric methods, superior performance on vision tasks does \textbf{not} necessarily translate to better embodied performance. Despite using 10 times more data, DINOV2 performs worse than MoCoV3 and MAE. Notably, MAE performs exceptionally well, likely due to its reconstruction objective, which enhances \textit{2D spatial awareness}. Interestingly, methods like MVP and VC-1, which are MAE models pre-trained on human interaction data, show \textbf{no clear advantage} over ImageNet~\citep{deng2009imagenet} pre-trained MAE. This suggests that while human activity data may seem more relevant, data diversity and thorough convergence are more critical.

\noindent \textcolor{red}{\textbf{\textit{Finding 3:}}} Multimodal methods \textbf{generally perform poorly} in embodied evaluations, except EVA, which combines image-language contrastive techniques with MAE reconstruction. Furthermore, InternViT-6B, despite having significantly more model parameters, does not demonstrate superiority and even performs worse on some benchmarks compared to InternViT-300M. This indicates that current scaling properties of multimodal approaches do not effectively translate to embodied AI.

\noindent \textcolor{red}{\textbf{\textit{Finding 4:}}} Focusing on a single benchmark can \textbf{lead to highly biased conclusions}. For instance, ImageNet pre-trained methods (\eg MoCoV3 and MAE) perform exceptionally well on the Franka Kitchen benchmark, suggesting a minimal domain gap between ImageNet and Franka Kitchen observations. Moreover, despite being based on MAE, previous SOTA embodied representations like MVP and VC-1 do not consistently outperform the original ImageNet version. These observations underscore the importance of our large-scale embodied evaluation. %

\begin{table}[!tb]\centering
\vspace{-0.25em}
\caption{\textbf{Additional comparisons of ViT-base models.} \rebuttal{S.R. denotes `Success Rate'.}}\label{tab:vitb-compare}
\vspace{-1.0em}
\resizebox{\linewidth}{!}{
\tablestyle{3pt}{0.9}
\begin{tabular}{l|l|ccccccccc}\toprule
\multicolumn{2}{l|}{Methods}
&\makecell{DINOV2-B\\\citesplit{oquab2023dinov2}} 
&\makecell{MAE-B\\\citesplit{he2022masked}} &\makecell{R3M-B\\\citesplit{nair2022r3m}} &\makecell{VC-1-B\\\citesplit{majumdar2023we}} &\makecell{STP-B\\\citesplit{yang2024spatiotemporal}} &\makecell{Voltron-B\\\citesplit{karamcheti2023voltron}} &\makecell{Theia-B\\\citesplit{shang2024theia}} &\makecell{SPA-B\\\vspace{-0.3em}(Ours)\\\;} \\\midrule
\multicolumn{2}{l|}{Is Vanilla?} &\xmark &\cmark &\cmark &\cmark &\cmark &\xmark &\cmark &\cmark \\
\multicolumn{2}{l|}{Embodied?} &\xmark &\xmark &\cmark &\cmark &\cmark &\cmark &\cmark &\cmark \\\midrule
\multirow{4}{*}{VC-1} &AD &36.67$\pm$2.31 &52.67$\pm$3.06 & 48.00$\pm$6.93 &50.00$\pm$5.29 &52.00$\pm$2.00 & 46.67$\pm$4.62 &53.33$\pm$5.03 &52.00$\pm$3.46 \\
&MW &60.80$\pm$0.80 &88.80$\pm$4.00 & 59.20$\pm$5.60 &86.67$\pm$0.92 &92.00$\pm$1.39 &84.00$\pm$3.20 &89.07$\pm$3.23 &92.00$\pm$4.16 \\
&DMC &35.19$\pm$4.87 &62.39$\pm$4.97 & 49.57$\pm$4.85 &60.92$\pm$0.70 &61.40$\pm$2.86 &56.36$\pm$2.01 &64.98$\pm$3.42 &64.21$\pm$3.52 \\
&TF &54.50$\pm$1.16 &70.78$\pm$0.17 & 56.18$\pm$7.00 &72.33$\pm$0.69 &67.96$\pm$0.95 &74.26$\pm$1.57 &69.41$\pm$0.60 &73.06$\pm$0.51 \\\midrule
\multicolumn{2}{l|}{Mean S.R.} &47.31 &71.63 & 54.37 &70.19 &71.92 &69.50 &\underline{72.55} &\textbf{73.66} \\
\bottomrule
\end{tabular}
}
\vspace{-1.4em}
\end{table}

\subsection{Additional Comparisons (Q1)}
\label{sec:additional-comparisons}
We primarily compare with SOTA methods using the ViT-L backbone, which is commonly available and pre-trained on large-scale datasets. However, some embodied-specific models are only offered in ViT-B variants. Therefore, we provide additional comparisons with several ViT-B models in \tabref{tab:vitb-compare}. 
Our ViT-B version, SPA-B, also outperforms other baselines. Furthermore, when compared to SPA-L on VC-1 benchmarks, the mean success rate increases by 4.16 (73.66 $\rightarrow$ 77.82). This indicates that increasing the model size positively impacts SPA's performance.

\subsection{Study on 3D Awareness of SPA (Q3)}
\label{sec:study-on-3d-awareness}
\begin{table}[!tb]\centering
\vspace{0.25em}
\caption{\textbf{Zero-shot camera pose estimation.} \rebuttal{Trans. and Rot. denote `translation' and `rotation' errors respectively. The detailed metrics on the error calculation are listed in Appendix~\ref{appx:camera-pose}.}}\label{tab:camera-pose}
\vspace{-1.0em}
\resizebox{\linewidth}{!}{
\tablestyle{1.5pt}{0.9}
\begin{tabular}{l|ccccccccccc}\toprule
Error &MoCoV3 &MAE &DINOV2 &CLIP &EVA &\small{InternViT-300M} &InternViT-6B &MVP &VC-1 &SPA(Ours) \\\cmidrule{1-11}
Trans. ($\times e^{-2}$) &2.29$\pm$0.07 &2.15$\pm$0.07 &6.55$\pm$0.07 &4.21$\pm$0.37 &5.49$\pm$0.24 &4.62$\pm$0.14 &5.39$\pm$0.41 &2.15$\pm$0.12 &2.02$\pm$0.07 &\textbf{1.65$\pm$0.09} \\
Rot. ($\times e^{-1}$) &0.79$\pm$0.07 &0.73$\pm$0.03 &2.12$\pm$0.25 &1.52$\pm$0.08 &1.83$\pm$0.09 &1.83$\pm$0.08 &1.91$\pm$0.12 &0.77$\pm$0.05 &0.72$\pm$0.01 &\textbf{0.61$\pm$0.01} \\
\bottomrule
\end{tabular}
}
\end{table}

\begin{wraptable}[5]{r}{0.48\textwidth}
\centering
\vspace{-3.35em}
\caption{Additional ablations on VC-1.}\label{tab:spa-mae-radio}
\label{tab:spamae_radio}
\vspace{-1.0em}
\resizebox{\linewidth}{!}{
\tablestyle{2.5pt}{1.05}
\begin{tabular}{l|l|ccccc}\toprule
\multicolumn{2}{c|}{Methods} &SPA-B &SPA-MAE &RADIO &E-RADIO \\\midrule
\multirow{4}{*}{VC-1} &AD &52.00$\pm$3.46 &55.33$\pm$3.06 &55.33$\pm$3.06 &56.67$\pm$2.31 \\
&MW &92.00$\pm$4.16 &90.67$\pm$6.00 &72.00$\pm$9.23 &83.47$\pm$4.11 \\
&DMC &64.21$\pm$3.52 &63.85$\pm$3.60 &67.38$\pm$7.35 &62.92$\pm$4.24 \\
&TF &73.06$\pm$0.51 &70.14$\pm$0.98 &71.75$\pm$0.14 &68.44$\pm$1.19 \\\midrule
\multicolumn{2}{c|}{Mean S. R.} &\textbf{73.66} &73.11 &67.93 &70.16 \\
\bottomrule
\end{tabular}
}
\vspace{-1.2em}
\end{wraptable}

Firstly, we aim to provide clear evidence that the performance improvements of SPA are due to its 3D awareness. To demonstrate this, we conducted two additional ablation studies on the VC-1 benchmarks: 1) To determine whether the performance gain is due to SPA's pre-training objectives or the datasets used, we continue pre-training the ImageNet pre-trained MAE-B (the most competitive method besides SPA) on the same datasets used by SPA-B, referring to this model as SPA-MAE. Hyperparameters, including mask ratio and batch size, are kept at their default settings, and both the ImageNet pre-trained encoder and decoder weights are initially loaded. 2) Since SPA uses the feature map of RADIO for semantic rendering supervision, we also evaluate the original RADIO (653M parameters) and its efficient version, E-RADIO (391M parameters). Results are presented in \tabref{tab:spa-mae-radio}.
\begin{figure}[!tb]
\centering
\includegraphics[width=0.9\linewidth]{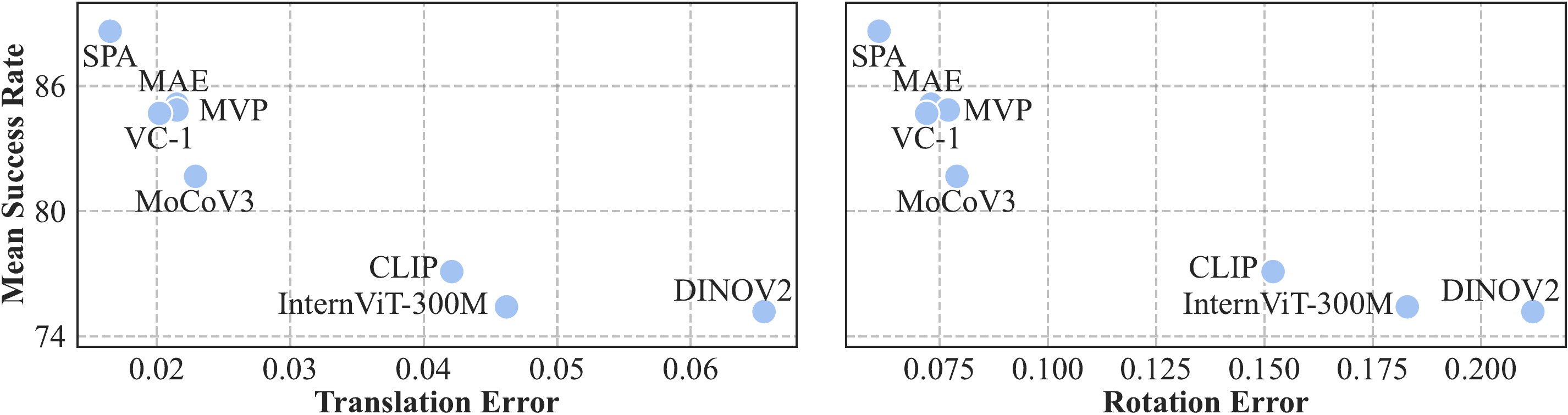}
\caption{\textbf{Correlation between mean success rate and camera pose regression error.}}
\label{fig:regression}
\vspace{-2.18em}
\end{figure}

\noindent \textcolor{red}{\textbf{\textit{Finding 5:}}} The 3D-aware pre-training objective significantly enhances SPA's performance. It surpasses the single-image naive MAE with the same data. %
Notably, SPA learns superior representations compared to its semantic rendering teacher by a substantial margin.

\begin{wrapfigure}[26]{r}{0.425\columnwidth}
    \centering
    \vspace{-0.45em}
    \includegraphics[width=0.425\columnwidth]{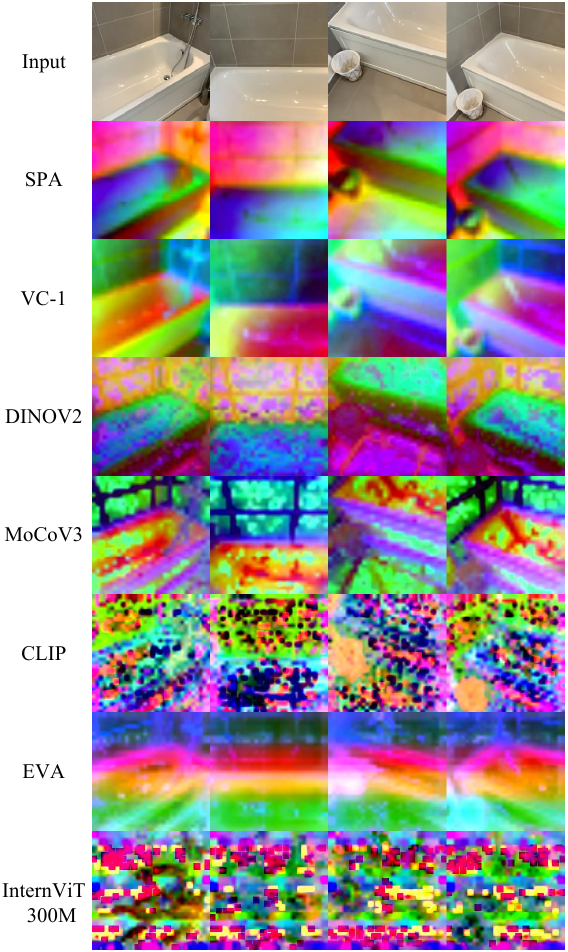}
    \vspace{-2em}
    \caption{\textbf{Feature map visualization.}}
    \label{fig:vis}
    \vspace{-1.2em}
\end{wrapfigure}

Moreover, we provide both quantitative and qualitative evidence to demonstrate that SPA has acquired 3D awareness. For qualitative analysis, we visualize the zero-shot feature maps on multiview images of different encoder outputs, as shown in \figref{fig:vis}. The images are taken from the unseen Arkitscenes dataset. For quantitative analysis, we evaluate the zero-shot 3D awareness of various methods using a camera pose estimation task on the NAVI dataset~\citep{jampani2023navi}.
Specifically, given a pair of images from different viewpoints, we use a frozen encoder to extract features and concatenate them. 
A small MLP then regresses the relative camera pose and we report rotation and translation errors in \tabref{tab:camera-pose}. Details are in Appendix~\ref{appx:camera-pose}. 
While \cite{el2024probing} has explored 3D awareness of different vision models, their context differs. Their tasks can allow strong semantic models like DINOV2 to `cheat'. For example, multiview correspondence can be achieved through semantic matching, and the relative depth estimation task involves transforming normalized values into discrete bins, resembling a per-pixel classification task. Additionally, they emphasize fine-grained dense local context, whereas, embodied AI focuses more on sparse, global information~\citep{nair2022r3m}. Thus, we believe camera pose estimation, which predicts a global `pose' from observations, is more relevant to embodied AI, where a policy must predict a global `action'.

\noindent \textcolor{red}{\textbf{\textit{Finding 6:}}} We observe that SPA %
outperforms all other methods in zero-shot camera pose estimation. It achieved \textbf{an 18.3\% improvement in translation and a 15.3\% reduction in rotation error} compared to the second-best model. Additionally, we identify a \textbf{clear positive correlation} between camera pose estimation and embodied evaluation performance, as demonstrated in \figref{fig:regression}. This finding supports our spatial hypothesis and may offer valuable insights for future research on embodied representation.

\noindent \textcolor{red}{\textbf{\textit{Finding 7:}}} The feature map visualization provides clear evidence that SPA has learned multi-view consistent knowledge, demonstrating its 3D awareness. Additionally, the features produced by SPA are \textbf{cleaner and more coherent}. Though VC-1 also generates smooth features, they are \textit{not consistent across viewpoints}. The feature maps from the multi-modal approach are highly noisy and lack details.

\subsection{\rebuttal{Hyperparameter Investigation}}

\begin{wraptable}[8]{r}{0.45\textwidth}
\centering
\vspace{-3.75em}
\caption{\textbf{Mask ratio and loss components.} C., D., S. denote color, depth, and semantic. %
}
\label{tab:mr_loss}
\vspace{-0.8em}
\resizebox{\linewidth}{!}{
\tablestyle{1.5pt}{1.05}
\begin{tabular}{c|ccc|cccc|c}\toprule
\multirow{2}{*}{\makecell{Mask\\Ratio}} &\multicolumn{3}{c|}{Loss} &\multicolumn{4}{c|}{VC-1 Benchmark} &\multirow{2}{*}{\makecell{Mean\\S.R.}} \\
&C. &D. &S. &AD &MW &DMC &TF & \\\midrule
0.00 &\cmark &\cmark &\cmark &53.3$\pm$4.6 &88.5$\pm$5.7 &57.5$\pm$2.6 &74.1$\pm$0.6 &70.36 \\
0.25 &\cmark &\cmark &\cmark &52.7$\pm$3.1 &89.6$\pm$4.5 &57.6$\pm$3.0 &70.4$\pm$1.7 &70.17 \\
0.50 &\cmark &\cmark &\cmark &53.3$\pm$4.2 &88.8$\pm$1.6 &60.1$\pm$3.1 &72.6$\pm$0.7 &\textbf{71.18} \\
0.75 &\cmark &\cmark &\cmark &51.3$\pm$1.2 &88.0$\pm$3.5 &61.1$\pm$3.5 &73.0$\pm$0.8 &71.01 \\
0.95 &\cmark &\cmark &\cmark &51.3$\pm$1.2 &85.6$\pm$4.0 &62.5$\pm$5.3 &73.1$\pm$0.2 &70.67 \\
0.50 &\cmark &\xmark &\cmark &51.3$\pm$1.2 &90.9$\pm$3.3 &58.8$\pm$5.6 &71.5$\pm$1.0 &71.01 \\
0.50 &\xmark &\cmark &\cmark &52.0$\pm$2.0 &89.3$\pm$3.3 &53.9$\pm$4.3 &70.9$\pm$1.3 &68.71 \\
0.50 &\cmark &\cmark &\xmark &52.7$\pm$3.1 &88.0$\pm$4.5 &61.5$\pm$3.4 &71.6$\pm$1.2 &71.16 \\
\bottomrule
\end{tabular}
}
\vspace{-1.1em}
\end{wraptable}

\label{sec:hyperparameter-investigation}
We conduct hyperparameter tuning with a ViT-B model on ScanNet~\citep{dai2017scannet}, and evaluate it on VC-1 benchmarks, as shown in \tabref{tab:mr_loss}. 
{\noindent \textbf{1) Mask Ratio.}} Our results indicate that a mask ratio of $0.5$ is the most effective.
{\noindent \textbf{2) Loss Components.}} As discussed in \secref{sec:method-loss}, our rendering loss consists of color, depth, and semantic components. We sequentially deactivate each and find that all three are valuable. However, deactivating the semantic loss has the least impact.

\subsection{Real-World Experiments (Q4)}
\label{sec:real-world}
\begin{wrapfigure}[9]{r}{0.42\columnwidth}
    \centering
    \vspace{-3.7em}
    \includegraphics[width=0.42\columnwidth]{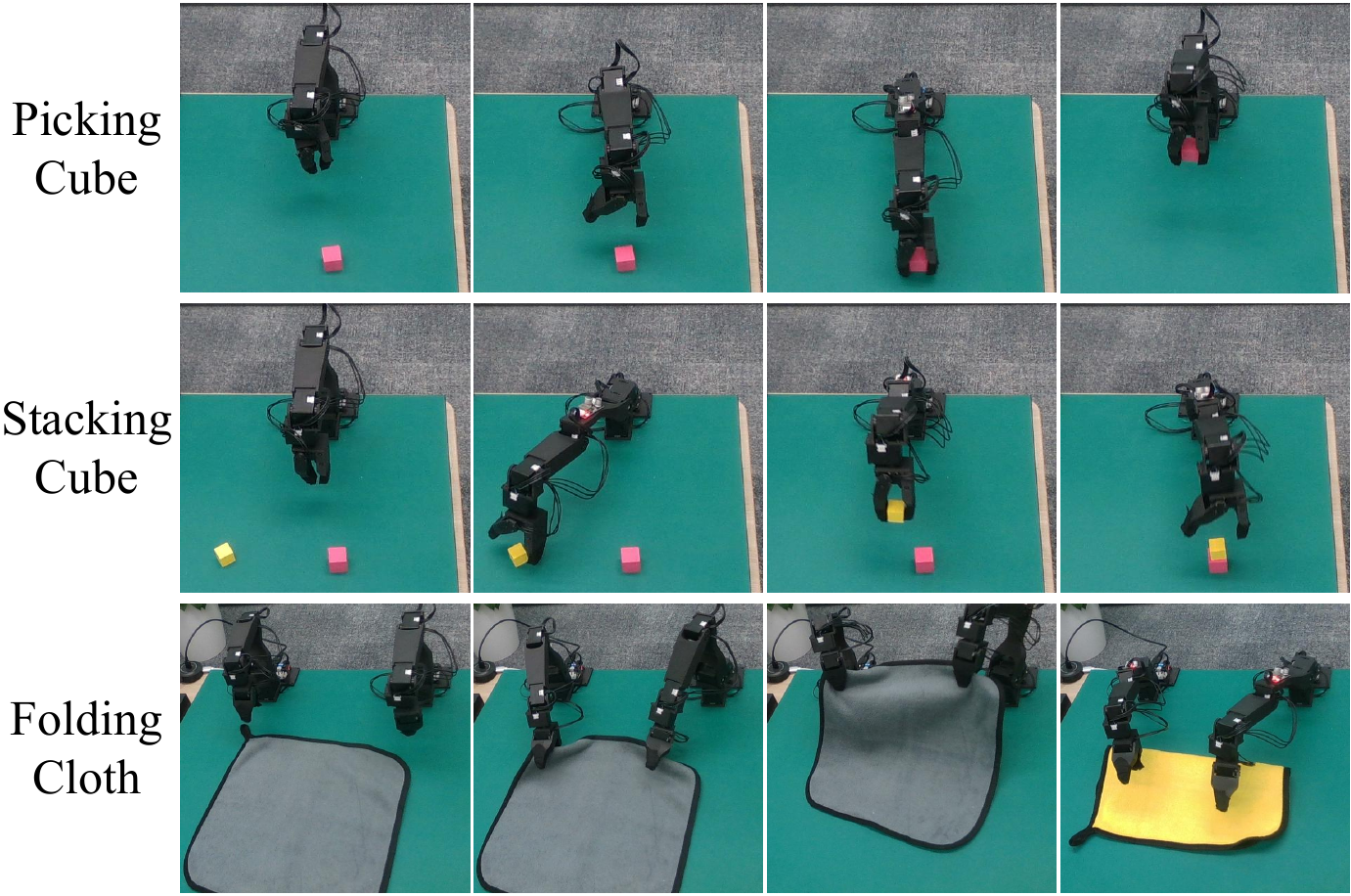}
    \vspace{-2em}
    \caption{\textbf{Real-world task illustrations.}}
    \label{fig:real-world}
    \vspace{-1.5em}
\end{wrapfigure}

We conduct several real-world experiments to further investigate the generalization ability of different representations. Specifically, we utilize the open-sourced Low-Cost Robot Arm~\citep{Alexander} to learn real-world tasks from pixels, with only 50 demonstrations per task using different frozen pre-trained representations. The robot performed two single-arm tasks: (1) picking a cube, and (2) stacking a yellow cube on a pink cube, as well as one dual-arm task: folding a cloth in half. Refer to \figref{fig:real-world} for illustrations and Appendix~\ref{appx:real-world} for more details. We evaluate each task with 25 rollouts, with the results presented in \tabref{tab:real-world}.
SPA consistently performs better on real-world tasks, %
suggesting that SPA's pre-trained representations can robustly adapt to real-world environments without finetuning.

\begin{table}[!tbp]\centering
\caption{\textbf{Real-world experiment results.} \rebuttal{S.R. denotes `Success Rate'.}}\label{tab:real-world}
\vspace{-1.2em}
\resizebox{0.9\linewidth}{!}{
\tablestyle{3pt}{1.05}
\begin{tabular}{l|ccccccccccc}\toprule
Methods&MoCoV3 &MAE &DINOV2 &CLIP &EVA &InternViT-300M &InternViT-6B &MVP &VC-1 &SPA (Ours) \\\midrule
Picking Cube &28.00 &64.00 &20.00 &28.00 &56.00 &32.00 &52.00 &36.00 &40.00 &64.00 \\
Stacking Cube &16.00 &32.00 &4.00 &16.00 &8.00 &8.00 &36.00 &20.00 &16.00 &48.00 \\
Folding Cloth &48.00 &64.00 &32.00 &24.00 &28.00 &48.00 &44.00 &64.00 &60.00 &84.00 \\\midrule
Mean S.R. &30.67 &53.33 &18.67 &22.67 &30.67 &29.33 &44.00 &40.00 &38.67 &\textbf{65.33} \\
\bottomrule
\end{tabular}
}
\vspace{-1.2em}
\end{table}

\section{Related Work}
\label{sec:related-work}
\noindent \textbf{Representation Learning for Computer Vision.} Recent advances in computer vision have increasingly focused on unsupervised and self-supervised learning to utilize large amounts of unlabeled data. Techniques like contrastive learning~\citep{chen2020simple,chen2021empirical,chen2020improved,he2020momentum}, masked autoencoders~\citep{he2022masked,feichtenhofer2022masked,bachmann2022multimae,tong2022videomae,wang2023videomaev2}, and self-distillation~\citep{caron2021emerging,oquab2023dinov2,ranzinger2024radio} have shown that effective representations can be learned without supervision. Moreover, multi-modal pre-training approaches~\citep{radford2021learning,fang2023eva,chen2024internvl} leverage language to learn more comprehensive representations. These developments have significantly improved transfer learning capabilities while also displaying zero-shot abilities.

\noindent \textbf{Representation Learning for Embodied AI.} Recent advances in embodied AI representation learning, inspired by computer vision, have applied techniques such as contrastive~\citep{nair2022r3m,yang2023comae} and masked autoencoders~\citep{radosavovic2023real,majumdar2023we,karamcheti2023voltron,yang2024spatiotemporal} to embodied AI. 
However, these approaches often emphasize semantic learning while overlooking the specific needs of embodied AI tasks. %
In this work, we propose a spatial hypothesis specifically for embodied AI representation learning, and we demonstrate how a standard 2D backbone can integrate 3D spatial awareness.

\rebuttal{
\noindent \textbf{3D Robot Learning and 3D-Aware Computer Vision.} Prior work in 3D robot learning has often relied on explicit 3D input~\citep{zhu2024point,ze20243d,wang2024rise,wang2024dexcap,shridhar2023perceiver,chen23polarnet}, or lifting 2D features into 3D spaces~\citep{ke20243d,goyal2024rvt2}, providing a strong foundation for our spatial hypothesis. Given the scalability challenges of explicit 3D observations, some computer vision research has explored integrating 3D spatial awareness into 2D backbones~\citep{yang2024unipad,zhu2023ponderv2,yue2025improving,zhang2024hvdistill}. To the best of our knowledge, SPA is the first to systematically investigate this approach in embodied AI.
}

\noindent \textbf{Neural Rendering.} Recent advances in 3D vision, particularly in neural rendering~\citep{mildenhall2021nerf}, have enabled the encoding of scenes using neural networks, which support differentiable rendering and reconstruction. Alongside improvements in neural rendering techniques themselves~\citep{wang2021neus,zhu2023x,gropp2020implicit,ortiz2022isdf,wang2022go}, the Ponder series~\citep{huang2023ponder,zhu2023ponderv2,yang2024unipad} and subsequent works~\citep{wang2024distillnerf,irshad2024nerfmae} have applied differentiable neural rendering for representation learning. However, they have primarily focused on 3D perception and autonomous driving scenarios. To the best of our knowledge, our work is the first to apply neural rendering for embodied AI representation learning using a standard 2D backbone, marking a novel contribution to this area of research.

\section{Conclusion, Limitations, and Future Work}
\label{sec:conclusion}
In this work, we propose that 3D spatial awareness is crucial for embodied AI and introduce SPA, a novel framework that pre-trains a standard ViT backbone with 3D spatial awareness. To validate our hypothesis, we conduct the largest-scale embodied evaluation to date, over 15 times larger than previous studies. Our experiments demonstrate the clear superiority of SPA and highlight the importance of 3D awareness. Despite strong results across simulated and real robotic tasks, limitations remain. Our evaluation is currently restricted to imitation learning (specifically behavior cloning), and exploring SPA's performance in other settings, such as reinforcement learning, presents an exciting future direction. \rebuttal{Incorporating SPA into VLMs to enhance their performance on spatial aware tasks~\citep{chen2024spatialvlm,majumdar2024openeqa} also represents an exciting research area.} Additionally, SPA currently focuses on static multi-view scenes; extending it to dynamic, temporal scenarios could enhance its generality. Lastly, while we use the ViT encoder for fair comparison, the volume decoder’s multi-view interaction knowledge could be leveraged in policy learning, offering further potential for improvement.

\subsubsection*{Acknowledgments}
This work is supported by the National Key R\&D Program of China (NO.2022ZD0160102) and Shanghai Artificial Intelligence Laboratory.

\bibliography{iclr2025_conference}

\begin{thebibliography}{84}
\providecommand{\natexlab}[1]{#1}
\providecommand{\url}[1]{\texttt{#1}}
\expandafter\ifx\csname urlstyle\endcsname\relax
  \providecommand{\doi}[1]{doi: #1}\else
  \providecommand{\doi}{doi: \begingroup \urlstyle{rm}\Url}\fi

\bibitem[Armeni et~al.(2017)Armeni, Sax, Zamir, and Savarese]{armeni2017joint}
Iro Armeni, Sasha Sax, Amir~R Zamir, and Silvio Savarese.
\newblock Joint 2d-3d-semantic data for indoor scene understanding.
\newblock \emph{arXiv preprint arXiv:1702.01105}, 2017.

\bibitem[Avetisyan et~al.(2024)Avetisyan, Xie, Howard-Jenkins, Yang, Aroudj, Patra, Zhang, Frost, Holland, Orme, et~al.]{avetisyan2024scenescript}
Armen Avetisyan, Christopher Xie, Henry Howard-Jenkins, Tsun-Yi Yang, Samir Aroudj, Suvam Patra, Fuyang Zhang, Duncan Frost, Luke Holland, Campbell Orme, et~al.
\newblock Scenescript: Reconstructing scenes with an autoregressive structured language model.
\newblock \emph{arXiv preprint arXiv:2403.13064}, 2024.

\bibitem[Bachmann et~al.(2022)Bachmann, Mizrahi, Atanov, and Zamir]{bachmann2022multimae}
Roman Bachmann, David Mizrahi, Andrei Atanov, and Amir Zamir.
\newblock Multimae: Multi-modal multi-task masked autoencoders.
\newblock In \emph{European Conference on Computer Vision}, pp.\  348--367. Springer, 2022.

\bibitem[Baruch et~al.(2021)Baruch, Chen, Dehghan, Dimry, Feigin, Fu, Gebauer, Joffe, Kurz, Schwartz, et~al.]{baruch2021arkitscenes}
Gilad Baruch, Zhuoyuan Chen, Afshin Dehghan, Tal Dimry, Yuri Feigin, Peter Fu, Thomas Gebauer, Brandon Joffe, Daniel Kurz, Arik Schwartz, et~al.
\newblock Arkitscenes: A diverse real-world dataset for 3d indoor scene understanding using mobile rgb-d data.
\newblock \emph{arXiv preprint arXiv:2111.08897}, 2021.

\bibitem[Caron et~al.(2021)Caron, Touvron, Misra, J{\'e}gou, Mairal, Bojanowski, and Joulin]{caron2021emerging}
Mathilde Caron, Hugo Touvron, Ishan Misra, Herv{\'e} J{\'e}gou, Julien Mairal, Piotr Bojanowski, and Armand Joulin.
\newblock Emerging properties in self-supervised vision transformers.
\newblock In \emph{Proceedings of the IEEE/CVF international conference on computer vision}, pp.\  9650--9660, 2021.

\bibitem[Chen et~al.(2024{\natexlab{a}})Chen, Xu, Kirmani, Ichter, Sadigh, Guibas, and Xia]{chen2024spatialvlm}
Boyuan Chen, Zhuo Xu, Sean Kirmani, Brain Ichter, Dorsa Sadigh, Leonidas Guibas, and Fei Xia.
\newblock Spatialvlm: Endowing vision-language models with spatial reasoning capabilities.
\newblock In \emph{Proceedings of the IEEE/CVF Conference on Computer Vision and Pattern Recognition}, pp.\  14455--14465, 2024{\natexlab{a}}.

\bibitem[Chen et~al.(2023)Chen, Garcia, Schmid, and Laptev]{chen23polarnet}
Shizhe Chen, Ricardo Garcia, Cordelia Schmid, and Ivan Laptev.
\newblock Polarnet: 3d point clouds for language-guided robotic manipulation.
\newblock 2023.

\bibitem[Chen et~al.(2020{\natexlab{a}})Chen, Kornblith, Norouzi, and Hinton]{chen2020simple}
Ting Chen, Simon Kornblith, Mohammad Norouzi, and Geoffrey Hinton.
\newblock A simple framework for contrastive learning of visual representations.
\newblock In \emph{International conference on machine learning}, pp.\  1597--1607. PMLR, 2020{\natexlab{a}}.

\bibitem[Chen et~al.(2020{\natexlab{b}})Chen, Fan, Girshick, and He]{chen2020improved}
Xinlei Chen, Haoqi Fan, Ross Girshick, and Kaiming He.
\newblock Improved baselines with momentum contrastive learning.
\newblock \emph{arXiv preprint arXiv:2003.04297}, 2020{\natexlab{b}}.

\bibitem[Chen et~al.(2021)Chen, Xie, and He]{chen2021empirical}
Xinlei Chen, Saining Xie, and Kaiming He.
\newblock An empirical study of training self-supervised vision transformers.
\newblock In \emph{Proceedings of the IEEE/CVF international conference on computer vision}, pp.\  9640--9649, 2021.

\bibitem[Chen et~al.(2024{\natexlab{b}})Chen, Wu, Wang, Su, Chen, Xing, Zhong, Zhang, Zhu, Lu, et~al.]{chen2024internvl}
Zhe Chen, Jiannan Wu, Wenhai Wang, Weijie Su, Guo Chen, Sen Xing, Muyan Zhong, Qinglong Zhang, Xizhou Zhu, Lewei Lu, et~al.
\newblock Internvl: Scaling up vision foundation models and aligning for generic visual-linguistic tasks.
\newblock In \emph{Proceedings of the IEEE/CVF Conference on Computer Vision and Pattern Recognition}, pp.\  24185--24198, 2024{\natexlab{b}}.

\bibitem[Chi et~al.(2023)Chi, Feng, Du, Xu, Cousineau, Burchfiel, and Song]{chi2023diffusion}
Cheng Chi, Siyuan Feng, Yilun Du, Zhenjia Xu, Eric Cousineau, Benjamin Burchfiel, and Shuran Song.
\newblock Diffusion policy: Visuomotor policy learning via action diffusion.
\newblock \emph{arXiv preprint arXiv:2303.04137}, 2023.

\bibitem[Contributors(2024)]{realrobot2024}
RealRobot Contributors.
\newblock Realrobot: A project for open-sourced robot learning research. \url{https://github.com/HaoyiZhu/RealRobot}.
\newblock 2024.

\bibitem[Dai et~al.(2017)Dai, Chang, Savva, Halber, Funkhouser, and Nie{\ss}ner]{dai2017scannet}
Angela Dai, Angel~X Chang, Manolis Savva, Maciej Halber, Thomas Funkhouser, and Matthias Nie{\ss}ner.
\newblock Scannet: Richly-annotated 3d reconstructions of indoor scenes.
\newblock In \emph{Proceedings of the IEEE conference on computer vision and pattern recognition}, pp.\  5828--5839, 2017.

\bibitem[Deng et~al.(2009)Deng, Dong, Socher, Li, Li, and Fei-Fei]{deng2009imagenet}
Jia Deng, Wei Dong, Richard Socher, Li-Jia Li, Kai Li, and Li~Fei-Fei.
\newblock Imagenet: A large-scale hierarchical image database.
\newblock In \emph{2009 IEEE conference on computer vision and pattern recognition}, pp.\  248--255. Ieee, 2009.

\bibitem[Dosovitskiy et~al.(2021)Dosovitskiy, Beyer, Kolesnikov, Weissenborn, Zhai, Unterthiner, Dehghani, Minderer, Heigold, Gelly, Uszkoreit, and Houlsby]{dosovitskiy2020vit}
Alexey Dosovitskiy, Lucas Beyer, Alexander Kolesnikov, Dirk Weissenborn, Xiaohua Zhai, Thomas Unterthiner, Mostafa Dehghani, Matthias Minderer, Georg Heigold, Sylvain Gelly, Jakob Uszkoreit, and Neil Houlsby.
\newblock An image is worth 16x16 words: Transformers for image recognition at scale.
\newblock \emph{ICLR}, 2021.

\bibitem[El~Banani et~al.(2024)El~Banani, Raj, Maninis, Kar, Li, Rubinstein, Sun, Guibas, Johnson, and Jampani]{el2024probing}
Mohamed El~Banani, Amit Raj, Kevis-Kokitsi Maninis, Abhishek Kar, Yuanzhen Li, Michael Rubinstein, Deqing Sun, Leonidas Guibas, Justin Johnson, and Varun Jampani.
\newblock Probing the 3d awareness of visual foundation models.
\newblock In \emph{Proceedings of the IEEE/CVF Conference on Computer Vision and Pattern Recognition}, pp.\  21795--21806, 2024.

\bibitem[Fang et~al.(2020)Fang, Wang, Gou, and Lu]{fang2020graspnet}
Hao-Shu Fang, Chenxi Wang, Minghao Gou, and Cewu Lu.
\newblock Graspnet-1billion: A large-scale benchmark for general object grasping.
\newblock In \emph{Proceedings of the IEEE/CVF conference on computer vision and pattern recognition}, pp.\  11444--11453, 2020.

\bibitem[Fang et~al.(2023{\natexlab{a}})Fang, Fang, Tang, Liu, Wang, Zhu, and Lu]{fang2023rh20t}
Hao-Shu Fang, Hongjie Fang, Zhenyu Tang, Jirong Liu, Junbo Wang, Haoyi Zhu, and Cewu Lu.
\newblock Rh20t: A robotic dataset for learning diverse skills in one-shot.
\newblock In \emph{RSS 2023 Workshop on Learning for Task and Motion Planning}, 2023{\natexlab{a}}.

\bibitem[Fang et~al.(2023{\natexlab{b}})Fang, Wang, Xie, Sun, Wu, Wang, Huang, Wang, and Cao]{fang2023eva}
Yuxin Fang, Wen Wang, Binhui Xie, Quan Sun, Ledell Wu, Xinggang Wang, Tiejun Huang, Xinlong Wang, and Yue Cao.
\newblock Eva: Exploring the limits of masked visual representation learning at scale.
\newblock In \emph{Proceedings of the IEEE/CVF Conference on Computer Vision and Pattern Recognition}, pp.\  19358--19369, 2023{\natexlab{b}}.

\bibitem[Feichtenhofer et~al.(2022)Feichtenhofer, Li, He, et~al.]{feichtenhofer2022masked}
Christoph Feichtenhofer, Yanghao Li, Kaiming He, et~al.
\newblock Masked autoencoders as spatiotemporal learners.
\newblock \emph{Advances in neural information processing systems}, 35:\penalty0 35946--35958, 2022.

\bibitem[Gou et~al.(2021)Gou, Fang, Zhu, Xu, Wang, and Lu]{gou2021rgb}
Minghao Gou, Hao-Shu Fang, Zhanda Zhu, Sheng Xu, Chenxi Wang, and Cewu Lu.
\newblock Rgb matters: Learning 7-dof grasp poses on monocular rgbd images.
\newblock In \emph{2021 IEEE International Conference on Robotics and Automation (ICRA)}, pp.\  13459--13466. IEEE, 2021.

\bibitem[Goyal et~al.(2024)Goyal, Blukis, Xu, Guo, Chao, and Fox]{goyal2024rvt2}
Ankit Goyal, Valts Blukis, Jie Xu, Yijie Guo, Yu-Wei Chao, and Dieter Fox.
\newblock Rvt2: Learning precise manipulation from few demonstrations.
\newblock \emph{RSS}, 2024.

\bibitem[Gropp et~al.(2020)Gropp, Yariv, Haim, Atzmon, and Lipman]{gropp2020implicit}
Amos Gropp, Lior Yariv, Niv Haim, Matan Atzmon, and Yaron Lipman.
\newblock Implicit geometric regularization for learning shapes.
\newblock \emph{arXiv preprint arXiv:2002.10099}, 2020.

\bibitem[Gupta et~al.(2019)Gupta, Kumar, Lynch, Levine, and Hausman]{gupta2019relay}
Abhishek Gupta, Vikash Kumar, Corey Lynch, Sergey Levine, and Karol Hausman.
\newblock Relay policy learning: Solving long-horizon tasks via imitation and reinforcement learning.
\newblock \emph{arXiv preprint arXiv:1910.11956}, 2019.

\bibitem[He et~al.(2020)He, Fan, Wu, Xie, and Girshick]{he2020momentum}
Kaiming He, Haoqi Fan, Yuxin Wu, Saining Xie, and Ross Girshick.
\newblock Momentum contrast for unsupervised visual representation learning.
\newblock In \emph{Proceedings of the IEEE/CVF conference on computer vision and pattern recognition}, pp.\  9729--9738, 2020.

\bibitem[He et~al.(2022)He, Chen, Xie, Li, Doll{\'a}r, and Girshick]{he2022masked}
Kaiming He, Xinlei Chen, Saining Xie, Yanghao Li, Piotr Doll{\'a}r, and Ross Girshick.
\newblock Masked autoencoders are scalable vision learners.
\newblock In \emph{Proceedings of the IEEE/CVF conference on computer vision and pattern recognition}, pp.\  16000--16009, 2022.

\bibitem[Hendrycks \& Gimpel(2016)Hendrycks and Gimpel]{hendrycks2016gaussian}
Dan Hendrycks and Kevin Gimpel.
\newblock Gaussian error linear units (gelus).
\newblock \emph{arXiv preprint arXiv:1606.08415}, 2016.

\bibitem[Hu et~al.(2023)Hu, Wang, Li, and Gao]{hu2023pre}
Yingdong Hu, Renhao Wang, Li~Erran Li, and Yang Gao.
\newblock For pre-trained vision models in motor control, not all policy learning methods are created equal.
\newblock In \emph{International Conference on Machine Learning}, pp.\  13628--13651. PMLR, 2023.

\bibitem[Huang et~al.(2023)Huang, Peng, He, Yang, Zhou, and Ouyang]{huang2023ponder}
Di~Huang, Sida Peng, Tong He, Honghui Yang, Xiaowei Zhou, and Wanli Ouyang.
\newblock Ponder: Point cloud pre-training via neural rendering.
\newblock In \emph{Proceedings of the IEEE/CVF International Conference on Computer Vision}, pp.\  16089--16098, 2023.

\bibitem[Irshad et~al.(2024)Irshad, Zakharov, Guizilini, Gaidon, Kira, and Ambrus]{irshad2024nerfmae}
Muhammad~Zubair Irshad, Sergey Zakharov, Vitor Guizilini, Adrien Gaidon, Zsolt Kira, and Rares Ambrus.
\newblock Nerf-mae: Masked autoencoders for self-supervised 3d representation learning for neural radiance fields.
\newblock In \emph{European Conference on Computer Vision (ECCV)}, 2024.

\bibitem[James et~al.(2020)James, Ma, Arrojo, and Davison]{james2020rlbench}
Stephen James, Zicong Ma, David~Rovick Arrojo, and Andrew~J Davison.
\newblock Rlbench: The robot learning benchmark \& learning environment.
\newblock \emph{IEEE Robotics and Automation Letters}, 5\penalty0 (2):\penalty0 3019--3026, 2020.

\bibitem[Jampani et~al.(2023)Jampani, Maninis, Engelhardt, Karpur, Truong, Sargent, Popov, Araujo, Martin~Brualla, Patel, et~al.]{jampani2023navi}
Varun Jampani, Kevis-Kokitsi Maninis, Andreas Engelhardt, Arjun Karpur, Karen Truong, Kyle Sargent, Stefan Popov, Andr{\'e} Araujo, Ricardo Martin~Brualla, Kaushal Patel, et~al.
\newblock Navi: Category-agnostic image collections with high-quality 3d shape and pose annotations.
\newblock \emph{Advances in Neural Information Processing Systems}, 36:\penalty0 76061--76084, 2023.

\bibitem[Karamcheti et~al.(2023)Karamcheti, Nair, Chen, Kollar, Finn, Sadigh, and Liang]{karamcheti2023voltron}
Siddharth Karamcheti, Suraj Nair, Annie~S. Chen, Thomas Kollar, Chelsea Finn, Dorsa Sadigh, and Percy Liang.
\newblock Language-driven representation learning for robotics.
\newblock In \emph{Robotics: Science and Systems (RSS)}, 2023.

\bibitem[Ke et~al.(2024)Ke, Gkanatsios, and Fragkiadaki]{ke20243d}
Tsung-Wei Ke, Nikolaos Gkanatsios, and Katerina Fragkiadaki.
\newblock 3d diffuser actor: Policy diffusion with 3d scene representations.
\newblock \emph{arXiv preprint arXiv:2402.10885}, 2024.

\bibitem[Khazatsky et~al.(2024)Khazatsky, Pertsch, Nair, Balakrishna, Dasari, Karamcheti, Nasiriany, Srirama, Chen, Ellis, et~al.]{khazatsky2024droid}
Alexander Khazatsky, Karl Pertsch, Suraj Nair, Ashwin Balakrishna, Sudeep Dasari, Siddharth Karamcheti, Soroush Nasiriany, Mohan~Kumar Srirama, Lawrence~Yunliang Chen, Kirsty Ellis, et~al.
\newblock Droid: A large-scale in-the-wild robot manipulation dataset.
\newblock \emph{arXiv preprint arXiv:2403.12945}, 2024.

\bibitem[Koch(2024)]{Alexander}
Alexander Koch.
\newblock Low-cost robot arm.
\newblock \url{https://github.com/AlexanderKoch-Koch/low_cost_robot}, 2024.
\newblock URL \url{https://github.com/AlexanderKoch-Koch/low_cost_robot}.
\newblock GitHub repository.

\bibitem[Kumar(2016)]{Kumar2016thesis}
Vikash Kumar.
\newblock \emph{Manipulators and Manipulation in high dimensional spaces}.
\newblock PhD thesis, University of Washington, Seattle, 2016.
\newblock URL \url{https://digital.lib.washington.edu/researchworks/handle/1773/38104}.

\bibitem[Li et~al.(2022)Li, Wang, Li, Xie, Sima, Lu, Qiao, and Dai]{li2022bevformer}
Zhiqi Li, Wenhai Wang, Hongyang Li, Enze Xie, Chonghao Sima, Tong Lu, Yu~Qiao, and Jifeng Dai.
\newblock Bevformer: Learning bird’s-eye-view representation from multi-camera images via spatiotemporal transformers.
\newblock In \emph{European conference on computer vision}, pp.\  1--18. Springer, 2022.

\bibitem[Liu et~al.(2024)Liu, Zhu, Gao, Feng, Liu, Zhu, and Stone]{liu2024libero}
Bo~Liu, Yifeng Zhu, Chongkai Gao, Yihao Feng, Qiang Liu, Yuke Zhu, and Peter Stone.
\newblock Libero: Benchmarking knowledge transfer for lifelong robot learning.
\newblock \emph{Advances in Neural Information Processing Systems}, 36, 2024.

\bibitem[Loshchilov et~al.(2017)Loshchilov, Hutter, et~al.]{loshchilov2017fixing}
Ilya Loshchilov, Frank Hutter, et~al.
\newblock Fixing weight decay regularization in adam.
\newblock \emph{arXiv preprint arXiv:1711.05101}, 5, 2017.

\bibitem[Majumdar et~al.(2023)Majumdar, Yadav, Arnaud, Ma, Chen, Silwal, Jain, Berges, Wu, Vakil, et~al.]{majumdar2023we}
Arjun Majumdar, Karmesh Yadav, Sergio Arnaud, Jason Ma, Claire Chen, Sneha Silwal, Aryan Jain, Vincent-Pierre Berges, Tingfan Wu, Jay Vakil, et~al.
\newblock Where are we in the search for an artificial visual cortex for embodied intelligence?
\newblock \emph{Advances in Neural Information Processing Systems}, 36:\penalty0 655--677, 2023.

\bibitem[Majumdar et~al.(2024)Majumdar, Ajay, Zhang, Putta, Yenamandra, Henaff, Silwal, Mcvay, Maksymets, Arnaud, et~al.]{majumdar2024openeqa}
Arjun Majumdar, Anurag Ajay, Xiaohan Zhang, Pranav Putta, Sriram Yenamandra, Mikael Henaff, Sneha Silwal, Paul Mcvay, Oleksandr Maksymets, Sergio Arnaud, et~al.
\newblock Openeqa: Embodied question answering in the era of foundation models.
\newblock In \emph{Proceedings of the IEEE/CVF Conference on Computer Vision and Pattern Recognition}, pp.\  16488--16498, 2024.

\bibitem[Mildenhall et~al.(2021)Mildenhall, Srinivasan, Tancik, Barron, Ramamoorthi, and Ng]{mildenhall2021nerf}
Ben Mildenhall, Pratul~P Srinivasan, Matthew Tancik, Jonathan~T Barron, Ravi Ramamoorthi, and Ren Ng.
\newblock Nerf: Representing scenes as neural radiance fields for view synthesis.
\newblock \emph{Communications of the ACM}, 65\penalty0 (1):\penalty0 99--106, 2021.

\bibitem[Nair et~al.(2022)Nair, Rajeswaran, Kumar, Finn, and Gupta]{nair2022r3m}
Suraj Nair, Aravind Rajeswaran, Vikash Kumar, Chelsea Finn, and Abhinav Gupta.
\newblock R3m: A universal visual representation for robot manipulation.
\newblock \emph{arXiv preprint arXiv:2203.12601}, 2022.

\bibitem[Oquab et~al.(2023)Oquab, Darcet, Moutakanni, Vo, Szafraniec, Khalidov, Fernandez, Haziza, Massa, El-Nouby, et~al.]{oquab2023dinov2}
Maxime Oquab, Timoth{\'e}e Darcet, Th{\'e}o Moutakanni, Huy Vo, Marc Szafraniec, Vasil Khalidov, Pierre Fernandez, Daniel Haziza, Francisco Massa, Alaaeldin El-Nouby, et~al.
\newblock Dinov2: Learning robust visual features without supervision.
\newblock \emph{arXiv preprint arXiv:2304.07193}, 2023.

\bibitem[Ortiz et~al.(2022)Ortiz, Clegg, Dong, Sucar, Novotny, Zollhoefer, and Mukadam]{ortiz2022isdf}
Joseph Ortiz, Alexander Clegg, Jing Dong, Edgar Sucar, David Novotny, Michael Zollhoefer, and Mustafa Mukadam.
\newblock isdf: Real-time neural signed distance fields for robot perception.
\newblock \emph{arXiv preprint arXiv:2204.02296}, 2022.

\bibitem[Pan et~al.(2023)Pan, Charron, Yang, Peters, Whelan, Kong, Parkhi, Newcombe, and Ren]{pan2023aria}
Xiaqing Pan, Nicholas Charron, Yongqian Yang, Scott Peters, Thomas Whelan, Chen Kong, Omkar Parkhi, Richard Newcombe, and Yuheng~Carl Ren.
\newblock Aria digital twin: A new benchmark dataset for egocentric 3d machine perception.
\newblock In \emph{Proceedings of the IEEE/CVF International Conference on Computer Vision}, pp.\  20133--20143, 2023.

\bibitem[Radford et~al.(2021)Radford, Kim, Hallacy, Ramesh, Goh, Agarwal, Sastry, Askell, Mishkin, Clark, et~al.]{radford2021learning}
Alec Radford, Jong~Wook Kim, Chris Hallacy, Aditya Ramesh, Gabriel Goh, Sandhini Agarwal, Girish Sastry, Amanda Askell, Pamela Mishkin, Jack Clark, et~al.
\newblock Learning transferable visual models from natural language supervision.
\newblock In \emph{International conference on machine learning}, pp.\  8748--8763. PMLR, 2021.

\bibitem[Radosavovic et~al.(2023)Radosavovic, Xiao, James, Abbeel, Malik, and Darrell]{radosavovic2023real}
Ilija Radosavovic, Tete Xiao, Stephen James, Pieter Abbeel, Jitendra Malik, and Trevor Darrell.
\newblock Real-world robot learning with masked visual pre-training.
\newblock In \emph{Conference on Robot Learning}, pp.\  416--426. PMLR, 2023.

\bibitem[Ranftl et~al.(2020)Ranftl, Lasinger, Hafner, Schindler, and Koltun]{Ranftl2020}
Ren\'{e} Ranftl, Katrin Lasinger, David Hafner, Konrad Schindler, and Vladlen Koltun.
\newblock Towards robust monocular depth estimation: Mixing datasets for zero-shot cross-dataset transfer.
\newblock \emph{IEEE Transactions on Pattern Analysis and Machine Intelligence (TPAMI)}, 2020.

\bibitem[Ranzinger et~al.(2024)Ranzinger, Heinrich, Kautz, and Molchanov]{ranzinger2024radio}
Mike Ranzinger, Greg Heinrich, Jan Kautz, and Pavlo Molchanov.
\newblock Am-radio: Agglomerative vision foundation model reduce all domains into one.
\newblock In \emph{Proceedings of the IEEE/CVF Conference on Computer Vision and Pattern Recognition}, pp.\  12490--12500, 2024.

\bibitem[Roberts et~al.(2021)Roberts, Ramapuram, Ranjan, Kumar, Bautista, Paczan, Webb, and Susskind]{roberts2021hypersim}
Mike Roberts, Jason Ramapuram, Anurag Ranjan, Atulit Kumar, Miguel~Angel Bautista, Nathan Paczan, Russ Webb, and Joshua~M Susskind.
\newblock Hypersim: A photorealistic synthetic dataset for holistic indoor scene understanding.
\newblock In \emph{Proceedings of the IEEE/CVF international conference on computer vision}, pp.\  10912--10922, 2021.

\bibitem[Shang et~al.(2024)Shang, Schmeckpeper, May, Minniti, Kelestemur, Watkins, and Herlant]{shang2024theia}
Jinghuan Shang, Karl Schmeckpeper, Brandon~B. May, Maria~Vittoria Minniti, Tarik Kelestemur, David Watkins, and Laura Herlant.
\newblock Theia: Distilling diverse vision foundation models for robot learning.
\newblock \emph{arXiv}, 2024.

\bibitem[Shi et~al.(2016)Shi, Caballero, Husz{\'a}r, Totz, Aitken, Bishop, Rueckert, and Wang]{shi2016real}
Wenzhe Shi, Jose Caballero, Ferenc Husz{\'a}r, Johannes Totz, Andrew~P Aitken, Rob Bishop, Daniel Rueckert, and Zehan Wang.
\newblock Real-time single image and video super-resolution using an efficient sub-pixel convolutional neural network.
\newblock In \emph{Proceedings of the IEEE conference on computer vision and pattern recognition}, pp.\  1874--1883, 2016.

\bibitem[Shridhar et~al.(2023)Shridhar, Manuelli, and Fox]{shridhar2023perceiver}
Mohit Shridhar, Lucas Manuelli, and Dieter Fox.
\newblock Perceiver-actor: A multi-task transformer for robotic manipulation.
\newblock In \emph{Conference on Robot Learning}, pp.\  785--799. PMLR, 2023.

\bibitem[Smith \& Topin(2019)Smith and Topin]{smith2019super}
Leslie~N Smith and Nicholay Topin.
\newblock Super-convergence: Very fast training of neural networks using large learning rates.
\newblock In \emph{Artificial intelligence and machine learning for multi-domain operations applications}, volume 11006, pp.\  369--386. SPIE, 2019.

\bibitem[Tong et~al.(2022)Tong, Song, Wang, and Wang]{tong2022videomae}
Zhan Tong, Yibing Song, Jue Wang, and Limin Wang.
\newblock Videomae: Masked autoencoders are data-efficient learners for self-supervised video pre-training.
\newblock \emph{Advances in neural information processing systems}, 35:\penalty0 10078--10093, 2022.

\bibitem[Tunyasuvunakool et~al.(2020)Tunyasuvunakool, Muldal, Doron, Liu, Bohez, Merel, Erez, Lillicrap, Heess, and Tassa]{tunyasuvunakool2020dm_control}
Saran Tunyasuvunakool, Alistair Muldal, Yotam Doron, Siqi Liu, Steven Bohez, Josh Merel, Tom Erez, Timothy Lillicrap, Nicolas Heess, and Yuval Tassa.
\newblock dm\_control: Software and tasks for continuous control.
\newblock \emph{Software Impacts}, 6:\penalty0 100022, 2020.

\bibitem[Wang et~al.(2024{\natexlab{a}})Wang, Shi, Wang, Zhang, Fei-Fei, and Liu]{wang2024dexcap}
Chen Wang, Haochen Shi, Weizhuo Wang, Ruohan Zhang, Li~Fei-Fei, and C~Karen Liu.
\newblock Dexcap: Scalable and portable mocap data collection system for dexterous manipulation.
\newblock \emph{arXiv preprint arXiv:2403.07788}, 2024{\natexlab{a}}.

\bibitem[Wang et~al.(2024{\natexlab{b}})Wang, Fang, Fang, and Lu]{wang2024rise}
Chenxi Wang, Hongjie Fang, Hao-Shu Fang, and Cewu Lu.
\newblock Rise: 3d perception makes real-world robot imitation simple and effective.
\newblock \emph{arXiv preprint arXiv:2404.12281}, 2024{\natexlab{b}}.

\bibitem[Wang et~al.(2022)Wang, Bleja, and Agapito]{wang2022go}
Jingwen Wang, Tymoteusz Bleja, and Lourdes Agapito.
\newblock Go-surf: Neural feature grid optimization for fast, high-fidelity rgb-d surface reconstruction.
\newblock In \emph{2022 International Conference on 3D Vision (3DV)}, pp.\  433--442. IEEE, 2022.

\bibitem[Wang et~al.(2024{\natexlab{c}})Wang, Kim, Yang, Yu, Ivanovic, Waslander, Wang, Fidler, Pavone, and Karkus]{wang2024distillnerf}
Letian Wang, Seung~Wook Kim, Jiawei Yang, Cunjun Yu, Boris Ivanovic, Steven~L Waslander, Yue Wang, Sanja Fidler, Marco Pavone, and Peter Karkus.
\newblock Distillnerf: Perceiving 3d scenes from single-glance images by distilling neural fields and foundation model features.
\newblock \emph{arXiv preprint arXiv:2406.12095}, 2024{\natexlab{c}}.

\bibitem[Wang et~al.(2023)Wang, Huang, Zhao, Tong, He, Wang, Wang, and Qiao]{wang2023videomaev2}
Limin Wang, Bingkun Huang, Zhiyu Zhao, Zhan Tong, Yinan He, Yi~Wang, Yali Wang, and Yu~Qiao.
\newblock Videomae {V2:} scaling video masked autoencoders with dual masking.
\newblock In \emph{CVPR}, pp.\  14549--14560, 2023.

\bibitem[Wang et~al.(2021)Wang, Liu, Liu, Theobalt, Komura, and Wang]{wang2021neus}
Peng Wang, Lingjie Liu, Yuan Liu, Christian Theobalt, Taku Komura, and Wenping Wang.
\newblock Neus: Learning neural implicit surfaces by volume rendering for multi-view reconstruction.
\newblock \emph{arXiv preprint arXiv:2106.10689}, 2021.

\bibitem[Weinzaepfel et~al.(2023)Weinzaepfel, Lucas, Leroy, Cabon, Arora, Br{\'e}gier, Csurka, Antsfeld, Chidlovskii, and Revaud]{croco_v2}
Philippe Weinzaepfel, Thomas Lucas, Vincent Leroy, Yohann Cabon, Vaibhav Arora, Romain Br{\'e}gier, Gabriela Csurka, Leonid Antsfeld, Boris Chidlovskii, and J{\'e}r{\^o}me Revaud.
\newblock {CroCo v2: Improved Cross-view Completion Pre-training for Stereo Matching and Optical Flow}.
\newblock In \emph{ICCV}, 2023.

\bibitem[W{\"u}thrich et~al.(2020)W{\"u}thrich, Widmaier, Grimminger, Akpo, Joshi, Agrawal, Hammoud, Khadiv, Bogdanovic, Berenz, et~al.]{wuthrich2020trifinger}
Manuel W{\"u}thrich, Felix Widmaier, Felix Grimminger, Joel Akpo, Shruti Joshi, Vaibhav Agrawal, Bilal Hammoud, Majid Khadiv, Miroslav Bogdanovic, Vincent Berenz, et~al.
\newblock Trifinger: An open-source robot for learning dexterity.
\newblock \emph{arXiv preprint arXiv:2008.03596}, 2020.

\bibitem[Yang et~al.(2024{\natexlab{a}})Yang, Zhang, Huang, Wu, Zhu, He, Tang, Zhao, Qiu, Lin, et~al.]{yang2024unipad}
Honghui Yang, Sha Zhang, Di~Huang, Xiaoyang Wu, Haoyi Zhu, Tong He, Shixiang Tang, Hengshuang Zhao, Qibo Qiu, Binbin Lin, et~al.
\newblock Unipad: A universal pre-training paradigm for autonomous driving.
\newblock In \emph{Proceedings of the IEEE/CVF Conference on Computer Vision and Pattern Recognition}, pp.\  15238--15250, 2024{\natexlab{a}}.

\bibitem[Yang et~al.(2023)Yang, Guo, Wu, and Wang]{yang2023comae}
Jiange Yang, Sheng Guo, Gangshan Wu, and Limin Wang.
\newblock Comae: single model hybrid pre-training on small-scale rgb-d datasets.
\newblock In \emph{Proceedings of the AAAI Conference on Artificial Intelligence}, volume~37, pp.\  3145--3154, 2023.

\bibitem[Yang et~al.(2024{\natexlab{b}})Yang, Liu, Fu, Pan, Wu, and Wang]{yang2024spatiotemporal}
Jiange Yang, Bei Liu, Jianlong Fu, Bocheng Pan, Gangshan Wu, and Limin Wang.
\newblock Spatiotemporal predictive pre-training for robotic motor control.
\newblock \emph{arXiv preprint arXiv:2403.05304}, 2024{\natexlab{b}}.

\bibitem[Yarats et~al.(2021)Yarats, Fergus, Lazaric, and Pinto]{yarats2021mastering}
Denis Yarats, Rob Fergus, Alessandro Lazaric, and Lerrel Pinto.
\newblock Mastering visual continuous control: Improved data-augmented reinforcement learning.
\newblock \emph{arXiv preprint arXiv:2107.09645}, 2021.

\bibitem[Yeshwanth et~al.(2023)Yeshwanth, Liu, Nie{\ss}ner, and Dai]{yeshwanth2023scannet++}
Chandan Yeshwanth, Yueh-Cheng Liu, Matthias Nie{\ss}ner, and Angela Dai.
\newblock Scannet++: A high-fidelity dataset of 3d indoor scenes.
\newblock In \emph{Proceedings of the IEEE/CVF International Conference on Computer Vision}, pp.\  12--22, 2023.

\bibitem[Yu et~al.(2021)Yu, Li, Tancik, Li, Ng, and Kanazawa]{yu2021plenoctrees}
Alex Yu, Ruilong Li, Matthew Tancik, Hao Li, Ren Ng, and Angjoo Kanazawa.
\newblock Plenoctrees for real-time rendering of neural radiance fields.
\newblock In \emph{Proceedings of the IEEE/CVF International Conference on Computer Vision}, pp.\  5752--5761, 2021.

\bibitem[Yu et~al.(2020)Yu, Quillen, He, Julian, Hausman, Finn, and Levine]{yu2020meta}
Tianhe Yu, Deirdre Quillen, Zhanpeng He, Ryan Julian, Karol Hausman, Chelsea Finn, and Sergey Levine.
\newblock Meta-world: A benchmark and evaluation for multi-task and meta reinforcement learning.
\newblock In \emph{Conference on robot learning}, pp.\  1094--1100. PMLR, 2020.

\bibitem[Yue et~al.(2025)Yue, Das, Engelmann, Tang, and Lenssen]{yue2025improving}
Yuanwen Yue, Anurag Das, Francis Engelmann, Siyu Tang, and Jan~Eric Lenssen.
\newblock Improving 2d feature representations by 3d-aware fine-tuning.
\newblock In \emph{European Conference on Computer Vision}, pp.\  57--74. Springer, 2025.

\bibitem[Ze et~al.(2024)Ze, Zhang, Zhang, Hu, Wang, and Xu]{ze20243d}
Yanjie Ze, Gu~Zhang, Kangning Zhang, Chenyuan Hu, Muhan Wang, and Huazhe Xu.
\newblock 3d diffusion policy.
\newblock \emph{arXiv preprint arXiv:2403.03954}, 2024.

\bibitem[Zhang et~al.(2024)Zhang, Deng, Bai, Li, Ouyang, and Zhang]{zhang2024hvdistill}
Sha Zhang, Jiajun Deng, Lei Bai, Houqiang Li, Wanli Ouyang, and Yanyong Zhang.
\newblock Hvdistill: Transferring knowledge from images to point clouds via unsupervised hybrid-view distillation.
\newblock \emph{International Journal of Computer Vision}, pp.\  1--15, 2024.

\bibitem[Zhao et~al.(2023)Zhao, Kumar, Levine, and Finn]{zhao2023learning}
Tony~Z Zhao, Vikash Kumar, Sergey Levine, and Chelsea Finn.
\newblock Learning fine-grained bimanual manipulation with low-cost hardware.
\newblock \emph{arXiv preprint arXiv:2304.13705}, 2023.

\bibitem[Zheng et~al.(2020)Zheng, Zhang, Li, Tang, Gao, and Zhou]{zheng2020structured3d}
Jia Zheng, Junfei Zhang, Jing Li, Rui Tang, Shenghua Gao, and Zihan Zhou.
\newblock Structured3d: A large photo-realistic dataset for structured 3d modeling.
\newblock In \emph{Computer Vision--ECCV 2020: 16th European Conference, Glasgow, UK, August 23--28, 2020, Proceedings, Part IX 16}, pp.\  519--535. Springer, 2020.

\bibitem[Zhu et~al.(2023{\natexlab{a}})Zhu, Fang, and Lu]{zhu2023x}
Haoyi Zhu, Hao-Shu Fang, and Cewu Lu.
\newblock X-nerf: Explicit neural radiance field for multi-scene 360deg insufficient rgb-d views.
\newblock In \emph{Proceedings of the IEEE/CVF Winter Conference on Applications of Computer Vision}, pp.\  5766--5775, 2023{\natexlab{a}}.

\bibitem[Zhu et~al.(2023{\natexlab{b}})Zhu, Yang, Wu, Huang, Zhang, He, He, Zhao, Shen, Qiao, et~al.]{zhu2023ponderv2}
Haoyi Zhu, Honghui Yang, Xiaoyang Wu, Di~Huang, Sha Zhang, Xianglong He, Tong He, Hengshuang Zhao, Chunhua Shen, Yu~Qiao, et~al.
\newblock Ponderv2: Pave the way for 3d foundataion model with a universal pre-training paradigm.
\newblock \emph{arXiv preprint arXiv:2310.08586}, 2023{\natexlab{b}}.

\bibitem[Zhu et~al.(2024)Zhu, Wang, Huang, Ye, Ouyang, and He]{zhu2024point}
Haoyi Zhu, Yating Wang, Di~Huang, Weicai Ye, Wanli Ouyang, and Tong He.
\newblock Point cloud matters: Rethinking the impact of different observation spaces on robot learning.
\newblock \emph{arXiv preprint arXiv:2402.02500}, 2024.

\bibitem[Zhu et~al.(2021)Zhu, Su, Lu, Li, Wang, and Dai]{zhu2021deformdetr}
Xizhou Zhu, Weijie Su, Lewei Lu, Bin Li, Xiaogang Wang, and Jifeng Dai.
\newblock Deformable {DETR:} deformable transformers for end-to-end object detection.
\newblock In \emph{International Conference on Learning Representations}, 2021.

\bibitem[Zhu et~al.(2020)Zhu, Wong, Mandlekar, Mart{\'\i}n-Mart{\'\i}n, Joshi, Nasiriany, and Zhu]{zhu2020robosuite}
Yuke Zhu, Josiah Wong, Ajay Mandlekar, Roberto Mart{\'\i}n-Mart{\'\i}n, Abhishek Joshi, Soroush Nasiriany, and Yifeng Zhu.
\newblock robosuite: A modular simulation framework and benchmark for robot learning.
\newblock \emph{arXiv preprint arXiv:2009.12293}, 2020.

\end{thebibliography}
\bibliographystyle{iclr2025_conference}

\newpage
\appendix
\section{Additional Rendering Losses}
\label{appx:loss}

Here we detail the three additional rendering losses we have applied in \secref{sec:method-loss}.

{\noindent \textbf{Eikonal Regularization Loss.}} The Eikonal regularization loss, denoted as $\mathcal{L}_{\text{eikonal}}$, is a widely used loss function for the regularization of signed distance functions (SDFs)~\citep{gropp2020implicit}. It is defined as:
\begin{equation}
    \mathcal{L}_{\text{eikonal}} = \frac{1}{N_r N_p} \sum_{i=1}^{N_r} \sum_{j=1}^{N_p} \left( 
    \left\| \nabla s(\mathbf{p}_{i,j}) \right\| - 1 \right)^2,
\end{equation}

where $\nabla s(\mathbf{p}_{i,j})$ represents the gradient of the SDF $s$ at the location $\mathbf{p}_{i,j}$. Since the SDF is a distance measure, $\mathcal{L}_{\text{eikonal}}$ encourages the gradients to have unit norm at the query point.

{\noindent \textbf{Near-Surface and Free Space Loss for SDF.}} To improve SDF estimation, we incorporate additional approximate SDF supervision, similar to iSDF~\citep{ortiz2022isdf} and GO-Surf~\citep{wang2022go}. Specifically, for near-surface points, the difference between rendered depth and ground-truth depth serves as pseudo-SDF ground-truth supervision. For points far from the surface, a free space loss is used to further regularize the SDF values.

To compute the approximate SDF supervision, we define an indicator $b(z)$ for each sampled ray point with ray length $z$ and corresponding ground-truth depth $D$:

\begin{equation}
    b(z) = D - z.
\end{equation}
The value $b(z)$ can be considered a credible approximate SDF value when it is small. Let $t$ be a user-defined threshold, set to $0.05$ in our experiments. For sampled ray points satisfying $b(z) \leq t$, we apply the near-surface SDF loss to constrain the SDF prediction $s(z_{i,j})$:

\begin{equation}
    \mathcal{L}_{\text{sdf}} = \frac{1}{N_r N_p} \sum_{i=1}^{N_r} \sum_{j=1}^{N_p} \left| s(z_{i,j}) - b(z_{i,j}) \right|.
\end{equation}

For the remaining sampled ray points, we utilize a free space loss:

\begin{equation}
    \mathcal{L}_{\text{free}} = \frac{1}{N_r N_p} \sum_{i=1}^{N_r} \sum_{j=1}^{N_p} \max\left(0, e^{-\alpha \cdot s(z_{i,j})} - 1, s(z_{i,j}) - b(z_{i,j})\right),
\end{equation}

where $\alpha$ is set to 5, following~\cite{ortiz2022isdf,wang2022go}. Due to the presence of noisy depth images, $\mathcal{L}_{\text{sdf}}$ and $\mathcal{L}_{\text{free}}$ are applied only to rays with valid depth values.

In our experiments, we adopt a similar weighting scheme to GO-Surf~\citep{wang2022go}, setting $\lambda_C = 10.0$, $\lambda_D = 1.0$, $\lambda_{\text{sdf}} = 10.0$, and $\lambda_{\text{free}} = 1.0$. We observe that the Eikonal term can lead to overly smooth reconstructions, so we use a small weight of $0.01$ for the Eikonal loss.

\section{Evaluation Setups}
\label{appx:eval-setups}
Here we detail the setups of our large-scale evaluation in \secref{sec:evaluation-setting}. \rebuttal{For the detailed visualizations of each task, we recommend the readers to read the original simulator's or benchmark's dataset.}

\subsection{Single-Task Benchmarks}
{\noindent \textbf{VC-1~\citep{majumdar2023we}.}} This benchmark includes several simulators. We selected four: Adroit~\citep{Kumar2016thesis}, Meta-World~\citep{yu2020meta}, DMControl~\citep{tunyasuvunakool2020dm_control}, and TriFinger~\citep{wuthrich2020trifinger}. 
The Adroit subset focuses on dexterous manipulation with 2 tasks: \rebuttal{\texttt{Relocate} and \texttt{Pen}}. 
The Meta-World subset addresses two-finger gripper manipulation with 5 tasks: \rebuttal{\texttt{Button Press Topdown}, \texttt{Drawer Open}, \texttt{Bin Picking}, \texttt{Hammer}, and \texttt{Assembly}}. 
The DMControl subset is for locomotion control, also with 5 tasks: \rebuttal{\texttt{Walker Stand}, \texttt{Walker Walk}, \texttt{Reacher Easy}, \texttt{Cheetah Run}, and \texttt{Finger Spin}}. 
The TriFinger subset targets three-finger manipulation with 2 tasks: \rebuttal{\texttt{Reach Cube} and \texttt{Move Cube}}. 
For all tasks, we use a 3-layer MLP as the policy network for each single-task training, following the original implementation. Each task is trained with 100 demonstrations, except for 25 on Meta-World, and evaluated 50 times using the specific seeds 100, 200, and 300. The \texttt{[CLS]} token of a frozen pre-trained ViT is used as the observation feature. All hyper-parameters are kept the same with the original implementation.

{\noindent \textbf{Franka Kitchen~\citep{gupta2019relay}.}} 
Franka-Kitchen is a MuJoCo-modeled simulation environment with a Franka robot in a kitchen scene. Its action space is the 9-dimensional joint velocity with 7 DoF for the arm and 2 DoF for the gripper. Following previous works~\citep{nair2022r3m,karamcheti2023voltron}, we evaluate five tasks: \rebuttal{\texttt{Sliding Door}, \texttt{Turning Light On}, \texttt{Opening Door}, \texttt{Turning Knob}, and \texttt{Opening Microwave}}. Each task spans two camera viewpoints and three random seeds. Similar to the evaluation scheme in VC-1, we utilize 25 demonstrations to train a policy model, which is a 2-layer MLP with hidden sizes [256, 256] preceded by a BatchNorm.

{\noindent \textbf{Meta-World~\citep{yu2020meta}.}} This benchmark comprises a series of tasks in which an agent directs a Sawyer robot arm to manipulate objects in a tabletop environment. We selected 48 tasks, encompassing easy, medium, and hard levels. We implemented the Diffusion Policy~\citep{chi2023diffusion} on this benchmark and adhered to the setup in \cite{ze20243d} to generate 10 demonstrations for each single-task training, followed by evaluation through 20 rollouts. The average results across three fixed seeds (100, 200, 300) are reported. The \texttt{[CLS]} token from a frozen pre-trained ViT serves as the observation feature. \rebuttal{The 48 tasks include: \texttt{Button Press Wall}, \texttt{Door Close}, \texttt{Door Unlock}, \texttt{Drawer Close}, \texttt{Drawer Open}, \texttt{Faucet Close}, \texttt{Plate Slide}, \texttt{Plate Slide Back}, \texttt{Plate Slide Side}, \texttt{Window Close}, \texttt{Basketball}, \texttt{Bin Picking}, \texttt{Box Close}, \texttt{Coffee Push}, \texttt{Assembly}, \texttt{Disassemble}, \texttt{Push Wall}, \texttt{Shelf Place}, \texttt{Door Open}, \texttt{Button Press}, \texttt{Sweep Into}, \texttt{Door Lock}, \texttt{Reach Wall}, \texttt{Hammer}, \texttt{Stick Push}, \texttt{Button Press Topdown}, \texttt{Handle Press Side}, \texttt{Plate Slide Back Side}, \texttt{Sweep}, \texttt{Button Press Topdown Wall}, \texttt{Handle Press}, \texttt{Push}, \texttt{Coffee Pull}, \texttt{Dial Turn}, \texttt{Reach}, \texttt{Coffee Button}, \texttt{Pick Place Wall}, \texttt{Stick Pull}, \texttt{Hand Insert}, \texttt{Peg Insert Side}, \texttt{Pick Place}, \texttt{Faucet Open}, \texttt{Push Back}, \texttt{Lever Pull}, \texttt{Handle Pull}, \texttt{Soccer}, \texttt{Window Open}, and \texttt{Pick Out Of Hole}.}

\subsection{Language-Conditioned Multi-Task Benchmarks}
{\noindent \textbf{RLBench~\citep{james2020rlbench}.}} This benchmark is a prominent language-conditioned multi-task robot learning framework. PolarNet~\citep{chen23polarnet} has categorized all tasks into 9 groups. We selected 71 tasks from RLBench that can be successfully executed and split them into two groups uniformly on categories: Group 1 with 35 tasks and Group 2 with 36 tasks. Each task includes 100 training demonstrations and 25 testing rollouts. For each group, we train a language-conditioned multi-task agent. We employ RVT-2~\citep{goyal2024rvt2}, the state-of-the-art (SOTA) method on this benchmark, as our policy. RVT-2 takes multiple images rendered from point clouds as inputs and uses a convolutional block to generate feature maps. We substitute the convolutional block with different pre-trained ViTs, unpatchifying the latent vectors concatenated with the global \texttt{[CLS]} token to obtain feature maps. All other architectures and hyperparameters remain consistent with the original RVT-2 implementation. 

\rebuttal{The 35 tasks in Group 1 include: \texttt{Basketball In Hoop}, \texttt{Put Rubbish In Bin}, \texttt{Meat Off Grill}, \texttt{Meat On Grill}, \texttt{Slide Block To Target}, \texttt{Reach And Drag}, \texttt{Take Frame Off Hanger}, \texttt{Water Plants}, \texttt{Hang Frame On Hanger}, \texttt{Wipe Desk}, \texttt{Stack Blocks}, \texttt{Reach Target}, \texttt{Push Button}, \texttt{Lamp On}, \texttt{Toilet Seat Down}, \texttt{Close Laptop Lid}, \texttt{Open Box}, \texttt{Open Drawer}, \texttt{Pick Up Cup}, \texttt{Turn Tap}, \texttt{Take Usb Out Of Computer}, \texttt{Play Jenga}, \texttt{Insert Onto Square Peg}, \texttt{Take Umbrella Out Of Umbrella Stand}, \texttt{Insert Usb In Computer}, \texttt{Straighten Rope}, \texttt{Turn Oven On}, \texttt{Change Clock}, \texttt{Close Microwave}, \texttt{Close Fridge}, \texttt{Close Grill}, \texttt{Open Grill}, \texttt{Unplug Charger}, \texttt{Press Switch}, and \texttt{Take Money Out Safe}.}

\rebuttal{The 36 tasks in Group 2 include: \texttt{Change Channel}, \texttt{Tv On}, \texttt{Push Buttons}, \texttt{Stack Wine}, \texttt{Scoop With Spatula}, \texttt{Place Hanger On Rack}, \texttt{Move Hanger}, \texttt{Sweep To Dustpan}, \texttt{Take Plate Off Colored Dish Rack}, \texttt{Screw Nail}, \texttt{Take Shoes Out Of Box}, \texttt{Slide Cabinet Open And Place Cups}, \texttt{Lamp Off}, \texttt{Pick And Lift}, \texttt{Take Lid Off Saucepan}, \texttt{Close Drawer}, \texttt{Close Box}, \texttt{Phone On Base}, \texttt{Toilet Seat Up}, \texttt{Put Books On Bookshelf}, \texttt{Beat The Buzz}, \texttt{Stack Cups}, \texttt{Put Knife On Chopping Board}, \texttt{Place Shape In Shape Sorter}, \texttt{Take Toilet Roll Off Stand}, \texttt{Put Umbrella In Umbrella Stand}, \texttt{Setup Checkers}, \texttt{Open Window}, \texttt{Open Wine Bottle}, \texttt{Open Microwave}, \texttt{Put Money In Safe}, \texttt{Open Door}, \texttt{Close Door}, \texttt{Open Fridge}, \texttt{Open Oven}, and \texttt{Plug Charger In Power Supply}.}

{\noindent \textbf{LIBERO~\citep{liu2024libero}.}} Built upon Robosuite~\citep{zhu2020robosuite}, LIBERO~\citep{liu2024libero} generates a total of 130 language-conditioned tasks across five suites: LIBERO-Spatial, LIBERO-Object, LIBERO-Goal, LIBERO-10, and LIBERO-90. Each suite contains 10 tasks, except for LIBERO-90, which includes 90 tasks. We train a language-conditioned multi-task policy for each suite, adopting the transformer policy provided by LIBERO. The image encoders are modified from default CNNs to frozen pre-trained ViTs, utilizing the \texttt{[CLS]} token for feature extraction. To expedite policy training, we use only 20 demonstrations per task and forgo augmentations, allowing for pre-extraction of all image features during training. After training for 25 epochs, the checkpoints from the 20th and 25th are evaluated with 20 rollouts per task, and the best checkpoint's performance is taken. Finally, the results are averaged on 3 random seeds. 

\rebuttal{The 10 tasks in LIBERO-Spatial are:} \\
\noindent \rebuttal{1. Pick up the black bowl between the plate and the ramekin and place it on the plate.} \\
\noindent \rebuttal{2. Pick up the black bowl next to the ramekin and place it on the plate.} \\
\noindent \rebuttal{3. Pick up the black bowl from table center and place it on the plate.} \\
\noindent \rebuttal{4. Pick up the black bowl on the cookie box and place it on the plate.} \\
\noindent \rebuttal{5. Pick up the black bowl in the top drawer of the wooden cabinet and place it on the plate.} \\
\noindent \rebuttal{6. Pick up the black bowl on the ramekin and place it on the plate.} \\
\noindent \rebuttal{7. Pick up the black bowl next to the cookie box and place it on the plate.} \\
\noindent \rebuttal{8. Pick up the black bowl on the stove and place it on the plate.} \\
\noindent \rebuttal{9. Pick up the black bowl next to the plate and place it on the plate.} \\
\noindent \rebuttal{10. Pick up the black bowl on the wooden cabinet and place it on the plate.}

\rebuttal{The 10 tasks in LIBERO-Object are:} \\
\noindent \rebuttal{1. Pick up the alphabet soup and place it in the basket.} \\
\noindent \rebuttal{2. Pick up the cream cheese and place it in the basket.} \\
\noindent \rebuttal{3. Pick up the salad dressing and place it in the basket.} \\
\noindent \rebuttal{4. Pick up the BBQ sauce and place it in the basket.} \\
\noindent \rebuttal{5. Pick up the ketchup and place it in the basket.} \\
\noindent \rebuttal{6. Pick up the tomato sauce and place it in the basket.} \\
\noindent \rebuttal{7. Pick up the butter and place it in the basket.} \\
\noindent \rebuttal{8. Pick up the milk and place it in the basket.} \\
\noindent \rebuttal{9. Pick up the chocolate pudding and place it in the basket.} \\
\noindent \rebuttal{10. Pick up the orange juice and place it in the basket.}

\rebuttal{The 10 tasks in LIBERO-Goal are:} \\
\noindent \rebuttal{1. Open the middle drawer of the cabinet.} \\
\noindent \rebuttal{2. Put the bowl on the stove.} \\
\noindent \rebuttal{3. Put the wine bottle on top of the cabinet.} \\
\noindent \rebuttal{4. Open the top drawer and put the bowl inside.} \\
\noindent \rebuttal{5. Put the bowl on top of the cabinet.} \\
\noindent \rebuttal{6. Push the plate to the front of the stove.} \\
\noindent \rebuttal{7. Put the cream cheese in the bowl.} \\
\noindent \rebuttal{8. Turn on the stove.} \\
\noindent \rebuttal{9. Put the bowl on the plate.} \\
\noindent \rebuttal{10. Put the wine bottle on the rack.}

\rebuttal{The 10 tasks in LIBERO-10 are:} \\
\noindent \rebuttal{1. LIVING ROOM SCENE2: Put both the alphabet soup and the tomato sauce in the basket.} \\
\noindent \rebuttal{2. LIVING ROOM SCENE2: Put both the cream cheese box and the butter in the basket.} \\
\noindent \rebuttal{3. KITCHEN SCENE3: Turn on the stove and put the moka pot on it.} \\
\noindent \rebuttal{4. KITCHEN SCENE4: Put the black bowl in the bottom drawer of the cabinet and close it.} \\
\noindent \rebuttal{5. LIVING ROOM SCENE5: Put the white mug on the left plate and put the yellow and white mug on the right plate.} \\
\noindent \rebuttal{6. STUDY SCENE1: Pick up the book and place it in the back compartment of the caddy.} \\
\noindent \rebuttal{7. LIVING ROOM SCENE6: Put the white mug on the plate and put the chocolate pudding to the right of the plate.} \\
\noindent \rebuttal{8. LIVING ROOM SCENE1: Put both the alphabet soup and the cream cheese box in the basket.} \\
\noindent \rebuttal{9. KITCHEN SCENE8: Put both moka pots on the stove.} \\
\noindent \rebuttal{10. KITCHEN SCENE6: Put the yellow and white mug in the microwave and close it.}

\rebuttal{The 10 tasks in LIBERO-90 are:} \\
\noindent \rebuttal{1. KITCHEN SCENE10: Close the top drawer of the cabinet.} \\
\noindent \rebuttal{2. KITCHEN SCENE10: Close the top drawer of the cabinet and put the black bowl on top of it.} \\
\noindent \rebuttal{3. KITCHEN SCENE10: Put the black bowl in the top drawer of the cabinet.} \\
\noindent \rebuttal{4. KITCHEN SCENE10: Put the butter at the back in the top drawer of the cabinet and close it.} \\
\noindent \rebuttal{5. KITCHEN SCENE10: Put the butter at the front in the top drawer of the cabinet and close it.} \\
\noindent \rebuttal{6. KITCHEN SCENE10: Put the chocolate pudding in the top drawer of the cabinet and close it.} \\
\noindent \rebuttal{7. KITCHEN SCENE1: Open the bottom drawer of the cabinet.} \\
\noindent \rebuttal{8. KITCHEN SCENE1: Open the top drawer of the cabinet.} \\
\noindent \rebuttal{9. KITCHEN SCENE1: Open the top drawer of the cabinet and put the bowl in it.} \\
\noindent \rebuttal{10. KITCHEN SCENE1: Put the black bowl on the plate.} \\
\noindent \rebuttal{11. KITCHEN SCENE1: Put the black bowl on top of the cabinet.} \\
\noindent \rebuttal{12. KITCHEN SCENE2: Open the top drawer of the cabinet.} \\
\noindent \rebuttal{13. KITCHEN SCENE2: Put the black bowl at the back on the plate.} \\
\noindent \rebuttal{14. KITCHEN SCENE2: Put the black bowl at the front on the plate.} \\
\noindent \rebuttal{15. KITCHEN SCENE2: Put the middle black bowl on the plate.} \\
\noindent \rebuttal{16. KITCHEN SCENE2: Put the middle black bowl on top of the cabinet.} \\
\noindent \rebuttal{17. KITCHEN SCENE2: Stack the black bowl at the front on the black bowl in the middle.} \\
\noindent \rebuttal{18. KITCHEN SCENE2: Stack the middle black bowl on the back black bowl.} \\
\noindent \rebuttal{19. KITCHEN SCENE3: Put the frying pan on the stove.} \\
\noindent \rebuttal{20. KITCHEN SCENE3: Put the moka pot on the stove.} \\
\noindent \rebuttal{21. KITCHEN SCENE3: Turn on the stove.} \\
\noindent \rebuttal{22. KITCHEN SCENE3: Turn on the stove and put the frying pan on it.} \\
\noindent \rebuttal{23. KITCHEN SCENE4: Close the bottom drawer of the cabinet.} \\
\noindent \rebuttal{24. KITCHEN SCENE4: Close the bottom drawer of the cabinet and open the top drawer.} \\
\noindent \rebuttal{25. KITCHEN SCENE4: Put the black bowl in the bottom drawer of the cabinet.} \\
\noindent \rebuttal{26. KITCHEN SCENE4: Put the black bowl on top of the cabinet.} \\
\noindent \rebuttal{27. KITCHEN SCENE4: Put the wine bottle in the bottom drawer of the cabinet.} \\
\noindent \rebuttal{28. KITCHEN SCENE4: Put the wine bottle on the wine rack.} \\
\noindent \rebuttal{29. KITCHEN SCENE5: Close the top drawer of the cabinet.} \\
\noindent \rebuttal{30. KITCHEN SCENE5: Put the black bowl in the top drawer of the cabinet.} \\
\noindent \rebuttal{31. KITCHEN SCENE5: Put the black bowl on the plate.} \\
\noindent \rebuttal{32. KITCHEN SCENE5: Put the black bowl on top of the cabinet.} \\
\noindent \rebuttal{33. KITCHEN SCENE5: Put the ketchup in the top drawer of the cabinet.} \\
\noindent \rebuttal{34. KITCHEN SCENE6: Close the microwave.} \\
\noindent \rebuttal{35. KITCHEN SCENE6: Put the yellow and white mug to the front of the white mug.} \\
\noindent \rebuttal{36. KITCHEN SCENE7: Open the microwave.} \\
\noindent \rebuttal{37. KITCHEN SCENE7: Put the white bowl on the plate.} \\
\noindent \rebuttal{38. KITCHEN SCENE7: Put the white bowl to the right of the plate.} \\
\noindent \rebuttal{39. KITCHEN SCENE8: Put the right moka pot on the stove.} \\
\noindent \rebuttal{40. KITCHEN SCENE8: Turn off the stove.} \\
\noindent \rebuttal{41. KITCHEN SCENE9: Put the frying pan on the cabinet shelf.} \\
\noindent \rebuttal{42. KITCHEN SCENE9: Put the frying pan on top of the cabinet.} \\
\noindent \rebuttal{43. KITCHEN SCENE9: Put the frying pan under the cabinet shelf.} \\
\noindent \rebuttal{44. KITCHEN SCENE9: Put the white bowl on top of the cabinet.} \\
\noindent \rebuttal{45. KITCHEN SCENE9: Turn on the stove.} \\
\noindent \rebuttal{46. KITCHEN SCENE9: Turn on the stove and put the frying pan on it.} \\
\noindent \rebuttal{47. LIVING ROOM SCENE1: Pick up the alphabet soup and put it in the basket.} \\
\noindent \rebuttal{48. LIVING ROOM SCENE1: Pick up the cream cheese box and put it in the basket.} \\
\noindent \rebuttal{49. LIVING ROOM SCENE1: Pick up the ketchup and put it in the basket.} \\
\noindent \rebuttal{50. LIVING ROOM SCENE1: Pick up the tomato sauce and put it in the basket.} \\
\noindent \rebuttal{51. LIVING ROOM SCENE2: Pick up the alphabet soup and put it in the basket.} \\
\noindent \rebuttal{52. LIVING ROOM SCENE2: Pick up the butter and put it in the basket.} \\
\noindent \rebuttal{53. LIVING ROOM SCENE2: Pick up the milk and put it in the basket.} \\
\noindent \rebuttal{54. LIVING ROOM SCENE2: Pick up the orange juice and put it in the basket.} \\
\noindent \rebuttal{55. LIVING ROOM SCENE2: Pick up the tomato sauce and put it in the basket.} \\
\noindent \rebuttal{56. LIVING ROOM SCENE3: Pick up the alphabet soup and put it in the tray.} \\
\noindent \rebuttal{57. LIVING ROOM SCENE3: Pick up the butter and put it in the tray.} \\
\noindent \rebuttal{58. LIVING ROOM SCENE3: Pick up the cream cheese and put it in the tray.} \\
\noindent \rebuttal{59. LIVING ROOM SCENE3: Pick up the ketchup and put it in the tray.} \\
\noindent \rebuttal{60. LIVING ROOM SCENE3: Pick up the tomato sauce and put it in the tray.} \\
\noindent \rebuttal{61. LIVING ROOM SCENE4: Pick up the black bowl on the left and put it in the tray.} \\
\noindent \rebuttal{62. LIVING ROOM SCENE4: Pick up the chocolate pudding and put it in the tray.} \\
\noindent \rebuttal{63. LIVING ROOM SCENE4: Pick up the salad dressing and put it in the tray.} \\
\noindent \rebuttal{64. LIVING ROOM SCENE4: Stack the left bowl on the right bowl and place them in the tray.} \\
\noindent \rebuttal{65. LIVING ROOM SCENE4: Stack the right bowl on the left bowl and place them in the tray.} \\
\noindent \rebuttal{66. LIVING ROOM SCENE5: Put the red mug on the left plate.} \\
\noindent \rebuttal{67. LIVING ROOM SCENE5: Put the red mug on the right plate.} \\
\noindent \rebuttal{68. LIVING ROOM SCENE5: Put the white mug on the left plate.} \\
\noindent \rebuttal{69. LIVING ROOM SCENE5: Put the yellow and white mug on the right plate.} \\
\noindent \rebuttal{70. LIVING ROOM SCENE6: Put the chocolate pudding to the left of the plate.} \\
\noindent \rebuttal{71. LIVING ROOM SCENE6: Put the chocolate pudding to the right of the plate.} \\
\noindent \rebuttal{72. LIVING ROOM SCENE6: Put the red mug on the plate.} \\
\noindent \rebuttal{73. LIVING ROOM SCENE6: Put the white mug on the plate.} \\
\noindent \rebuttal{74. STUDY SCENE1: Pick up the book and place it in the front compartment of the caddy.} \\
\noindent \rebuttal{75. STUDY SCENE1: Pick up the book and place it in the left compartment of the caddy.} \\
\noindent \rebuttal{76. STUDY SCENE1: Pick up the book and place it in the right compartment of the caddy.} \\
\noindent \rebuttal{77. STUDY SCENE1: Pick up the yellow and white mug and place it to the right of the caddy.} \\
\noindent \rebuttal{78. STUDY SCENE2: Pick up the book and place it in the back compartment of the caddy.} \\
\noindent \rebuttal{79. STUDY SCENE2: Pick up the book and place it in the front compartment of the caddy.} \\
\noindent \rebuttal{80. STUDY SCENE2: Pick up the book and place it in the left compartment of the caddy.} \\
\noindent \rebuttal{81. STUDY SCENE2: Pick up the book and place it in the right compartment of the caddy.} \\
\noindent \rebuttal{82. STUDY SCENE3: Pick up the book and place it in the front compartment of the caddy.} \\
\noindent \rebuttal{83. STUDY SCENE3: Pick up the book and place it in the left compartment of the caddy.} \\
\noindent \rebuttal{84. STUDY SCENE3: Pick up the book and place it in the right compartment of the caddy.} \\
\noindent \rebuttal{85. STUDY SCENE3: Pick up the red mug and place it to the right of the caddy.} \\
\noindent \rebuttal{86. STUDY SCENE3: Pick up the white mug and place it to the right of the caddy.} \\
\noindent \rebuttal{87. STUDY SCENE4: Pick up the book in the middle and place it on the cabinet shelf.} \\
\noindent \rebuttal{88. STUDY SCENE4: Pick up the book on the left and place it on top of the shelf.} \\
\noindent \rebuttal{89. STUDY SCENE4: Pick up the book on the right and place it on the cabinet shelf.} \\
\noindent \rebuttal{90. STUDY SCENE4: Pick up the book on the right and place it under the cabinet shelf.}

\section{More Implementation Details}
\label{appx:impl-details}
\subsection{Dataset Details}
The datasets used for SPA include ScanNet, ScanNet++, Hypersim, ADT, S3DIS, and Droid. 

\noindent \textbf{ScanNet} consists of 1.89 million frames in total. Each epoch includes 1.5 times the dataset size. For each scene, a random starting frame is selected, followed by the sampling of 1 to 8 frames at random, with an interval of 8 frames between them.

\noindent \textbf{ScanNet++} comprises 0.11 million frames. Each epoch includes 5 times the dataset size. For each scene, a random starting frame is selected, followed by the sampling of 1 to 8 frames at random, with an interval of 5 frames between them.

\noindent \textbf{Hypersim} contains 0.03 million frames. Each epoch includes 8 times the dataset size. For each scene, we randomly select 1 to 8 continuous frames.

\noindent \textbf{ADT} consists of 0.0015M frames in total. Each epoch includes 1 times the dataset size. For each scene, 1 to 8 continuous frames are randomly selected.

\noindent \textbf{S3DIS} consists of 0.015 million frames. Each epoch includes 5 times the dataset size. For each scene, a random starting frame is selected, followed by the sampling of 1 to 8 frames at random, with an interval of 5 frames between them.

\noindent \textbf{Droid} contains a large number of videos, but due to the high similarity between frames, the videos are first downsampled by a factor of 15 during pre-processing, resulting in 1.78 million frames. Since Droid does not provide depth data, we utilize Croco-Stereo~\cite{croco_v2} to estimate dense depth maps for rendering supervision. Additionally, due to the significant noise in the camera pose data, only a single frame is sampled at a time during training.

During pre-training, we first resize the multi-view input images to slightly larger than \(224 \times 224\), and then randomly crop them to a final size of \(224 \times 224\). Random photometric distortions with a probability of $0.5$ are applied for augmentation, including brightness ranging from 0.875 to 1.125, contrast ranging from 0.5 to 1.5, saturation ranging from 0.5 to 1.5, and hue ranging from -0.05 to 0.05. Frames with very small valid depth areas or scene boxes are filtered out. 

For semantic rendering supervision, we observe that using larger image sizes improves the quality of feature maps generated by RADIO. Consequently, we resize the images to \(1024 \times 1024\) before feeding them into RADIO, which outputs a feature map of size \(64 \times 64\). We then apply bilinear sampling to query the semantic feature labels for each pixel.

\subsection{Pre-Training Details}

For stability during pre-training, we apply the Exponential Moving Average (EMA) with a decay rate of 0.999. The model is trained for 2000 epochs on 80 NVIDIA A100-80G GPUs, using a gradient clipping threshold of 1.0. Each GPU processes a batch size of 2, with 8 gradient accumulation steps, resulting in a total effective batch size of \(2 \times 80 \times 8 = 1280\). We employ the AdamW optimizer with a weight decay of 0.04. The base learning rate is set to \(5 \times 10^{-6}\), and the actual learning rate is scaled by a factor of 8 times the effective batch size. A OneCycle learning rate scheduler is used, with a percentage start of 0.05, a divide factor of 100, and a final divide factor of 1000.

To facilitate faster convergence and improve stability, we initialize the encoder with ImageNet pre-trained weights from the Masked Autoencoder (MAE), applying a learning rate layer decay of 0.8. This initialization does not affect the validity of our conclusions, as demonstrated by the ablation study of SPA-MAE in \secref{sec:study-on-3d-awareness}. The ViT encoder and upsampling layers are trained with FP16 precision, while the volume decoder is trained with FP32 precision.

We set the loss weights to \(\lambda_{\text{color}} = 10\), \(\lambda_{\text{depth}} = 1\), \(\lambda_{\text{semantic}} = 1\), \(\lambda_{\text{eikonal}} = 0.01\), \(\lambda_{\text{free}} = 1\), and \(\lambda_{\text{sdf}} = 10\). For the NeuS sampler, the initial number of samples is set to 72, with 24 importance samples. In each iteration, we randomly sample \(512\) pixels per view for rendering and supervision.

\section{Detailed Results of Each Task}
\label{appx:detailed-results}
We present the results of all individual tasks in \tabref{tab:appx-vc1}, \tabref{tab:appx-fk}, \tabref{tab:appx-mw}, \tabref{tab:appx-rlbench}, \tabref{tab:appx-libero}, and \tabref{tab:appx-libero90}.

\section{Camera Pose Estimation Details}
\label{appx:camera-pose}
We adopt a setup similar to that of \cite{el2024probing} for camera pose estimation using the NAVI dataset~\citep{jampani2023navi}. Given an image pair from different viewpoints, we first extract features from each image using a frozen, pre-trained Vision Transformer (ViT) encoder. Following standard protocols for embodied evaluation, we use the \texttt{[CLS]} token as the feature representation. The two \texttt{[CLS]} tokens are then concatenated and passed through a BatchNorm layer and a Multi-Layer Perceptron (MLP) to regress the camera pose. The MLP consists of four linear layers with three ReLU activations, using hidden sizes of 512, 256, and 128 units, and outputs a 7-dimensional pose vector. The first three dimensions represent the \(xyz\) translation, while the last four dimensions correspond to the rotation quaternions. 

We employ the Mean Squared Error (MSE) loss function and optimize the model using the AdamW optimizer with a OneCycle learning rate scheduler. The model is trained for 100 epochs with a base learning rate of \(1 \times 10^{-3}\) and a starting percentage of 0.1. For evaluation, we use Euclidean distance as the translation error metric and geodesic distance as the rotation error metric. The geodesic distance between two quaternions \(q_1\) and \(q_2\) is defined as:

\begin{equation}
\theta = 2 \cdot \arccos\left( \left| q_1 \cdot q_2 \right| \right),
\end{equation}
where \(q_1\) and \(q_2\) are normalized quaternions, and \(\cdot\) denotes the quaternion dot product.
The Euclidean distance \(d\) between two translation vectors \(\mathbf{t}_1 = (x_1, y_1, z_1)\) and \(\mathbf{t}_2 = (x_2, y_2, z_2)\) is given by:

\begin{equation}
d = \sqrt{(x_2 - x_1)^2 + (y_2 - y_1)^2 + (z_2 - z_1)^2}.
\end{equation}

\section{Real-World Experiment Details}
\label{appx:real-world}
\begin{figure}[!tb]
\centering
\includegraphics[width=0.65\linewidth]{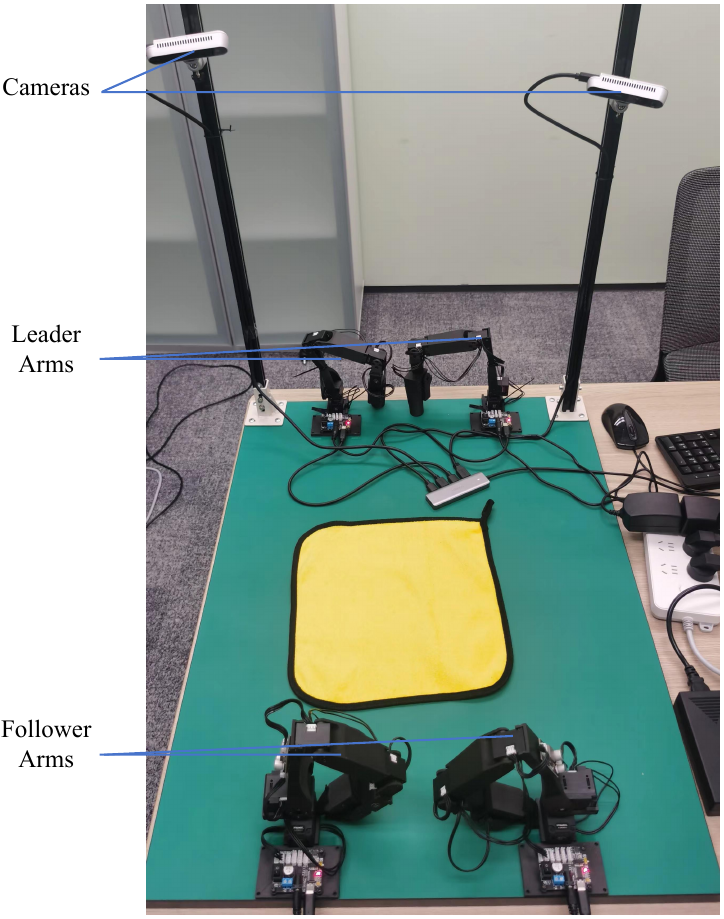}
\vspace{-0.9em}
\caption{\textbf{Real-world hardware platform.}}
\label{fig:platform}
\vspace{-1.8em}
\end{figure}

Our real-world hardware setup is based on the open-source Low-Cost-Robot project~\citep{Alexander}. We utilize two Intel RealSense D415 cameras for image capture. A visualization of our platform is provided in \figref{fig:platform}. For teleoperation, policy training, and evaluation, we leverage the open-source RealRobot project~\citep{realrobot2024}. The policy used is the ACT policy~\citep{zhao2023learning}. 

For each task, we collect 50 demonstrations, and during evaluation, we conduct 25 rollouts, each with randomized object locations and orientations. The model is trained for 10,000 epochs using four NVIDIA A100 GPUs. We employ the AdamW optimizer with a learning rate of \(5 \times 10^{-5}\) and a weight decay of 0.05. Additionally, a OneCycle learning rate scheduler is used, with a starting percentage of 0.1, a division factor of 10, and a final division factor of 100.

\section{\rebuttal{Additional Reinforcement Learning Experiments}}

\begin{figure}[!htb]
\centering
\includegraphics[width=\linewidth]{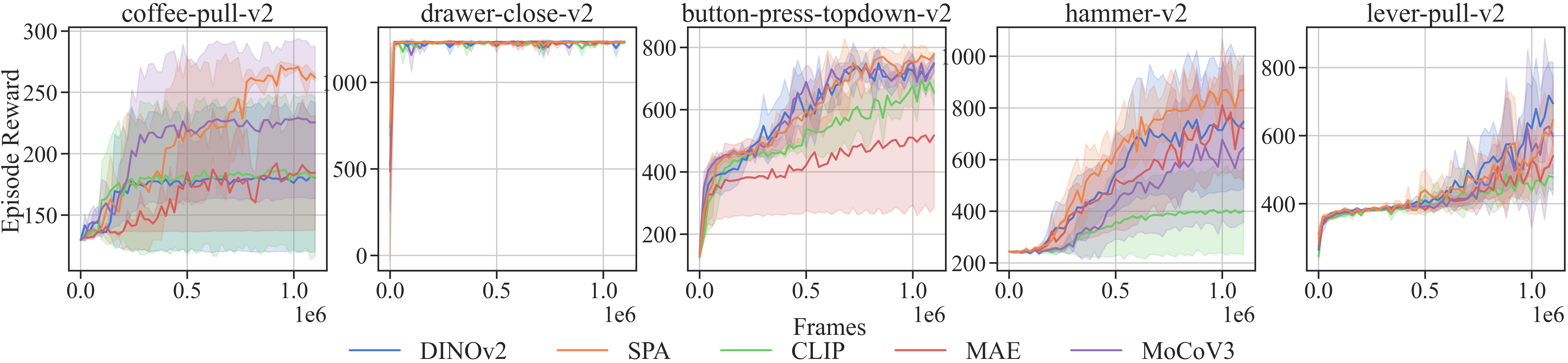}
\vspace{-0.9em}
\caption{\rebuttal{\textbf{Reinforcement learning reward curves visualization.}}}
\label{fig:appx-rl}
\vspace{-1.8em}
\end{figure}

\rebuttal{We conduct additional RL experiments following the settings in \citet{hu2023pre} to use DrQ-v2~\citep{yarats2021mastering}, a state-of-the-art off-policy actor-critic approach for continuous vision-based control. We train some RL experiments with different pre-trained vision representations with ViT-Base architectures. The vision encoders are frozen during RL training. Five tasks in the Meta-World benchmark are chosen, as shown below. We train for a total of 1.1M frames and all other hyper-parameters including random seeds are kept as default and same. We run three seeds for each experiment. We report the evaluation success rate and episode reward below in \tabref{tab:appx-rl}. The reward curves are visualized in \figref{fig:appx-rl}. From the results, it is evident that the reinforcement learning outcomes exhibit high variance. Nevertheless, overall, our 3D spatial-aware representation outperforms other representation learning methods.} %

\begin{table}[!hbt]\centering
\caption{\rebuttal{\textbf{Reinforcement learning comparison results.}}}\label{tab:appx-rl}
\vspace{-0.75em}
\tablestyle{6pt}{0.9}
\begin{tabular}{l|l|ccc}\toprule
Meta-World RL Task &Method (ViT-B) &Success Rate &Episode Reward \\\cmidrule{1-4}
\multirow{5}{*}{button-press-topdown-v2} &CLIP &0.93 &653.97 \\
&DINOv2 &\textbf{1.00} &746.04 \\
&MAE &0.46 &517.54 \\
&MoCoV3 &0.99 &\underline{749.93} \\
&SPA (Ours) &\textbf{1.00} &\textbf{778.47} \\\cmidrule{1-4}
\multirow{5}{*}{hammer-v2} &CLIP &0.00 &401.41 \\
&DINOv2 &\underline{0.67} &\underline{746.74} \\
&MAE &0.66 &720.19 \\
&MoCoV3 &0.59 &645.46 \\
&SPA (Ours) &\textbf{1.00} &\textbf{870.32} \\\cmidrule{1-4}
\multirow{5}{*}{lever-pull-v2} &CLIP &0.00 &478.18 \\
&DINOv2 &0.00 &\textbf{694.73} \\
&MAE &0.00 &540.44 \\
&MoCoV3 &\textbf{0.23} &598.54 \\
&SPA (Ours) &\underline{0.15} &\underline{646.33} \\\cmidrule{1-4}
\multirow{5}{*}{coffee-pull-v2} &CLIP &0.00 &181.40 \\
&DINOv2 &0.00 &180.72 \\
&MAE &0.00 &184.56 \\
&MoCoV3 &0.00 &\underline{225.73} \\
&SPA (Ours) &\textbf{0.00} &\textbf{262.11} \\\cmidrule{1-4}
\multirow{5}{*}{drawer-close-v2} &CLIP &1.00 &1228.90 \\
&DINOv2 &1.00 &\textbf{1236.30} \\
&MAE &1.00 &1233.91 \\
&MoCoV3 &1.00 &1233.46 \\
&SPA (Ours) &\textbf{1.00} &\underline{1235.81} \\\cmidrule{1-4}
\multirow{5}{*}{Mean} &CLIP &0.39 &588.77 \\
&DINOv2 &0.53 &\underline{720.91} \\
&MAE &0.42 &639.33 \\
&MoCoV3 &\underline{0.56} &690.63 \\
&SPA (Ours) &\textbf{0.63} &\textbf{758.61} \\
\bottomrule
\end{tabular}
\end{table}

\section{\rebuttal{Additional Monocular Grasp Pose Detection Experiments}}
\rebuttal{We conduct a monocular grasp pose detection experiment to further investigate more complex robotics learning paradigms. We follow similar settings in \citet{gou2021rgb}, which train a neural network to detect the 7-DoF grasp poses on monocular image observations. The experiment is conducted on GraspNet-1Billion~\citep{fang2020graspnet}, a large-scale real-world object grasping benchmark. We follow the hyper-parameters and setups in the official implementation, except that we replace the default ResNet with different pre-trained ViT models for feature extraction. All pre-trained representations are with ViT-Base architecture and are frozen during training. We report the overall Top-K accuracy on the test set below. The results align well with our findings and indicate that SPA also outperforms other representation learning methods in the monocular grasp pose detection task.}

\begin{table}[!hbt]\centering
\vspace{0.25em}
\caption{\textbf{\rebuttal{Overall top-k monocular grasp pose detection accuracy of various methods (ViT-Base).}}}\label{tab:overall-grasp-accuracy}
\tablestyle{12pt}{0.9}
\begin{tabular}{l|cccccc}\toprule
Method (ViT-Base) & CLIP & DINOv2 & MoCoV3 & MAE & SPA \\\cmidrule{1-6}
Overall Accuracy & 21.10 & 22.08 & 29.39 & 31.03 & \textbf{31.20} \\
\bottomrule
\end{tabular}
\end{table}

\section{\rebuttal{Additional Ablation Study on Neural Rendering}}

\rebuttal{To clarify the contribution of neural rendering to the overall performance of SPA, we conducted an additional ablation study. In this study, we maintained all settings identical—data loading, training techniques, hyperparameters, and the encoder—while replacing the volume neural rendering decoder with a multiview transformer-based decoder, similar to the MAE decoder. This alternative decoder receives masked patches filled with mask tokens corresponding to multiview images. Additional camera pose embeddings are added, and attention layers are used to fuse the multiview information and reconstruct RGB and depth images. We refer to this baseline as MV-MAE. It was trained on the ScanNet dataset without semantic supervision, ensuring a fair comparison with the result in the last line of \tabref{tab:mr_loss}. The results from this experiment demonstrate that neural rendering is crucial for incorporating explicit 3D spatial information. Simple multiview attention-based interaction, as used in MV-MAE, does not perform as effectively in learning 3D spatial awareness.}

\begin{table}[!hbt]\centering
\vspace{0.25em}
\caption{\rebuttal{\textbf{Additional Ablation Study on Neural Rendering.} The models are evaluated on two subsets of the VC-1 benchmark. The model architectures are both ViT-base.}}\label{tab:abl-neural-rendering}
\tablestyle{12pt}{0.9}
\begin{tabular}{l|cc}\toprule
Method & Meta-World & DMControl \\\cmidrule{1-3}
MV-MAE & 84.8$\pm$5.8 & 59.6$\pm$3.2 \\
SPA & \textbf{88.0$\pm$4.5} & \textbf{61.5$\pm$3.4} \\
\bottomrule
\end{tabular}
\end{table}

\begin{table}[!htp]\centering
\caption{\textbf{All results on Vc-1 benchmarks.}}\label{tab:appx-vc1}
\vspace{-1.2em}
\resizebox{\linewidth}{!}{
\tablestyle{1.25pt}{0.95}
\begin{tabular}{l|l|cc|ccccc|ccccc|ccc}\toprule
\multicolumn{2}{r|}{Benchmark} &\multicolumn{2}{c|}{AD} &\multicolumn{5}{c|}{MW} &\multicolumn{5}{c|}{DMC} &\multicolumn{2}{c}{TF} \\\cmidrule{1-16}
Methods &Seed &\makecell{Relo-\\cate} &Pen &\makecell{Button\\Press\\Topdown} &\makecell{Drawer\\Open} &\makecell{Bin\\Picking} &\makecell{Ham-\\mer} &\makecell{Assem-\\bly} &\makecell{Walker\\Stand} &\makecell{Walker\\Walk} &\makecell{Reacher\\Easy} &\makecell{Cheetah\\Run} &\makecell{Finger\\Spin} &\makecell{Reach\\Cube} &\makecell{Move\\Cube} \\\cmidrule{1-16}
\multicolumn{16}{l}{\textit{ViT-L Methods}} \\\cmidrule{1-16}
\multirow{3}{*}{MoCoV3} &100 &40.00 &92.00 &88.00 &100.00 &88.00 &100.00 &88.00 &84.88 &57.59 &92.29 &56.28 &70.49 &84.37 &61.20 \\
&200 &36.00 &80.00 &88.00 &100.00 &80.00 &100.00 &84.00 &82.95 &55.02 &92.08 &43.17 &69.49 &84.20 &61.26 \\
&300 &28.00 &76.00 &84.00 &100.00 &68.00 &92.00 &72.00 &81.42 &53.59 &91.96 &41.27 &68.23 &84.09 &64.24 \\\cmidrule{1-16}
\multirow{3}{*}{MAE} &100 &36.00 &84.00 &84.00 &100.00 &88.00 &100.00 &100.00 &951.27 &680.69 &976.50 &482.47 &703.30 &85.46 &59.46 \\
&200 &36.00 &80.00 &84.00 &100.00 &76.00 &96.00 &96.00 &933.67 &676.92 &952.20 &49.22 &695.00 &86.88 &59.45 \\
&300 &32.00 &80.00 &68.00 &100.00 &72.00 &98.00 &88.00 &873.53 &659.41 &895.60 &501.91 &691.80 &85.26 &61.69 \\\cmidrule{1-16}
\multirow{3}{*}{DINOV2} &100 &32.00 &68.00 &68.00 &100.00 &84.00 &100.00 &88.00 &87.01 &56.52 &94.50 &26.98 &70.87 &86.16 &50.84 \\
&200 &28.00 &68.00 &60.00 &100.00 &80.00 &96.00 &80.00 &86.00 &53.97 &89.97 &21.84 &68.78 &86.87 &50.78 \\
&300 &28.00 &60.00 &60.00 &100.00 &80.00 &92.00 &72.00 &82.41 &51.69 &88.36 &21.34 &67.41 &86.05 &50.17 \\\cmidrule{1-16}
\multirow{3}{*}{CLIP} &100 &24.00 &80.00 &28.00 &100.00 &88.00 &100.00 &84.00 &66.16 &43.94 &90.71 &18.40 &68.30 &73.28 &41.09 \\
&200 &24.00 &72.00 &24.00 &100.00 &84.00 &100.00 &80.00 &64.04 &34.51 &88.26 &16.52 &66.68 &75.11 &33.45 \\
&300 &24.00 &68.00 &16.00 &100.00 &84.00 &96.00 &72.00 &52.70 &31.60 &85.17 &14.51 &67.49 &74.73 &38.95 \\\cmidrule{1-16}
\multirow{3}{*}{EVA} &100 &44.00 &84.00 &72.00 &100.00 &76.00 &100.00 &100.00 &77.92 &51.64 &98.17 &31.04 &70.17 &82.56 &52.04 \\
&200 &40.00 &76.00 &84.00 &100.00 &72.00 &100.00 &100.00 &77.63 &50.81 &86.66 &19.37 &67.43 &81.72 &52.07 \\
&300 &32.00 &72.00 &96.00 &100.00 &68.00 &96.00 &96.00 &77.21 &47.71 &88.41 &29.19 &67.43 &82.13 &52.70 \\\cmidrule{1-16}
\multirow{3}{*}{\makecell{InternViT-\\300M}} &100 &40.00 &72.00 &80.00 &100.00 &72.00 &100.00 &84.00 &70.04 &30.44 &80.80 &16.70 &67.05 &78.59 &53.67 \\
&200 &28.00 &72.00 &68.00 &100.00 &76.00 &96.00 &84.00 &67.63 &31.55 &82.33 &19.39 &67.57 &77.07 &55.21 \\
&300 &28.00 &80.00 &60.00 &100.00 &72.00 &96.00 &72.00 &66.95 &29.28 &81.87 &18.80 &68.55 &78.27 &48.58 \\\cmidrule{1-16}
\multirow{3}{*}{\makecell{InternViT-\\6B}} &100 &32.00 &72.00 &88.00 &100.00 &80.00 &100.00 &84.00 &88.53 &70.02 &93.09 &26.54 &70.62 &85.96 &57.52 \\
&200 &40.00 &76.00 &84.00 &100.00 &76.00 &100.00 &80.00 &85.28 &60.09 &90.86 &22.84 &69.04 &86.30 &54.11 \\
&300 &60.00 &80.00 &80.00 &100.00 &88.00 &100.00 &76.00 &81.88 &59.17 &87.87 &21.53 &67.20 &85.68 &54.86 \\\cmidrule{1-16}
\multirow{3}{*}{MVP} &100 &32.00 &84.00 &96.00 &100.00 &96.00 &100.00 &100.00 &84.88 &57.59 &92.29 &56.28 &70.49 &84.37 &61.20 \\
&200 &28.00 &76.00 &92.00 &100.00 &84.00 &100.00 &96.00 &82.95 &55.02 &92.08 &43.17 &69.49 &84.20 &61.26 \\
&300 &24.00 &76.00 &84.00 &100.00 &68.00 &100.00 &88.00 &81.42 &53.59 &91.96 &41.27 &68.23 &84.09 &64.24 \\\cmidrule{1-16}
\multirow{3}{*}{VC-1} &100 &32.00 &84.00 &84.00 &100.00 &76.00 &96.00 &96.00 &82.36 &55.33 &98.09 &35.31 &72.60 &83.36 &58.00 \\
&200 &28.00 &80.00 &68.00 &100.00 &72.00 &92.00 &84.00 &80.21 &53.90 &89.83 &34.10 &70.15 &83.17 &61.00 \\
&300 &24.00 &76.00 &76.00 &100.00 &96.00 &88.00 &84.00 &68.62 &50.13 &87.89 &31.18 &70.11 &82.75 &57.16 \\\cmidrule{1-16}
\multirow{3}{*}{SPA-L} &100 &40.00 &88.00 &76.00 &100.00 &92.00 &100.00 &100.00 &94.19 &66.34 &95.57 &52.53 &73.95 &87.37 &56.68 \\
&200 &44.00 &76.00 &84.00 &100.00 &88.00 &100.00 &84.00 &92.28 &60.60 &81.43 &44.99 &71.83 &87.26 &64.35 \\
&300 &36.00 &76.00 &96.00 &100.00 &88.00 &96.00 &96.00 &87.87 &51.75 &83.86 &39.10 &70.91 &87.62 &58.02 \\\cmidrule{1-16}
\multicolumn{16}{l}{\textit{ViT-B Methods and Others}} \\\cmidrule{1-16}
\multirow{3}{*}{STP-B} &100 &20.00 &80.00 &88.00 &100.00 &84.00 &100.00 &96.00 &77.02 &45.34 &87.97 &40.01 &72.72 &80.41 &54.66 \\
&200 &28.00 &76.00 &92.00 &100.00 &84.00 &100.00 &80.00 &71.50 &33.60 &84.08 &34.30 &72.18 &80.13 &57.97 \\
&300 &32.00 &76.00 &88.00 &100.00 &72.00 &100.00 &96.00 &71.44 &42.86 &79.67 &39.17 &69.12 &80.65 &53.95 \\\cmidrule{1-16}
\multirow{3}{*}{R3M-B} &100 &20.00 &92.00 &52.00 &96.00 &32.00 &88.00 &48.00 &668.49 &301.54 &842.90 &256.56 &678.00 &75.08 &45.66 \\
&200 &12.00 &76.00 &48.00 &96.00 &32.00 &88.00 &44.00 &634.62 &256.82 &661.40 &198.63 &660.90 &75..62 &48.09 \\
&300 &12.00 &76.00 &32.00 &88.00 &28.00 &76.00 &40.00 &633.39 &211.90 &585.50 &188.15 &657.30 &74.54 &45.59 \\\cmidrule{1-16}
\multirow{3}{*}{Theia-B} &100 &32.00 &76.00 &88.00 &100.00 &80.00 &96.00 &96.00 &72.90 &43.97 &82.09 &37.02 &70.50 &84.55 &55.62 \\
&200 &36.00 &80.00 &60.00 &100.00 &84.00 &100.00 &84.00 &79.05 &56.99 &94.36 &39.59 &70.22 &83.27 &54.59 \\
&300 &24.00 &72.00 &80.00 &100.00 &72.00 &96.00 &100.00 &79.64 &54.39 &82.89 &39.09 &72.00 &84.01 &54.43 \\\cmidrule{1-16}
\multirow{3}{*}{Voltron-B} &100 &16.00 &72.00 &76.00 &100.00 &64.00 &100.00 &96.00 &74.31 &42.05 &68.88 &36.94 &70.91 &86.28 &65.11 \\
&200 &32.00 &72.00 &76.00 &100.00 &60.00 &96.00 &88.00 &71.57 &38.17 &67.53 &31.01 &70.17 &86.61 &62.39 \\
&300 &20.00 &68.00 &72.00 &100.00 &52.00 &96.00 &84.00 &71.25 &36.50 &66.14 &30.11 &69.88 &86.16 &59.02 \\\cmidrule{1-16}
\multirow{3}{*}{MAE-B} &100 &24.00 &88.00 &88.00 &100.00 &84.00 &96.00 &96.00 &88.28 &42.55 &95.18 &44.08 &69.26 &85.63 &55.68 \\
&200 &28.00 &76.00 &84.00 &100.00 &84.00 &88.00 &88.00 &77.13 &38.49 &88.22 &32.75 &69.02 &85.14 &56.81 \\
&300 &28.00 &72.00 &76.00 &100.00 &80.00 &88.00 &80.00 &75.60 &36.93 &78.35 &31.03 &69.01 &84.11 &57.30 \\\midrule
\multirow{3}{*}{DINOV2-B} &100 &8.00 &60.00 &40.00 &100.00 &44.00 &96.00 &20.00 &45.95 &16.61 &63.57 &13.38 &60.11 &74.07 &36.29 \\
&200 &8.00 &68.00 &40.00 &100.00 &64.00 &88.00 &16.00 &37.96 &15.81 &51.44 &12.59 &59.56 &74.18 &32.14 \\
&300 &12.00 &64.00 &48.00 &100.00 &64.00 &88.00 &4.00 &32.43 &14.31 &36.01 &11.67 &56.54 &73.77 &36.53 \\\cmidrule{1-16}
\multirow{3}{*}{VC-1-B} &100 &20.00 &76.00 &76.00 &100.00 &76.00 &100.00 &76.00 &72.35 &43.14 &92.77 &27.31 &68.67 &84.19 &62.00 \\
&200 &32.00 &80.00 &68.00 &100.00 &76.00 &100.00 &92.00 &81.83 &44.05 &83.62 &27.80 &70.98 &83.88 &59.63 \\
&300 &24.00 &68.00 &80.00 &100.00 &80.00 &88.00 &88.00 &83.01 &41.25 &77.60 &28.53 &70.89 &84.76 &59.51 \\\cmidrule{1-16}
\multirow{3}{*}{RADIO} &100 &28.00 &76.00 &48.00 &100.00 &72.00 &100.00 &84.00 &87.84 &62.72 &96.53 &15.71 &67.88 &85.70 &57.52 \\
&200 &36.00 &76.00 &44.00 &100.00 &72.00 &96.00 &52.00 &80.26 &57.39 &95.93 &15.26 &67.51 &85.67 &57.81 \\
&300 &44.00 &72.00 &32.00 &100.00 &40.00 &92.00 &48.00 &79.62 &53.51 &89.16 &14.80 &66.57 &85.64 &58.14 \\\cmidrule{1-16}
\multirow{3}{*}{E-RADIO} &100 &32.00 &84.00 &64.00 &100.00 &84.00 &96.00 &96.00 &71.47 &53.41 &93.01 &50.19 &70.75 &87.17 &46.97 \\
&200 &32.00 &84.00 &60.00 &100.00 &68.00 &88.00 &96.00 &68.80 &44.56 &88.54 &33.19 &70.46 &87.39 &50.72 \\
&300 &28.00 &80.00 &60.00 &100.00 &72.00 &88.00 &80.00 &65.96 &33.56 &98.14 &32.64 &69.18 &87.09 &51.31 \\\cmidrule{1-16}
\multirow{3}{*}{SPA-B} &100 &20.00 &84.00 &84.00 &100.00 &88.00 &100.00 &100.00 &80.50 &45.08 &91.38 &48.90 &71.16 &86.04 &59.03 \\
&200 &28.00 &80.00 &68.00 &100.00 &84.00 &100.00 &84.00 &79.71 &46.65 &85.75 &40.84 &71.01 &86.16 &60.05 \\
&300 &24.00 &80.00 &88.00 &100.00 &92.00 &100.00 &92.00 &74.70 &48.97 &81.60 &34.92 &71.16 &85.16 &61.94 \\
\bottomrule
\end{tabular}
}
\end{table}

\begin{table}[!htp]\centering
\caption{\textbf{All results on Franka Kitchen.}}\label{tab:appx-fk}
\vspace{-1.2em}
\tablestyle{4.5pt}{1.05}
\begin{tabular}{l|c|c|cccccccccc}\toprule
Task &View &Seed &MoCoV3 &MAE &DINOV2 &CLIP &EVA &\makecell{InternViT-\\300M} &MVP &VC-1 &SPA \\\cmidrule{1-12}
\multirow{6}{*}{Task 1} &\multirow{3}{*}{Left} &100 &86.00 &76.00 &84.00 &72.00 &78.00 &74.00 &66.00 &74.00 &84.00 \\
& &200 &78.00 &78.00 &74.00 &72.00 &76.00 &72.00 &58.00 &74.00 &92.00 \\
& &300 &80.00 &80.00 &78.00 &62.00 &82.00 &70.00 &64.00 &74.00 &80.00 \\\cmidrule{2-12}
&\multirow{3}{*}{Right} &100 &82.00 &80.00 &86.00 &78.00 &78.00 &72.00 &82.00 &78.00 &86.00 \\
& &200 &88.00 &62.00 &90.00 &70.00 &86.00 &86.00 &86.00 &84.00 &72.00 \\
& &300 &86.00 &82.00 &92.00 &82.00 &86.00 &76.00 &92.00 &78.00 &86.00 \\\cmidrule{1-12}
\multirow{6}{*}{Task 2} &\multirow{3}{*}{Left} &100 &60.00 &56.00 &48.00 &26.00 &40.00 &22.00 &40.00 &32.00 &48.00 \\
& &200 &64.00 &60.00 &46.00 &44.00 &40.00 &32.00 &32.00 &42.00 &60.00 \\
& &300 &58.00 &54.00 &40.00 &26.00 &32.00 &34.00 &30.00 &50.00 &66.00 \\\cmidrule{2-12}
&\multirow{3}{*}{Right} &100 &62.00 &54.00 &56.00 &26.00 &44.00 &26.00 &32.00 &54.00 &48.00 \\
& &200 &64.00 &54.00 &60.00 &36.00 &40.00 &24.00 &28.00 &56.00 &42.00 \\
& &300 &64.00 &52.00 &50.00 &38.00 &40.00 &30.00 &34.00 &44.00 &42.00 \\\cmidrule{1-12}
\multirow{6}{*}{Task 3} &\multirow{3}{*}{Left} &100 &16.00 &24.00 &18.00 &18.00 &22.00 &24.00 &6.00 &24.00 &28.00 \\
& &200 &28.00 &20.00 &14.00 &18.00 &20.00 &16.00 &6.00 &30.00 &38.00 \\
& &300 &22.00 &16.00 &14.00 &10.00 &26.00 &22.00 &8.00 &26.00 &30.00 \\\cmidrule{2-12}
&\multirow{3}{*}{Right} &100 &46.00 &26.00 &38.00 &22.00 &14.00 &8.00 &32.00 &12.00 &10.00 \\
& &200 &48.00 &22.00 &38.00 &24.00 &18.00 &4.00 &32.00 &12.00 &12.00 \\
& &300 &54.00 &34.00 &52.00 &14.00 &12.00 &6.00 &26.00 &14.00 &16.00 \\\cmidrule{1-12}
\multirow{6}{*}{Task 4} &\multirow{3}{*}{Left} &100 &32.00 &36.00 &26.00 &22.00 &34.00 &12.00 &16.00 &36.00 &22.00 \\
& &200 &30.00 &30.00 &32.00 &14.00 &20.00 &8.00 &14.00 &24.00 &10.00 \\
& &300 &24.00 &46.00 &28.00 &14.00 &32.00 &4.00 &20.00 &36.00 &16.00 \\\cmidrule{2-12}
&\multirow{3}{*}{Right} &100 &38.00 &24.00 &28.00 &22.00 &32.00 &12.00 &26.00 &12.00 &30.00 \\
& &200 &42.00 &24.00 &24.00 &24.00 &24.00 &12.00 &30.00 &8.00 &38.00 \\
& &300 &46.00 &16.00 &32.00 &28.00 &36.00 &16.00 &26.00 &12.00 &30.00 \\\cmidrule{1-12}
\multirow{6}{*}{Task 5} &\multirow{3}{*}{Left} &100 &36.00 &18.00 &8.00 &16.00 &24.00 &22.00 &26.00 &28.00 &20.00 \\
& &200 &30.00 &24.00 &8.00 &10.00 &24.00 &16.00 &20.00 &22.00 &16.00 \\
& &300 &22.00 &22.00 &10.00 &12.00 &16.00 &14.00 &28.00 &30.00 &18.00 \\\cmidrule{2-12}
&\multirow{3}{*}{Right} &100 &24.00 &46.00 &20.00 &4.00 &14.00 &10.00 &26.00 &22.00 &30.00 \\
& &200 &24.00 &30.00 &16.00 &8.00 &18.00 &18.00 &22.00 &22.00 &26.00 \\
& &300 &14.00 &36.00 &16.00 &12.00 &10.00 &12.00 &20.00 &16.00 &22.00 \\
\bottomrule
\end{tabular}
\end{table}

\begin{table}[!htp]\centering
\caption{\textbf{All results on Meta-World.}}\label{tab:appx-mw}
\vspace{-1.2em}
\resizebox{\linewidth}{!}{
\tablestyle{0.75pt}{1.05}
\begin{tabular}{l|ccc|ccc|ccc|ccc|ccc|ccc|ccc|ccc|cccc}\toprule
Method &\multicolumn{3}{c|}{MoCoV3} &\multicolumn{3}{c|}{MAE} &\multicolumn{3}{c|}{DINOV2} &\multicolumn{3}{c|}{CLIP} &\multicolumn{3}{c|}{EVA} &\multicolumn{3}{c|}{\makecell{InternViT-\\300M}} &\multicolumn{3}{c|}{MVP} &\multicolumn{3}{c|}{VC-1} &\multicolumn{3}{c}{SPA} \\\cmidrule{1-28}
Seed &100 &200 &300 &100 &200 &300 &100 &200 &300 &100 &200 &300 &100 &200 &300 &100 &200 &300 &100 &200 &300 &100 &200 &300 &100 &200 &300 \\\cmidrule{1-28}
ButtonPressWall &100 &100 &100 &100 &100 &100 &95 &90 &95 &100 &100 &100 &95 &95 &100 &95 &100 &95 &100 &100 &100 &100 &100 &100 &100 &100 &100 \\
DoorClose &100 &100 &100 &100 &100 &100 &100 &100 &100 &100 &100 &100 &100 &100 &100 &100 &100 &100 &100 &100 &100 &100 &100 &100 &100 &100 &100 \\
DoorUnlock &70 &65 &80 &70 &65 &85 &35 &35 &30 &60 &50 &60 &85 &85 &90 &80 &65 &75 &75 &70 &85 &80 &75 &90 &80 &75 &80 \\
DrawerClose &100 &100 &100 &100 &100 &100 &100 &100 &100 &100 &100 &100 &100 &100 &100 &100 &100 &100 &100 &100 &100 &100 &100 &100 &100 &100 &100 \\
DrawerOpen &80 &70 &85 &85 &65 &75 &60 &40 &60 &90 &80 &80 &60 &55 &75 &70 &75 &75 &95 &75 &85 &85 &85 &80 &90 &60 &75 \\
FaucetClose &70 &80 &55 &60 &80 &60 &55 &65 &35 &70 &80 &50 &65 &75 &60 &50 &65 &50 &70 &75 &60 &65 &75 &100 &70 &80 &65 \\
PlateSlide &90 &95 &100 &95 &100 &100 &65 &70 &80 &85 &95 &80 &100 &100 &100 &85 &95 &90 &100 &100 &100 &100 &95 &100 &95 &95 &95 \\
PlateSlideBack &80 &65 &85 &85 &65 &85 &90 &75 &90 &85 &75 &90 &80 &65 &90 &85 &80 &90 &85 &70 &90 &80 &70 &90 &80 &70 &85 \\
PlateSlideSide &85 &90 &95 &95 &90 &95 &90 &85 &85 &80 &85 &95 &100 &95 &100 &90 &95 &90 &80 &90 &95 &100 &95 &90 &90 &90 &95 \\
WindowClose &100 &100 &100 &100 &100 &100 &70 &90 &100 &100 &100 &95 &95 &100 &100 &100 &100 &100 &100 &100 &100 &100 &100 &100 &100 &100 &100 \\
Basketball &85 &95 &95 &95 &100 &100 &70 &55 &65 &85 &85 &70 &95 &85 &80 &90 &80 &95 &95 &100 &95 &90 &100 &95 &95 &100 &100 \\
BinPicking & 30 & 45 & 40 &20 &10 &35 &10 &15 &10 &30 &30 &25 &20 &5 &45 &10 &10 &10 &25 &20 &15 &30 &30 &30 &40 &25 &30 \\
BoxClose &80 &80 &80 &75 &80 &70 &35 &45 &30 &80 &70 &60 &55 &70 &55 &60 &65 &40 &80 &80 &65 &80 &95 &60 &80 &80 &65 \\
CoffeePush &45 &50 &40 &40 &55 &30 &30 &25 &15 &45 &40 &55 &45 &35 &40 &45 &30 &25 &25 &45 &25 &30 &45 &55 &35 &40 &30 \\
Assembly &70 &60 &55 &55 &65 &45 &30 &35 &25 &45 &55 &50 &45 &50 &40 &30 &25 &30 &60 &60 &45 &60 &60 &50 &50 &55 &50 \\
Disassemble &40 &55 &50 &30 &45 &45 &30 &20 &45 &55 &50 &60 &30 &50 &50 &40 &45 &50 &40 &45 &45 &40 &30 &35 &40 &45 &55 \\
PushWall &25 &35 &30 &20 &30 &40 &40 &30 &45 &35 &30 &35 &25 &35 &30 &15 &15 &25 &30 &35 &35 &55 &55 &60 &30 &40 &45 \\
ShelfPlace &35 &35 &20 &25 &45 &20 &30 &30 &35 &30 &35 &30 &25 &20 &15 &15 &15 &15 &25 &25 &15 &25 &15 &15 &15 &35 &15 \\
DoorOpen &95 &90 &90 &100 &95 &95 &95 &80 &95 &85 &75 &90 &80 &90 &100 &50 &55 &60 &80 &75 &95 &95 &95 &95 &85 &100 &95 \\
ButtonPress &75 &85 &85 &85 &100 &100 &80 &95 &85 &55 &70 &75 &80 &90 &90 &85 &90 &80 &85 &90 &95 &80 &90 &85 &100 &95 &100 \\
SweepInto &45 &45 &40 &55 &55 &45 &50 &50 &40 &45 &45 &40 &45 &25 &30 &35 &25 &30 &45 &50 &40 &55 &50 &50 &50 &50 &45 \\
DoorLock &100 &85 &85 &90 &100 &100 &95 &90 &85 &85 &85 &75 &95 &100 &95 &80 &90 &95 &85 &100 &95 &100 &100 &90 &80 &95 &85 \\
ReachWall &70 &70 &80 &75 &85 &70 &85 &80 &85 &90 &90 &80 &65 &75 &85 &60 &55 &55 &75 &65 &75 &75 &80 &80 &75 &80 &70 \\
Hammer &25 &45 &30 &30 &30 &30 &40 &45 &35 &25 &35 &25 &30 &35 &20 &20 &20 &20 &30 &35 &30 &45 &40 &55 &30 &40 &30 \\
StickPush &95 &95 &100 &80 &90 &95 &90 &90 &90 &75 &85 &85 &90 &85 &100 &60 &80 &85 &90 &95 &90 &85 &85 &95 &90 &90 &85 \\
ButtonPressTopdown &80 &80 &75 &80 &85 &80 &80 &90 &90 &80 &90 &70 &55 &65 &55 &45 &55 &55 &80 &85 &80 &75 &80 &70 &80 &85 &80 \\
HandlePressSide &100 &100 &100 &100 &100 &100 &95 &100 &90 &80 &100 &90 &100 &100 &95 &85 &100 &90 &95 &100 &100 &75 &100 &80 &90 &100 &100 \\
PlateSlideBackSide &100 &100 &100 &100 &95 &95 &100 &100 &100 &100 &100 &100 &100 &100 &100 &100 &100 &100 &100 &100 &100 &100 &100 &100 &100 &100 &100 \\
Sweep &50 &80 &70 &35 &60 &60 &65 &85 &95 &60 &85 &75 &35 &50 &55 &15 &60 &35 &35 &70 &65 &35 &65 &65 &30 &65 &55 \\
ButtonPressTopdownWall &45 &70 &80 &45 &75 &75 &70 &75 &75 &45 &60 &70 &30 &45 &65 &20 &55 &45 &30 &60 &70 &50 &85 &80 &45 &65 &75 \\
HandlePress &85 &95 &95 &80 &100 &75 &75 &100 &80 &90 &100 &90 &85 &100 &85 &65 &90 &75 &80 &95 &75 &85 &100 &90 &85 &100 &80 \\
Push &25 &30 &30 &25 &30 &30 &30 &25 &40 &25 &15 &30 &30 &25 &20 &25 &20 &20 &25 &20 &25 &40 &30 &25 &30 &15 &35 \\
CoffeePull &55 &55 &55 &40 &45 &40 &20 &30 &20 &50 &70 &40 &40 &45 &45 &40 &40 &25 &55 &55 &40 &55 &55 &60 &55 &55 &55 \\
DialTurn &80 &65 &80 &85 &75 &75 &40 &30 &35 &80 &95 &90 &65 &65 &80 &70 &70 &55 &70 &65 &75 &80 &95 &75 &85 &85 &75 \\
Reach &90 &75 &80 &90 &75 &80 &70 &75 &85 &95 &95 &100 &85 &80 &75 &95 &80 &90 &80 &70 &75 &70 &80 &80 &85 &70 &85 \\
CoffeeButton &85 &95 &75 &100 &100 &95 &85 &80 &60 &95 &100 &85 &90 &100 &95 &90 &85 &85 &100 &100 &90 &90 &100 &80 &100 &100 &100 \\
PickPlaceWall &45 &35 &65 &40 &25 &45 &15 &10 &20 &35 &40 &50 &20 &25 &35 &30 &25 &25 &25 &35 &40 &35 &20 &45 &40 &35 &55 \\
StickPull &35 &35 &25 &15 &40 &20 &25 &10 &5 &45 &40 &45 &25 &35 &15 &25 &25 &15 &15 &30 &25 &30 &30 &30 &25 &35 &30 \\
HandInsert &35 &30 &30 &30 &25 &25 &20 &20 &20 &45 &45 &40 &20 &25 &20 &25 &30 &25 &40 &40 &35 &40 &30 &40 &40 &50 &40 \\
PegInsertSide &40 &35 &40 &50 &35 &45 &25 &15 &10 &45 &30 &20 &45 &25 &30 &30 &20 &25 &45 &45 &30 &50 &25 &35 &55 &45 &60 \\
PickPlace &35 &30 &30 &25 &45 &15 &15 &10 &10 &25 &15 &30 &25 &30 &25 &30 &10 &30 &20 &35 &25 &25 &20 &20 &25 &30 &40 \\
FaucetOpen &95 &95 &100 &100 &100 &100 &80 &80 &100 &100 &95 &100 &95 &95 &100 &80 &85 &95 &100 &100 &100 &100 &100 &100 &95 &100 &100 \\
PushBack &65 &70 &60 &40 &55 &40 &15 &15 &25 &30 &45 &25 &35 &35 &45 &15 &35 &25 &40 &45 &45 &45 &55 &25 &35 &55 &50 \\
LeverPull &70 &80 &80 &80 &70 &75 &15 &30 &35 &55 &85 &80 &70 &65 &70 &60 &55 &55 &65 &80 &65 &65 &80 &70 &85 &70 &80 \\
HandlePull &85 &85 &80 &85 &80 &85 &45 &55 &40 &75 &75 &80 &80 &60 &65 &45 &70 &65 &80 &90 &80 &70 &85 &75 &100 &90 &90 \\
Soccer &25 &40 &35 &50 &30 &25 &15 &10 &15 &45 &40 &30 &35 &35 &25 &20 &20 &30 &30 &30 &35 &20 &20 &20 &25 &50 &25 \\
WindowOpen &65 &80 &80 &55 &80 &85 &60 &50 &65 &50 &65 &60 &65 &80 &75 &55 &75 &75 &60 &70 &70 &60 &70 &75 &55 &85 &65 \\
PickOutOfHole &65 &75 &80 &65 &65 &60 &60 &55 &60 &70 &70 &50 &60 &55 &50 &60 &55 &55 &65 &55 &60 &70 &75 &60 &65 &60 &75 \\
\bottomrule
\end{tabular}
}
\end{table}

\begin{table}[!htp]\centering
\caption{\textbf{All results on RLBench.}}\label{tab:appx-rlbench}
\vspace{-1.2em}
\resizebox{0.95\linewidth}{!}{
\tablestyle{4pt}{0.95}
\begin{tabular}{l|cccccccccc}\toprule
Method &MoCoV3 &MAE &DINOV2 &CLIP &EVA &\makecell{InternViT\\300M} &MVP &VC-1 &SPA \\\midrule
\textit{Group 1} & & & & & & & & & \\
basketball in hoop &100 &100 &100 &100 &100 &100 &100 &100 &100 \\
put rubbish in bin &100 &100 &96 &96 &96 &100 &96 &100 &100 \\
meat off grill &100 &100 &100 &100 &100 &100 &100 &100 &100 \\
meat on grill &80 &76 &76 &68 &80 &72 &68 &76 &80 \\
slide block to target &0 &84 &96 &24 &4 &0 &100 &100 &4 \\
reach and drag &100 &96 &88 &100 &96 &100 &96 &100 &100 \\
take frame off hanger &88 &88 &92 &88 &84 &84 &88 &88 &96 \\
water plants &64 &60 &28 &64 &60 &44 &52 &60 &68 \\
hang frame on hanger &8 &4 &0 &4 &8 &8 &12 &4 &4 \\
wipe desk &0 &0 &0 &0 &0 &0 &0 &0 &0 \\
stack blocks &60 &72 &72 &68 &56 &60 &84 &68 &68 \\
reach target &60 &96 &88 &100 &96 &80 &92 &96 &92 \\
push button &100 &100 &100 &100 &100 &100 &100 &100 &100 \\
lamp on &88 &68 &84 &88 &52 &80 &28 &88 &64 \\
toilet seat down &100 &100 &100 &100 &100 &100 &96 &96 &100 \\
close laptop lid &96 &96 &96 &96 &84 &80 &80 &96 &100 \\
open box &12 &12 &20 &4 &16 &4 &0 &12 &16 \\
open drawer &88 &96 &92 &100 &88 &88 &92 &96 &96 \\
pick up cup &92 &92 &88 &96 &96 &88 &96 &96 &96 \\
turn tap &88 &84 &84 &96 &88 &92 &96 &100 &100 \\
take usb out of computer &100 &100 &100 &100 &100 &100 &100 &88 &100 \\
play jenga &96 &96 &96 &100 &96 &100 &96 &96 &96 \\
insert onto square peg &28 &84 &80 &44 &88 &40 &64 &92 &84 \\
take umbrella out of umbrella stand &92 &100 &100 &92 &100 &96 &100 &100 &100 \\
insert usb in computer &12 &20 &20 &24 &24 &20 &16 &8 &68 \\
straighten rope &56 &44 &72 &80 &48 &72 &52 &60 &84 \\
turn oven on &96 &96 &96 &96 &96 &96 &100 &100 &100 \\
change clock &64 &68 &48 &68 &64 &72 &64 &60 &68 \\
close microwave &100 &100 &100 &100 &100 &100 &100 &100 &100 \\
close fridge &80 &92 &92 &88 &92 &96 &88 &92 &100 \\
close grill &96 &96 &96 &96 &96 &96 &100 &100 &96 \\
open grill &100 &100 &100 &100 &100 &100 &96 &100 &100 \\
unplug charger &44 &32 &48 &36 &48 &40 &40 &44 &44 \\
press switch &92 &92 &88 &72 &76 &84 &76 &88 &92 \\
take money out safe &100 &96 &100 &100 &100 &100 &100 &100 &100 \\\midrule
\textit{Group 2} & & & & & & & & & \\
change channel &0 &8 &4 &0 &0 &4 &0 &0 &4 \\
tv on &4 &8 &0 &4 &4 &8 &4 &4 &8 \\
push buttons &12 &4 &4 &0 &0 &0 &0 &12 &4 \\
stack wine &12 &16 &40 &4 &12 &0 &28 &8 &28 \\
scoop with spatula &0 &0 &0 &0 &0 &0 &0 &0 &0 \\
place hanger on rack &0 &0 &0 &0 &0 &0 &0 &0 &0 \\
move hanger &0 &0 &0 &0 &0 &0 &0 &0 &0 \\
sweep to dustpan &92 &96 &96 &96 &92 &100 &100 &88 &96 \\
take plate off colored dish rack &96 &100 &96 &92 &84 &96 &88 &92 &96 \\
screw nail &52 &36 &36 &36 &36 &52 &32 &32 &48 \\
take shoes out of box &20 &28 &24 &36 &40 &12 &32 &36 &36 \\
slide cabinet open and place cups &0 &0 &0 &0 &0 &4 &0 &0 &4 \\
lamp off &100 &96 &96 &100 &96 &96 &100 &100 &100 \\
pick and lift &88 &96 &92 &96 &92 &80 &96 &96 &96 \\
take lid off saucepan &100 &100 &100 &100 &100 &100 &100 &100 &100 \\
close drawer &100 &100 &100 &100 &96 &100 &100 &100 &100 \\
close box &92 &92 &96 &96 &100 &96 &100 &96 &100 \\
phone on base &100 &100 &100 &100 &100 &96 &100 &100 &100 \\
toilet seat up &80 &88 &100 &88 &88 &80 &88 &92 &96 \\
put books on bookshelf &12 &24 &24 &28 &28 &20 &20 &28 &16 \\
beat the buzz &88 &92 &96 &88 &92 &84 &88 &88 &100 \\
stack cups &40 &56 &52 &52 &48 &56 &64 &68 &64 \\
put knife on chopping board &72 &76 &68 &72 &80 &88 &80 &76 &80 \\
place shape in shape sorter &20 &36 &32 &28 &36 &20 &44 &36 &56 \\
take toilet roll off stand &100 &92 &76 &96 &92 &88 &84 &92 &96 \\
put umbrella in umbrella stand &8 &0 &12 &12 &0 &4 &12 &8 &12 \\
setup checkers &76 &80 &68 &68 &88 &92 &92 &80 &80 \\
open window &96 &96 &100 &100 &96 &100 &96 &100 &100 \\
open wine bottle &80 &100 &88 &92 &92 &88 &96 &88 &88 \\
open microwave &100 &100 &88 &96 &100 &80 &96 &100 &100 \\
put money in safe &96 &100 &88 &92 &100 &96 &100 &100 &100 \\
open door &100 &96 &96 &96 &96 &96 &84 &96 &96 \\
close door &32 &68 &56 &60 &80 &20 &24 &20 &60 \\
open fridge &44 &52 &48 &44 &36 &64 &52 &32 &64 \\
open oven &8 &4 &12 &8 &4 &20 &4 &4 &16 \\
plug charger in power supply &32 &36 &32 &24 &44 &36 &24 &32 &60 \\
\bottomrule
\end{tabular}
}
\end{table}

\begin{table}[!htp]\centering
\caption{\textbf{All results on LIBERO-OBJECT, LIBERO-SPATIAL, LIBERO-GOAL, LIBERO-10.}}\label{tab:appx-libero}
\vspace{-1.2em}
\resizebox{\linewidth}{!}{
\tablestyle{1pt}{1.05}
\begin{tabular}{l|ccc|ccc|ccc|ccc|ccc|ccc|ccc|ccc|ccc|ccc}\toprule
 &\multicolumn{3}{c|}{MoCoV3} &\multicolumn{3}{c|}{MAE} &\multicolumn{3}{c|}{DINOV2} &\multicolumn{3}{c|}{CLIP} &\multicolumn{3}{c|}{EVA} &\multicolumn{3}{c|}{\makecell{InternViT-\\300M}} &\multicolumn{3}{c|}{\makecell{InternViT-\\6B}} &\multicolumn{3}{c|}{MVP} &\multicolumn{3}{c|}{VC-1} &\multicolumn{3}{c}{SPA} \\\midrule
Seed &100 &200 &300 &100 &200 &300 &100 &200 &300 &100 &200 &300 &100 &200 &300 &100 &200 &300 &100 &200 &300 &100 &200 &300 &100 &200 &300 &100 &200 &300 \\\midrule
\multicolumn{31}{l}{\textit{LIBERO-OBJECT}}  \\\midrule
0 &0.65 &0.60 &0.65 &0.65 &0.45 &0.55 &0.65 &0.80 &0.85 &0.80 &0.75 &0.65 &1.00 &0.70 &0.95 &0.80 &0.65 &0.60 &0.70 &0.85 &0.50 &0.80 &0.90 &0.65 &0.80 &0.50 &0.60 &0.90 &0.95 &0.95 \\
1 &0.35 &0.35 &0.55 &0.90 &0.75 &0.80 &0.30 &0.50 &0.75 &0.40 &0.30 &0.05 &0.65 &0.30 &0.70 &0.15 &0.40 &0.20 &0.60 &0.25 &0.45 &0.05 &0.80 &0.60 &0.40 &0.65 &0.45 &0.65 &0.70 &0.45 \\
2 &0.90 &0.85 &0.95 &0.90 &0.40 &0.95 &0.85 &0.50 &0.90 &0.70 &0.80 &0.75 &0.85 &0.75 &0.75 &0.90 &0.85 &0.80 &0.85 &0.45 &0.85 &0.80 &0.85 &0.90 &1.00 &0.95 &0.95 &0.90 &0.95 &0.80 \\
3 &0.55 &0.70 &0.65 &0.90 &0.15 &0.90 &0.30 &0.65 &0.90 &0.25 &0.45 &0.60 &0.80 &0.80 &0.90 &0.75 &0.70 &0.40 &1.00 &0.50 &0.55 &0.70 &0.65 &0.85 &0.95 &0.75 &0.60 &0.70 &0.90 &0.90 \\
4 &0.65 &0.85 &0.85 &0.80 &0.90 &0.75 &0.75 &0.55 &0.75 &0.35 &0.75 &0.65 &0.95 &0.75 &1.00 &0.90 &1.00 &0.85 &0.90 &0.70 &0.80 &0.80 &0.75 &0.70 &0.90 &0.85 &0.90 &0.90 &1.00 &0.95 \\
5 &0.50 &0.70 &0.80 &0.70 &0.35 &0.60 &0.55 &0.75 &0.60 &0.25 &0.70 &0.45 &0.75 &0.75 &0.65 &0.85 &0.60 &0.75 &0.60 &0.35 &0.50 &0.55 &0.40 &0.80 &0.65 &0.70 &0.70 &0.25 &0.15 &0.65 \\
6 &0.35 &0.50 &0.65 &0.60 &0.65 &0.65 &0.55 &0.70 &0.70 &0.35 &0.55 &0.60 &0.40 &0.35 &0.25 &0.65 &0.60 &0.55 &0.30 &0.10 &0.35 &0.25 &0.50 &0.65 &0.50 &0.50 &0.30 &0.50 &0.70 &0.80 \\
7 &0.75 &0.75 &0.80 &0.90 &0.40 &0.75 &0.55 &0.30 &0.70 &0.40 &0.35 &0.40 &0.55 &0.75 &0.70 &0.80 &0.40 &0.60 &0.60 &0.65 &0.70 &0.60 &0.45 &0.65 &0.80 &0.75 &0.50 &0.80 &0.75 &0.65 \\
8 &0.50 &0.95 &0.90 &1.00 &0.95 &1.00 &0.50 &0.35 &0.50 &0.45 &0.45 &0.35 &1.00 &0.75 &0.85 &0.70 &0.65 &0.75 &0.50 &0.40 &0.75 &0.65 &0.55 &0.70 &0.80 &0.90 &0.50 &0.85 &0.95 &0.90 \\
9 &0.45 &0.50 &0.40 &0.60 &0.65 &0.95 &0.80 &0.90 &0.95 &0.50 &0.70 &0.30 &0.85 &0.75 &0.75 &0.85 &0.95 &0.65 &0.90 &0.60 &0.15 &0.65 &0.60 &0.30 &0.70 &0.70 &0.65 &0.60 &0.95 &0.90 \\\midrule
\multicolumn{31}{l}{\textit{LIBERO-SPATIAL}}  \\\midrule
0 &0.35 &0.55 &0.45 &0.45 &0.40 &0.70 &0.65 &0.50 &0.60 &0.25 &0.20 &0.35 &0.70 &0.75 &0.65 &0.55 &0.65 &0.55 &0.55 &0.50 &0.30 &0.75 &0.75 &0.60 &0.35 &0.55 &0.60 &0.45 &0.50 &0.35 \\
1 &0.65 &0.70 &0.70 &0.80 &0.80 &0.50 &0.55 &0.30 &0.35 &0.75 &0.75 &0.70 &0.55 &0.70 &0.25 &0.35 &0.50 &0.50 &1.00 &1.00 &0.90 &0.60 &0.40 &0.60 &0.45 &0.65 &0.80 &0.65 &0.65 &0.85 \\
2 &0.55 &0.50 &0.50 &0.35 &0.60 &0.40 &0.20 &0.05 &0.55 &0.10 &0.00 &0.40 &0.70 &0.80 &0.50 &0.70 &0.75 &0.60 &0.75 &0.60 &0.20 &0.85 &0.55 &0.75 &0.45 &0.45 &0.70 &0.50 &0.50 &0.40 \\
3 &0.50 &0.70 &0.75 &0.55 &0.60 &0.75 &0.80 &0.70 &0.95 &0.15 &0.40 &0.30 &0.85 &0.90 &0.85 &0.35 &0.50 &0.40 &0.40 &0.30 &0.15 &0.95 &0.55 &0.60 &0.50 &0.70 &0.65 &0.55 &0.85 &0.60 \\
4 &0.15 &0.15 &0.20 &0.55 &0.70 &0.80 &0.50 &0.05 &0.45 &0.35 &0.30 &0.20 &0.45 &0.55 &0.40 &0.35 &0.25 &0.40 &0.25 &0.15 &0.15 &0.60 &0.50 &0.70 &0.60 &0.60 &0.80 &0.70 &0.70 &0.50 \\
5 &0.45 &0.10 &0.10 &0.65 &0.40 &0.30 &0.30 &0.20 &0.35 &0.55 &0.45 &0.45 &0.65 &0.50 &0.45 &0.40 &0.30 &0.70 &0.55 &0.60 &0.60 &0.55 &0.30 &0.25 &0.05 &0.05 &0.15 &0.35 &0.35 &0.30 \\
6 &0.30 &0.35 &0.45 &0.55 &0.25 &0.95 &0.40 &0.30 &0.40 &0.20 &0.25 &0.10 &0.75 &0.70 &0.85 &0.45 &0.55 &0.55 &0.40 &0.35 &0.05 &0.45 &0.75 &0.65 &0.60 &0.70 &0.35 &0.35 &0.45 &0.30 \\
7 &0.10 &0.20 &0.25 &0.50 &0.35 &0.45 &0.05 &0.00 &0.10 &0.30 &0.15 &0.25 &0.30 &0.30 &0.20 &0.30 &0.60 &0.65 &0.15 &0.05 &0.05 &0.60 &0.65 &0.60 &0.10 &0.20 &0.40 &0.40 &0.15 &0.45 \\
8 &0.55 &0.70 &0.50 &0.35 &0.65 &0.70 &0.40 &0.15 &0.55 &0.30 &0.55 &0.30 &0.85 &0.70 &0.60 &0.30 &0.55 &0.60 &0.65 &0.40 &0.35 &0.70 &0.40 &0.55 &0.70 &0.60 &0.70 &0.75 &0.70 &0.40 \\
9 &0.55 &0.05 &0.10 &0.85 &0.75 &0.50 &0.20 &0.05 &0.25 &0.20 &0.20 &0.20 &0.55 &0.50 &0.30 &0.35 &0.50 &0.30 &0.45 &0.40 &0.35 &0.45 &0.45 &0.30 &0.50 &0.55 &0.65 &0.35 &0.50 &0.45 \\\midrule
\multicolumn{31}{l}{\textit{LIBERO-GOAL}}  \\\midrule
0 &0.45 &0.70 &0.75 &0.70 &0.85 &0.80 &0.15 &0.10 &0.30 &0.25 &0.40 &0.35 &0.70 &0.60 &0.60 &0.75 &0.65 &0.75 &0.25 &0.35 &0.35 &0.75 &0.60 &0.95 &0.45 &0.85 &1.00 &0.85 &1.00 &0.85 \\
1 &0.70 &0.60 &0.80 &0.65 &0.50 &0.90 &0.25 &0.55 &0.25 &0.20 &0.15 &0.25 &0.70 &0.80 &0.80 &0.90 &0.90 &1.00 &0.40 &0.15 &0.15 &0.90 &0.80 &0.95 &0.65 &0.65 &0.65 &1.00 &0.85 &0.90 \\
2 &0.50 &0.20 &0.15 &0.10 &0.40 &0.35 &0.10 &0.05 &0.15 &0.30 &0.25 &0.30 &0.65 &0.75 &0.75 &0.40 &0.75 &0.45 &0.50 &0.35 &0.35 &0.45 &0.25 &0.65 &0.40 &0.60 &0.35 &0.50 &0.55 &0.35 \\
3 &0.75 &0.45 &0.60 &0.40 &0.75 &0.55 &0.20 &0.10 &0.10 &0.05 &0.20 &0.55 &0.30 &0.15 &0.15 &0.30 &0.50 &0.65 &0.20 &0.25 &0.25 &0.70 &0.70 &0.15 &0.75 &0.55 &0.50 &0.65 &0.35 &0.80 \\
4 &0.20 &0.25 &0.05 &0.35 &0.40 &0.25 &0.10 &0.00 &0.05 &0.40 &0.30 &0.15 &0.15 &0.10 &0.10 &0.20 &0.15 &0.10 &0.15 &0.20 &0.20 &0.55 &0.60 &0.25 &0.15 &0.30 &0.30 &0.30 &0.35 &0.35 \\
5 &0.10 &0.75 &0.80 &0.60 &0.85 &0.80 &0.65 &0.50 &0.50 &0.35 &0.45 &0.50 &0.80 &0.75 &0.75 &0.70 &0.55 &0.45 &0.55 &0.45 &0.45 &0.80 &0.75 &0.85 &0.65 &0.75 &0.80 &0.80 &0.65 &0.65 \\
6 &0.45 &0.05 &0.15 &0.00 &0.10 &0.05 &0.00 &0.05 &0.00 &0.10 &0.10 &0.00 &0.00 &0.65 &0.65 &0.50 &0.40 &0.30 &0.00 &0.00 &0.00 &0.15 &0.35 &0.70 &0.25 &0.50 &0.45 &0.40 &0.30 &0.35 \\
7 &0.25 &0.75 &0.90 &0.80 &0.65 &1.00 &0.45 &0.45 &0.35 &0.50 &0.65 &0.80 &1.00 &1.00 &1.00 &1.00 &1.00 &0.95 &0.70 &0.85 &0.85 &0.95 &1.00 &0.95 &1.00 &0.95 &0.70 &0.95 &1.00 &1.00 \\
8 &0.50 &0.80 &0.75 &0.85 &0.55 &0.90 &0.45 &0.40 &0.20 &0.60 &0.25 &0.50 &0.90 &0.35 &0.35 &0.95 &0.65 &0.55 &0.65 &0.25 &0.25 &0.70 &0.65 &0.70 &0.50 &0.75 &0.55 &0.80 &0.65 &0.80 \\
9 &0.10 &0.65 &0.60 &0.50 &0.20 &0.50 &0.10 &0.00 &0.10 &0.10 &0.10 &0.00 &0.15 &0.70 &0.70 &0.60 &0.40 &0.20 &0.15 &0.35 &0.35 &0.30 &0.50 &0.55 &0.20 &0.35 &0.70 &0.55 &0.60 &0.45 \\\midrule
\multicolumn{31}{l}{\textit{LIBERO-10}}  \\\midrule
0 &0.15 &0.20 &0.10 &0.15 &0.25 &0.10 &0.00 &0.05 &0.10 &0.00 &0.05 &0.05 &0.25 &0.35 &0.10 &0.35 &0.10 &0.25 &0.15 &0.15 &0.00 &0.05 &0.15 &0.20 &0.10 &0.45 &0.25 &0.05 &0.10 &0.05 \\
1 &0.25 &0.20 &0.20 &0.30 &0.15 &0.25 &0.15 &0.15 &0.15 &0.40 &0.30 &0.15 &0.65 &0.10 &0.60 &0.15 &0.50 &0.45 &0.00 &0.25 &0.35 &0.15 &0.10 &0.15 &0.20 &0.40 &0.15 &0.25 &0.05 &0.45 \\
2 &0.70 &0.60 &0.75 &0.30 &0.60 &0.75 &0.55 &0.45 &0.50 &0.25 &0.45 &0.40 &0.75 &0.55 &0.65 &0.45 &0.80 &0.55 &0.70 &0.80 &0.75 &0.75 &0.65 &0.55 &0.85 &1.00 &0.90 &0.70 &0.80 &0.50 \\
3 &0.50 &0.80 &0.55 &0.55 &0.60 &0.80 &0.40 &0.45 &0.45 &0.60 &0.65 &0.60 &0.80 &0.90 &0.75 &0.75 &0.65 &0.50 &0.75 &0.60 &0.60 &0.75 &0.70 &0.65 &0.80 &0.70 &0.70 &0.70 &0.90 &0.70 \\
4 &0.25 &0.20 &0.05 &0.35 &0.25 &0.30 &0.10 &0.10 &0.05 &0.20 &0.05 &0.05 &0.15 &0.10 &0.15 &0.15 &0.15 &0.05 &0.15 &0.20 &0.15 &0.25 &0.20 &0.35 &0.30 &0.25 &0.30 &0.40 &0.30 &0.25 \\
5 &0.40 &0.60 &0.75 &0.55 &0.70 &0.80 &0.50 &0.65 &0.75 &0.40 &0.40 &0.30 &0.85 &0.65 &0.75 &0.75 &0.55 &0.75 &0.45 &0.55 &0.45 &0.60 &0.70 &0.75 &0.80 &0.90 &0.60 &0.70 &0.70 &0.45 \\
6 &0.20 &0.25 &0.10 &0.40 &0.35 &0.40 &0.05 &0.20 &0.05 &0.25 &0.15 &0.15 &0.20 &0.30 &0.35 &0.10 &0.15 &0.20 &0.20 &0.25 &0.05 &0.40 &0.30 &0.35 &0.30 &0.10 &0.20 &0.20 &0.20 &0.15 \\
7 &0.40 &0.30 &0.25 &0.50 &0.50 &0.50 &0.10 &0.30 &0.60 &0.30 &0.40 &0.20 &0.45 &0.30 &0.25 &0.40 &0.35 &0.35 &0.35 &0.70 &0.25 &0.35 &0.30 &0.25 &0.30 &0.25 &0.30 &0.50 &0.45 &0.40 \\
8 &0.10 &0.10 &0.15 &0.10 &0.30 &0.20 &0.35 &0.20 &0.05 &0.10 &0.10 &0.10 &0.15 &0.20 &0.10 &0.20 &0.00 &0.25 &0.05 &0.20 &0.05 &0.10 &0.30 &0.20 &0.25 &0.30 &0.25 &0.25 &0.05 &0.15 \\
9 &0.20 &0.60 &0.35 &0.40 &0.65 &0.30 &0.35 &0.25 &0.45 &0.45 &0.45 &0.30 &0.40 &0.70 &0.50 &0.50 &0.45 &0.60 &0.50 &0.25 &0.40 &0.40 &0.55 &0.50 &0.00 &0.00 &0.00 &0.50 &0.65 &0.50 \\
\bottomrule
\end{tabular}
}
\end{table}

\begin{table}[!htp]\centering
\caption{\textbf{All results on LIBERO-90.}}\label{tab:appx-libero90}
\vspace{-1.2em}
\resizebox{\linewidth}{!}{
\tablestyle{2pt}{1.05}
\begin{tabular}{l|ccc|ccc|ccc|ccc|ccc|ccc|ccc|ccc|ccc|ccc}\toprule
 &\multicolumn{3}{c|}{MoCoV3} &\multicolumn{3}{c|}{MAE} &\multicolumn{3}{c|}{DINOV2} &\multicolumn{3}{c|}{CLIP} &\multicolumn{3}{c|}{EVA} &\multicolumn{3}{c|}{\makecell{InternViT-\\300M}} &\multicolumn{3}{c|}{\makecell{InternViT-\\6B}} &\multicolumn{3}{c|}{MVP} &\multicolumn{3}{c|}{VC-1} &\multicolumn{3}{c}{SPA} \\\midrule
Seed &100 &200 &300 &100 &200 &300 &100 &200 &300 &100 &200 &300 &100 &200 &300 &100 &200 &300 &100 &200 &300 &100 &200 &300 &100 &200 &300 &100 &200 &300 \\\midrule
\multicolumn{31}{l}{\textit{LIBERO-90}}  \\\midrule
0 &0.95 &0.85 &0.90 &1.00 &0.90 &0.80 &0.80 &1.00 &0.60 &0.90 &0.80 &0.80 &1.00 &1.00 &1.00 &0.90 &0.80 &0.85 &0.75 &0.80 &0.95 &0.95 &0.95 &1.00 &0.95 &1.00 &0.95 &1.00 &1.00 &0.95 \\
1 &0.60 &0.35 &0.60 &0.35 &0.50 &0.15 &0.50 &0.55 &0.30 &0.40 &0.65 &0.35 &0.70 &0.50 &0.25 &0.30 &0.45 &0.40 &0.25 &0.35 &0.55 &0.80 &0.55 &0.30 &0.40 &0.50 &0.05 &0.65 &0.40 &0.50 \\
2 &0.85 &0.50 &0.80 &0.55 &0.55 &0.20 &0.65 &0.60 &0.30 &0.45 &0.30 &0.50 &0.35 &0.50 &0.70 &0.85 &0.65 &0.80 &0.25 &0.35 &0.30 &0.45 &0.70 &0.70 &0.75 &0.55 &0.35 &0.70 &0.85 &0.60 \\
3 &0.10 &0.10 &0.00 &0.05 &0.00 &0.00 &0.05 &0.00 &0.00 &0.15 &0.00 &0.00 &0.00 &0.10 &0.15 &0.00 &0.05 &0.05 &0.00 &0.00 &0.05 &0.10 &0.00 &0.05 &0.05 &0.10 &0.00 &0.10 &0.10 &0.00 \\
4 &0.40 &0.05 &0.20 &0.30 &0.25 &0.30 &0.15 &0.40 &0.55 &0.40 &0.40 &0.35 &0.10 &0.25 &0.15 &0.40 &0.05 &0.25 &0.20 &0.45 &0.40 &0.30 &0.40 &0.15 &0.25 &0.05 &0.15 &0.15 &0.15 &0.35 \\
5 &0.05 &0.05 &0.05 &0.10 &0.05 &0.20 &0.00 &0.20 &0.00 &0.05 &0.05 &0.20 &0.25 &0.25 &0.10 &0.10 &0.05 &0.30 &0.20 &0.10 &0.25 &0.05 &0.15 &0.10 &0.10 &0.05 &0.05 &0.35 &0.00 &0.15 \\
6 &0.10 &0.00 &0.00 &0.00 &0.00 &0.05 &0.05 &0.05 &0.10 &0.05 &0.10 &0.00 &0.00 &0.00 &0.05 &0.00 &0.00 &0.00 &0.05 &0.10 &0.10 &0.00 &0.05 &0.00 &0.00 &0.00 &0.00 &0.00 &0.05 &0.05 \\
7 &0.35 &0.30 &0.65 &0.20 &0.60 &0.30 &0.35 &0.25 &0.40 &0.50 &0.60 &0.35 &0.60 &0.10 &0.20 &0.15 &0.25 &0.10 &0.40 &0.25 &0.45 &0.50 &0.20 &0.40 &0.20 &0.30 &0.25 &0.30 &0.30 &0.65 \\
8 &0.10 &0.15 &0.00 &0.05 &0.20 &0.10 &0.15 &0.25 &0.10 &0.20 &0.10 &0.10 &0.10 &0.05 &0.00 &0.05 &0.00 &0.10 &0.20 &0.15 &0.20 &0.10 &0.00 &0.20 &0.05 &0.15 &0.20 &0.05 &0.05 &0.15 \\
9 &0.30 &0.25 &0.35 &0.50 &0.25 &0.30 &0.35 &0.60 &0.70 &0.25 &0.20 &0.50 &0.25 &0.10 &0.50 &0.10 &0.10 &0.25 &0.60 &0.25 &0.30 &0.25 &0.15 &0.45 &0.25 &0.05 &0.35 &0.25 &0.20 &0.25 \\
10 &0.50 &0.75 &0.50 &0.50 &0.60 &0.55 &0.65 &0.60 &0.60 &0.90 &0.45 &0.55 &0.40 &0.85 &0.35 &0.35 &0.05 &0.25 &0.45 &0.45 &0.65 &0.40 &0.50 &0.55 &0.45 &0.75 &0.40 &0.40 &0.35 &0.35 \\
11 &0.45 &0.35 &0.75 &0.45 &0.70 &0.65 &0.35 &0.20 &0.15 &0.40 &0.70 &0.55 &0.80 &0.25 &0.70 &0.50 &0.50 &0.10 &0.35 &0.25 &0.45 &0.80 &0.60 &0.95 &0.70 &0.75 &0.60 &0.60 &0.60 &0.65 \\
12 &0.15 &0.15 &0.10 &0.15 &0.15 &0.05 &0.20 &0.20 &0.15 &0.10 &0.05 &0.05 &0.10 &0.25 &0.05 &0.05 &0.00 &0.00 &0.25 &0.30 &0.10 &0.15 &0.10 &0.10 &0.20 &0.25 &0.10 &0.05 &0.10 &0.15 \\
13 &0.20 &0.35 &0.30 &0.15 &0.30 &0.20 &0.30 &0.35 &0.10 &0.30 &0.40 &0.35 &0.30 &0.10 &0.45 &0.20 &0.35 &0.40 &0.25 &0.15 &0.55 &0.30 &0.30 &0.15 &0.45 &0.10 &0.10 &0.10 &0.20 &0.10 \\
14 &0.05 &0.10 &0.00 &0.30 &0.30 &0.20 &0.10 &0.10 &0.15 &0.15 &0.40 &0.20 &0.25 &0.35 &0.10 &0.15 &0.15 &0.05 &0.20 &0.15 &0.10 &0.20 &0.35 &0.10 &0.20 &0.10 &0.20 &0.15 &0.15 &0.10 \\
15 &0.60 &0.75 &0.45 &0.70 &0.50 &0.65 &0.35 &0.50 &0.55 &0.45 &0.65 &0.40 &0.70 &0.75 &0.40 &0.40 &0.65 &0.50 &0.35 &0.55 &0.45 &0.70 &0.60 &0.55 &0.80 &0.80 &0.70 &0.65 &0.80 &0.55 \\
16 &0.05 &0.20 &0.00 &0.30 &0.15 &0.05 &0.10 &0.10 &0.05 &0.10 &0.00 &0.10 &0.20 &0.20 &0.15 &0.15 &0.15 &0.20 &0.05 &0.00 &0.10 &0.15 &0.05 &0.10 &0.00 &0.15 &0.15 &0.05 &0.10 &0.15 \\
17 &0.05 &0.15 &0.15 &0.10 &0.25 &0.05 &0.05 &0.10 &0.05 &0.05 &0.00 &0.05 &0.05 &0.20 &0.15 &0.10 &0.10 &0.15 &0.00 &0.10 &0.00 &0.20 &0.10 &0.20 &0.15 &0.10 &0.00 &0.25 &0.10 &0.10 \\
18 &0.45 &0.40 &0.60 &0.40 &0.75 &0.65 &0.30 &0.35 &0.40 &0.45 &0.25 &0.35 &0.25 &0.35 &0.60 &0.40 &0.05 &0.70 &0.60 &0.50 &0.35 &0.35 &0.25 &0.45 &0.30 &0.60 &0.35 &0.60 &0.35 &0.55 \\
19 &0.30 &0.30 &0.25 &0.35 &0.40 &0.20 &0.20 &0.05 &0.35 &0.45 &0.45 &0.30 &0.30 &0.35 &0.25 &0.15 &0.25 &0.20 &0.35 &0.30 &0.15 &0.55 &0.30 &0.40 &0.40 &0.45 &0.35 &0.45 &0.20 &0.35 \\
20 &0.85 &0.75 &0.80 &1.00 &1.00 &0.95 &0.95 &1.00 &1.00 &0.75 &0.85 &0.30 &1.00 &1.00 &1.00 &0.95 &0.95 &0.90 &1.00 &0.50 &0.80 &0.90 &1.00 &1.00 &1.00 &1.00 &1.00 &1.00 &1.00 &1.00 \\
21 &0.40 &0.20 &0.40 &0.35 &0.25 &0.30 &0.25 &0.40 &0.20 &0.25 &0.10 &0.45 &0.30 &0.70 &0.05 &0.00 &0.05 &0.10 &0.35 &0.10 &0.30 &0.40 &0.15 &0.30 &0.30 &0.70 &0.60 &0.65 &0.40 &0.60 \\
22 &0.90 &0.95 &0.95 &1.00 &0.85 &0.95 &0.25 &0.60 &0.40 &0.75 &0.75 &0.75 &0.95 &1.00 &0.95 &0.85 &0.95 &0.60 &0.45 &0.25 &0.25 &0.90 &1.00 &1.00 &0.90 &0.90 &0.95 &1.00 &0.95 &1.00 \\
23 &0.15 &0.05 &0.15 &0.05 &0.10 &0.00 &0.05 &0.10 &0.05 &0.00 &0.00 &0.05 &0.00 &0.00 &0.00 &0.00 &0.00 &0.00 &0.20 &0.25 &0.05 &0.00 &0.00 &0.00 &0.00 &0.00 &0.00 &0.00 &0.00 &0.00 \\
24 &0.80 &0.30 &0.85 &0.85 &0.50 &0.80 &0.60 &0.50 &0.65 &0.70 &0.45 &0.60 &0.70 &0.70 &0.80 &0.40 &0.65 &0.60 &0.55 &0.80 &0.45 &0.65 &0.60 &0.90 &0.90 &0.80 &0.80 &0.90 &0.80 &0.75 \\
25 &1.00 &0.80 &0.85 &1.00 &1.00 &0.90 &0.75 &0.90 &0.90 &0.80 &0.95 &0.90 &0.90 &1.00 &0.95 &0.70 &0.70 &0.85 &0.95 &0.60 &0.65 &1.00 &0.85 &1.00 &1.00 &1.00 &0.90 &1.00 &1.00 &1.00 \\
26 &0.15 &0.20 &0.25 &0.25 &0.40 &0.40 &0.05 &0.30 &0.40 &0.45 &0.05 &0.15 &0.05 &0.30 &0.15 &0.25 &0.40 &0.20 &0.25 &0.15 &0.20 &0.20 &0.30 &0.60 &0.25 &0.45 &0.25 &0.25 &0.20 &0.20 \\
27 &0.30 &0.15 &0.20 &0.35 &0.35 &0.10 &0.05 &0.10 &0.00 &0.35 &0.05 &0.10 &0.05 &0.20 &0.05 &0.05 &0.15 &0.00 &0.10 &0.10 &0.05 &0.35 &0.20 &0.30 &0.10 &0.45 &0.40 &0.10 &0.40 &0.20 \\
28 &0.90 &0.90 &1.00 &0.95 &0.70 &0.70 &0.80 &0.50 &0.80 &0.90 &0.75 &0.90 &0.95 &1.00 &0.85 &0.90 &0.85 &1.00 &0.50 &0.60 &0.45 &0.85 &0.75 &0.90 &0.75 &0.95 &0.60 &0.90 &0.65 &0.90 \\
29 &0.15 &0.50 &0.35 &0.60 &0.55 &0.20 &0.50 &0.50 &0.40 &0.50 &0.50 &0.30 &0.65 &0.30 &0.30 &0.15 &0.30 &0.40 &0.25 &0.35 &0.25 &0.35 &0.50 &0.25 &0.60 &0.60 &0.10 &0.30 &0.35 &0.70 \\
30 &0.15 &0.25 &0.15 &0.60 &0.35 &0.35 &0.35 &0.10 &0.50 &0.25 &0.20 &0.45 &0.30 &0.70 &0.20 &0.10 &0.15 &0.20 &0.50 &0.20 &0.25 &0.00 &0.05 &0.20 &0.10 &0.25 &0.10 &0.40 &0.25 &0.40 \\
31 &0.70 &0.60 &0.80 &0.70 &0.75 &0.60 &0.45 &0.70 &0.75 &0.95 &0.65 &0.95 &0.80 &0.75 &0.45 &0.50 &0.55 &0.35 &0.95 &0.80 &0.85 &0.80 &0.70 &0.75 &0.75 &0.75 &0.45 &0.70 &0.90 &0.80 \\
32 &0.30 &0.05 &0.20 &0.05 &0.10 &0.05 &0.00 &0.00 &0.05 &0.35 &0.10 &0.10 &0.20 &0.10 &0.15 &0.10 &0.15 &0.10 &0.05 &0.15 &0.05 &0.20 &0.25 &0.10 &0.05 &0.10 &0.05 &0.20 &0.10 &0.05 \\
33 &0.50 &0.55 &0.40 &0.15 &0.30 &0.30 &0.10 &0.20 &0.35 &0.30 &0.25 &0.30 &0.00 &0.50 &0.40 &0.35 &0.20 &0.25 &0.25 &0.25 &0.30 &0.20 &0.35 &0.40 &0.35 &0.45 &0.65 &0.15 &0.15 &0.15 \\
34 &0.30 &0.35 &0.30 &0.40 &0.40 &0.25 &0.35 &0.40 &0.15 &0.40 &0.40 &0.50 &0.10 &0.40 &0.10 &0.15 &0.05 &0.25 &0.35 &0.30 &0.50 &0.25 &0.30 &0.30 &0.55 &0.25 &0.05 &0.05 &0.30 &0.10 \\
35 &0.65 &0.40 &0.60 &0.85 &0.95 &0.75 &0.80 &0.80 &0.60 &0.65 &0.55 &0.90 &0.90 &1.00 &0.80 &0.10 &0.20 &0.55 &0.85 &0.75 &0.70 &0.85 &0.80 &0.85 &0.25 &0.80 &0.55 &1.00 &0.70 &1.00 \\
36 &0.05 &0.10 &0.15 &0.05 &0.00 &0.00 &0.00 &0.25 &0.05 &0.05 &0.05 &0.00 &0.15 &0.20 &0.10 &0.15 &0.10 &0.25 &0.00 &0.00 &0.10 &0.20 &0.00 &0.15 &0.30 &0.05 &0.20 &0.05 &0.20 &0.10 \\
37 &0.35 &0.30 &0.40 &0.55 &0.20 &0.25 &0.65 &0.50 &0.35 &0.30 &0.60 &0.60 &0.70 &0.80 &0.60 &0.45 &0.55 &0.45 &0.50 &0.55 &0.55 &0.45 &0.45 &0.50 &0.45 &0.75 &0.35 &0.75 &0.50 &0.55 \\
38 &0.50 &0.30 &0.45 &0.55 &0.50 &0.35 &0.25 &0.15 &0.30 &0.50 &0.45 &0.30 &0.50 &0.20 &0.35 &0.35 &0.55 &0.45 &0.35 &0.50 &0.35 &0.70 &0.40 &0.55 &0.70 &0.65 &0.30 &0.70 &0.45 &0.45 \\
39 &0.80 &0.80 &0.75 &0.45 &0.70 &0.60 &0.60 &0.55 &0.65 &0.60 &0.65 &0.70 &0.65 &0.85 &0.55 &0.60 &0.15 &0.60 &0.65 &0.60 &0.60 &0.45 &0.40 &0.70 &0.30 &0.65 &0.35 &0.60 &0.55 &0.60 \\
40 &0.40 &0.50 &0.20 &0.40 &0.30 &0.40 &0.30 &0.45 &0.55 &0.50 &0.25 &0.30 &0.70 &0.65 &0.30 &0.25 &0.60 &0.25 &0.55 &0.40 &0.25 &0.65 &0.70 &0.80 &0.35 &0.45 &0.25 &0.55 &0.35 &0.30 \\
41 &0.20 &0.60 &0.40 &0.50 &0.45 &0.60 &0.35 &0.25 &0.55 &0.20 &0.45 &0.50 &0.85 &0.45 &0.50 &0.45 &0.65 &0.45 &0.20 &0.30 &0.35 &0.55 &0.80 &0.20 &0.35 &0.50 &0.65 &0.65 &0.80 &0.40 \\
42 &0.20 &0.40 &0.20 &0.35 &0.15 &0.05 &0.30 &0.30 &0.55 &0.35 &0.05 &0.20 &0.70 &0.45 &0.20 &0.40 &0.65 &0.60 &0.25 &0.10 &0.05 &0.30 &0.50 &0.65 &0.35 &0.60 &0.50 &0.60 &0.60 &0.40 \\
43 &0.50 &0.50 &0.50 &0.35 &0.40 &0.30 &0.35 &0.35 &0.15 &0.35 &0.25 &0.45 &0.20 &0.60 &0.50 &0.35 &0.35 &0.05 &0.30 &0.30 &0.15 &0.30 &0.60 &0.40 &0.55 &0.40 &0.20 &0.55 &0.70 &0.40 \\
44 &0.90 &0.90 &0.95 &1.00 &1.00 &0.85 &0.80 &0.90 &0.95 &0.75 &0.85 &0.85 &0.95 &0.80 &1.00 &1.00 &0.95 &0.85 &0.85 &0.65 &0.95 &0.85 &0.95 &0.90 &0.95 &0.95 &0.95 &0.95 &1.00 &0.95 \\
45 &0.55 &0.40 &0.60 &0.80 &0.75 &0.60 &0.45 &0.55 &0.60 &0.70 &0.70 &0.55 &0.45 &0.40 &0.30 &0.55 &0.15 &0.55 &0.40 &0.65 &0.30 &0.65 &0.50 &0.35 &0.55 &0.65 &0.45 &0.60 &0.75 &0.60 \\
46 &0.00 &0.00 &0.00 &0.05 &0.35 &0.10 &0.00 &0.15 &0.00 &0.10 &0.10 &0.05 &0.05 &0.10 &0.00 &0.00 &0.00 &0.10 &0.10 &0.00 &0.05 &0.05 &0.05 &0.20 &0.00 &0.05 &0.10 &0.10 &0.00 &0.10 \\
47 &0.45 &0.35 &0.25 &0.25 &0.20 &0.20 &0.10 &0.20 &0.15 &0.60 &0.25 &0.45 &0.45 &0.30 &0.30 &0.15 &0.30 &0.30 &0.15 &0.35 &0.20 &0.40 &0.40 &0.30 &0.20 &0.25 &0.10 &0.55 &0.45 &0.30 \\
48 &0.00 &0.15 &0.30 &0.10 &0.00 &0.15 &0.20 &0.05 &0.20 &0.10 &0.10 &0.25 &0.05 &0.10 &0.25 &0.15 &0.00 &0.25 &0.20 &0.10 &0.20 &0.05 &0.10 &0.10 &0.15 &0.20 &0.10 &0.10 &0.05 &0.20 \\
49 &0.05 &0.05 &0.25 &0.25 &0.30 &0.15 &0.10 &0.20 &0.25 &0.25 &0.40 &0.30 &0.15 &0.25 &0.05 &0.05 &0.00 &0.15 &0.25 &0.30 &0.00 &0.00 &0.00 &0.15 &0.15 &0.05 &0.25 &0.15 &0.05 &0.10 \\
50 &0.25 &0.05 &0.25 &0.00 &0.05 &0.10 &0.00 &0.00 &0.00 &0.00 &0.00 &0.00 &0.00 &0.05 &0.10 &0.05 &0.05 &0.00 &0.00 &0.05 &0.00 &0.05 &0.05 &0.05 &0.10 &0.00 &0.00 &0.05 &0.15 &0.05 \\
51 &0.05 &0.15 &0.05 &0.10 &0.15 &0.05 &0.15 &0.15 &0.20 &0.15 &0.20 &0.15 &0.15 &0.10 &0.05 &0.05 &0.20 &0.05 &0.35 &0.15 &0.35 &0.15 &0.20 &0.30 &0.15 &0.00 &0.10 &0.05 &0.10 &0.10 \\
52 &0.00 &0.00 &0.00 &0.00 &0.00 &0.05 &0.05 &0.10 &0.00 &0.00 &0.05 &0.05 &0.10 &0.00 &0.15 &0.10 &0.00 &0.00 &0.05 &0.05 &0.00 &0.05 &0.15 &0.10 &0.00 &0.00 &0.10 &0.15 &0.00 &0.05 \\
53 &0.05 &0.05 &0.10 &0.15 &0.00 &0.05 &0.20 &0.20 &0.05 &0.00 &0.10 &0.20 &0.05 &0.25 &0.05 &0.05 &0.00 &0.00 &0.25 &0.15 &0.25 &0.00 &0.00 &0.00 &0.05 &0.15 &0.10 &0.10 &0.05 &0.15 \\
54 &0.05 &0.20 &0.05 &0.05 &0.05 &0.00 &0.05 &0.05 &0.05 &0.25 &0.05 &0.05 &0.05 &0.05 &0.10 &0.20 &0.10 &0.00 &0.20 &0.20 &0.05 &0.05 &0.00 &0.00 &0.00 &0.00 &0.00 &0.05 &0.15 &0.05 \\
55 &0.10 &0.05 &0.05 &0.00 &0.00 &0.00 &0.10 &0.05 &0.00 &0.05 &0.00 &0.00 &0.00 &0.00 &0.05 &0.00 &0.00 &0.00 &0.05 &0.05 &0.05 &0.00 &0.05 &0.05 &0.00 &0.00 &0.00 &0.00 &0.10 &0.00 \\
56 &0.05 &0.15 &0.10 &0.10 &0.10 &0.05 &0.15 &0.15 &0.30 &0.20 &0.10 &0.25 &0.10 &0.10 &0.05 &0.10 &0.20 &0.00 &0.25 &0.15 &0.10 &0.05 &0.00 &0.05 &0.20 &0.15 &0.10 &0.10 &0.15 &0.20 \\
57 &0.10 &0.00 &0.05 &0.10 &0.10 &0.05 &0.05 &0.20 &0.00 &0.15 &0.10 &0.05 &0.10 &0.15 &0.00 &0.05 &0.05 &0.10 &0.20 &0.25 &0.10 &0.00 &0.00 &0.20 &0.10 &0.00 &0.20 &0.20 &0.05 &0.05 \\
58 &0.05 &0.10 &0.20 &0.00 &0.05 &0.10 &0.10 &0.25 &0.10 &0.25 &0.20 &0.15 &0.20 &0.15 &0.15 &0.05 &0.05 &0.15 &0.05 &0.15 &0.05 &0.15 &0.10 &0.10 &0.05 &0.10 &0.20 &0.15 &0.05 &0.10 \\
59 &0.05 &0.10 &0.10 &0.05 &0.10 &0.10 &0.05 &0.15 &0.00 &0.05 &0.10 &0.05 &0.05 &0.10 &0.25 &0.05 &0.15 &0.25 &0.05 &0.10 &0.05 &0.10 &0.00 &0.10 &0.15 &0.15 &0.00 &0.00 &0.20 &0.05 \\
60 &0.55 &0.65 &0.70 &0.30 &0.60 &0.45 &0.50 &0.50 &0.40 &0.40 &0.35 &0.65 &0.20 &0.75 &0.55 &0.35 &0.35 &0.50 &0.40 &0.40 &0.50 &0.45 &0.50 &0.75 &0.20 &0.50 &0.55 &0.45 &0.65 &0.90 \\
61 &0.05 &0.15 &0.05 &0.00 &0.10 &0.00 &0.00 &0.10 &0.05 &0.05 &0.10 &0.10 &0.05 &0.10 &0.05 &0.00 &0.00 &0.10 &0.15 &0.05 &0.20 &0.10 &0.15 &0.20 &0.05 &0.15 &0.05 &0.15 &0.05 &0.15 \\
62 &0.45 &0.40 &0.40 &0.45 &0.20 &0.20 &0.25 &0.30 &0.25 &0.55 &0.15 &0.70 &0.40 &0.50 &0.45 &0.30 &0.10 &0.50 &0.40 &0.15 &0.05 &0.20 &0.15 &0.35 &0.25 &0.35 &0.50 &0.40 &0.10 &0.35 \\
63 &0.10 &0.00 &0.05 &0.15 &0.05 &0.00 &0.15 &0.20 &0.15 &0.05 &0.00 &0.00 &0.00 &0.00 &0.00 &0.00 &0.00 &0.00 &0.00 &0.00 &0.00 &0.00 &0.00 &0.10 &0.00 &0.00 &0.00 &0.00 &0.00 &0.00 \\
64 &0.10 &0.05 &0.05 &0.00 &0.05 &0.00 &0.00 &0.05 &0.05 &0.00 &0.05 &0.15 &0.00 &0.00 &0.00 &0.00 &0.00 &0.00 &0.05 &0.05 &0.00 &0.00 &0.00 &0.00 &0.00 &0.00 &0.00 &0.10 &0.05 &0.00 \\
65 &0.05 &0.00 &0.15 &0.20 &0.05 &0.10 &0.00 &0.05 &0.05 &0.10 &0.05 &0.05 &0.05 &0.00 &0.05 &0.00 &0.05 &0.05 &0.05 &0.10 &0.10 &0.10 &0.00 &0.10 &0.15 &0.05 &0.05 &0.05 &0.05 &0.05 \\
66 &0.20 &0.35 &0.05 &0.05 &0.30 &0.15 &0.40 &0.40 &0.50 &0.20 &0.15 &0.25 &0.10 &0.20 &0.05 &0.25 &0.10 &0.25 &0.15 &0.20 &0.30 &0.20 &0.20 &0.20 &0.10 &0.10 &0.20 &0.35 &0.40 &0.25 \\
67 &0.00 &0.00 &0.05 &0.05 &0.00 &0.15 &0.05 &0.05 &0.05 &0.05 &0.00 &0.05 &0.05 &0.00 &0.00 &0.10 &0.05 &0.00 &0.00 &0.05 &0.00 &0.00 &0.10 &0.00 &0.05 &0.10 &0.10 &0.00 &0.00 &0.00 \\
68 &0.05 &0.40 &0.20 &0.30 &0.25 &0.15 &0.30 &0.40 &0.60 &0.25 &0.20 &0.25 &0.20 &0.25 &0.15 &0.20 &0.25 &0.15 &0.40 &0.50 &0.40 &0.35 &0.30 &0.35 &0.15 &0.15 &0.35 &0.30 &0.15 &0.30 \\
69 &0.15 &0.00 &0.20 &0.10 &0.25 &0.05 &0.00 &0.00 &0.00 &0.10 &0.05 &0.10 &0.30 &0.05 &0.50 &0.20 &0.20 &0.00 &0.10 &0.20 &0.10 &0.30 &0.15 &0.25 &0.30 &0.30 &0.10 &0.25 &0.35 &0.10 \\
70 &0.20 &0.45 &0.05 &0.20 &0.20 &0.20 &0.30 &0.35 &0.20 &0.20 &0.30 &0.25 &0.20 &0.20 &0.30 &0.20 &0.30 &0.25 &0.20 &0.20 &0.20 &0.25 &0.55 &0.40 &0.25 &0.15 &0.50 &0.40 &0.20 &0.40 \\
71 &0.35 &0.25 &0.10 &0.20 &0.20 &0.35 &0.35 &0.20 &0.15 &0.15 &0.30 &0.05 &0.15 &0.15 &0.30 &0.25 &0.10 &0.30 &0.35 &0.05 &0.45 &0.35 &0.10 &0.15 &0.05 &0.10 &0.25 &0.20 &0.45 &0.30 \\
72 &0.10 &0.35 &0.25 &0.05 &0.10 &0.40 &0.05 &0.05 &0.25 &0.05 &0.20 &0.10 &0.20 &0.35 &0.35 &0.10 &0.10 &0.30 &0.05 &0.20 &0.05 &0.25 &0.25 &0.50 &0.15 &0.40 &0.30 &0.05 &0.30 &0.25 \\
73 &0.20 &0.05 &0.25 &0.10 &0.05 &0.15 &0.05 &0.00 &0.10 &0.20 &0.15 &0.15 &0.00 &0.15 &0.00 &0.25 &0.10 &0.25 &0.15 &0.05 &0.15 &0.05 &0.05 &0.25 &0.20 &0.25 &0.05 &0.10 &0.10 &0.30 \\
74 &0.30 &0.40 &0.25 &0.20 &0.25 &0.40 &0.20 &0.30 &0.15 &0.40 &0.15 &0.25 &0.35 &0.30 &0.25 &0.15 &0.10 &0.25 &0.45 &0.35 &0.15 &0.40 &0.25 &0.45 &0.40 &0.50 &0.20 &0.50 &0.15 &0.25 \\
75 &0.20 &0.30 &0.15 &0.20 &0.10 &0.25 &0.10 &0.20 &0.10 &0.05 &0.20 &0.15 &0.20 &0.20 &0.20 &0.25 &0.15 &0.20 &0.10 &0.10 &0.10 &0.20 &0.35 &0.55 &0.20 &0.20 &0.15 &0.10 &0.20 &0.25 \\
76 &0.15 &0.15 &0.40 &0.75 &0.50 &0.20 &0.40 &0.40 &0.35 &0.45 &0.25 &0.30 &0.00 &0.30 &0.05 &0.00 &0.00 &0.00 &0.60 &0.20 &0.30 &0.05 &0.25 &0.40 &0.10 &0.05 &0.10 &0.10 &0.05 &0.05 \\
77 &0.35 &0.35 &0.10 &0.05 &0.05 &0.30 &0.25 &0.50 &0.35 &0.15 &0.00 &0.40 &0.15 &0.15 &0.50 &0.30 &0.00 &0.10 &0.15 &0.20 &0.20 &0.15 &0.25 &0.30 &0.55 &0.45 &0.55 &0.35 &0.15 &0.40 \\
78 &0.10 &0.20 &0.25 &0.00 &0.20 &0.15 &0.10 &0.00 &0.00 &0.10 &0.10 &0.05 &0.10 &0.40 &0.20 &0.05 &0.00 &0.10 &0.05 &0.05 &0.00 &0.20 &0.05 &0.05 &0.20 &0.30 &0.10 &0.15 &0.00 &0.25 \\
79 &0.15 &0.40 &0.35 &0.25 &0.20 &0.40 &0.15 &0.45 &0.40 &0.35 &0.35 &0.20 &0.35 &0.25 &0.35 &0.35 &0.00 &0.20 &0.30 &0.15 &0.30 &0.25 &0.15 &0.60 &0.50 &0.65 &0.25 &0.35 &0.05 &0.25 \\
80 &0.15 &0.30 &0.40 &0.20 &0.40 &0.45 &0.25 &0.20 &0.35 &0.10 &0.05 &0.15 &0.40 &0.20 &0.20 &0.20 &0.00 &0.15 &0.30 &0.25 &0.50 &0.25 &0.40 &0.45 &0.60 &0.20 &0.30 &0.40 &0.00 &0.30 \\
81 &0.15 &0.10 &0.15 &0.00 &0.05 &0.20 &0.25 &0.15 &0.05 &0.10 &0.05 &0.10 &0.05 &0.20 &0.05 &0.15 &0.00 &0.05 &0.00 &0.00 &0.05 &0.25 &0.25 &0.25 &0.00 &0.15 &0.15 &0.20 &0.25 &0.20 \\
82 &0.20 &0.35 &0.45 &0.55 &0.30 &0.45 &0.15 &0.45 &0.25 &0.40 &0.60 &0.65 &0.70 &0.45 &0.60 &0.20 &0.00 &0.15 &0.35 &0.45 &0.50 &0.25 &0.45 &0.65 &0.45 &0.65 &0.15 &0.40 &0.30 &0.45 \\
83 &0.10 &0.35 &0.40 &0.10 &0.15 &0.20 &0.10 &0.15 &0.10 &0.30 &0.00 &0.10 &0.20 &0.15 &0.20 &0.05 &0.10 &0.15 &0.40 &0.20 &0.15 &0.15 &0.00 &0.25 &0.15 &0.25 &0.20 &0.15 &0.10 &0.20 \\
84 &0.05 &0.20 &0.20 &0.15 &0.25 &0.05 &0.20 &0.15 &0.40 &0.35 &0.20 &0.10 &0.30 &0.20 &0.05 &0.05 &0.05 &0.10 &0.50 &0.25 &0.20 &0.20 &0.05 &0.20 &0.20 &0.05 &0.20 &0.05 &0.15 &0.00 \\
85 &0.05 &0.00 &0.05 &0.00 &0.10 &0.05 &0.05 &0.20 &0.00 &0.20 &0.10 &0.15 &0.00 &0.05 &0.05 &0.00 &0.00 &0.00 &0.20 &0.10 &0.05 &0.00 &0.00 &0.05 &0.10 &0.05 &0.00 &0.05 &0.05 &0.00 \\
86 &0.05 &0.25 &0.15 &0.20 &0.60 &0.25 &0.00 &0.05 &0.00 &0.15 &0.25 &0.30 &0.25 &0.25 &0.30 &0.20 &0.05 &0.15 &0.15 &0.20 &0.10 &0.10 &0.00 &0.45 &0.05 &0.20 &0.05 &0.40 &0.30 &0.30 \\
87 &0.40 &0.65 &0.65 &0.25 &0.55 &0.40 &0.30 &0.20 &0.20 &0.40 &0.30 &0.35 &0.55 &0.65 &0.55 &0.25 &0.50 &0.40 &0.45 &0.30 &0.30 &0.30 &0.45 &0.90 &0.35 &0.65 &0.70 &0.50 &0.45 &0.55 \\
88 &0.40 &0.20 &0.55 &0.20 &0.55 &0.35 &0.45 &0.65 &0.65 &0.55 &0.60 &0.55 &0.45 &0.50 &0.50 &0.40 &0.20 &0.50 &0.55 &0.25 &0.20 &0.75 &0.45 &0.40 &0.35 &0.65 &0.30 &0.45 &0.55 &0.50 \\
89 &0.10 &0.00 &0.05 &0.05 &0.10 &0.10 &0.05 &0.20 &0.05 &0.20 &0.05 &0.05 &0.00 &0.10 &0.00 &0.10 &0.25 &0.05 &0.10 &0.15 &0.05 &0.00 &0.25 &0.30 &0.30 &0.05 &0.10 &0.20 &0.20 &0.10 \\
\bottomrule
\end{tabular}
}
\end{table}

\end{document}